\documentclass[]{fairmeta}

\providecommand{\doi}[1]{doi: \href{https://doi.org/#1}{\begingroup\urlstyle{tt}\nolinkurl{#1}\endgroup}}

\usepackage{tikz}
\usetikzlibrary{arrows.meta, positioning, calc, fit, decorations.pathreplacing, shapes.geometric, backgrounds}
\usepackage{algorithm}
\usepackage{algpseudocode}
\usepackage{amsmath}
\usepackage{amssymb}
\usepackage{wrapfig}
\usepackage{tabularx}
\usepackage{titletoc}

\definecolor{tfpred}{HTML}{1B6CA8}
\definecolor{arpred}{HTML}{C05A18}

\title{RoboJEPA: Scaling Robotic {} Latent World Models}

\author[1,2,3,4,*]{Artem Zholus}
\author[1]{Nicolas Beltran-Velez}
\author[1]{Jianhao Yuan}
\author[2,3,4]{Sarath Chandar}
\author[1]{Tushar Nagarajan}
\author[1]{Daniel Severo}
\author[1]{Koustuv Sinha}
\author[1]{Michal Drozdzal}
\author[1]{Adriana Romero Soriano}
\author[1\dagger]{Jeannette Bohg}
\author[1,\dagger]{Nicolas Ballas}
\author[1,\dagger]{Mahmoud Assran}
\affiliation[1]{FAIR at Meta}
\affiliation[2]{Chandar Research Lab}
\affiliation[3]{Mila - Quebec AI Institute}
\affiliation[4]{Polytechnique Montréal}

\contribution[*]{Work done at Meta}
\contribution[\dagger]{Joint last author}

\newcommand{\paperabstract}{

Latent world models have shown a remarkable ability to predict future states and to plan in the real world. In practice, however, we lack a principled way to estimate how their capabilities scale with model size, data, and compute, an open problem that slows progress in the field. In this work we present RoboJEPA, a world model based on the Joint Embedding Predictive Architecture (JEPA) and trained on a large-scale dataset spanning 12 robotic embodiments. We show that RoboJEPA's imagination error, the error of its latent rollouts, follows a second-order power law in compute, allowing us to predict model quality well beyond the scale at which the law is fit. We further show that downstream robotic planning performance improves predictably with compute, and that imagination error is strongly correlated with it, making it a reliable proxy for real-robot evaluation. Finally, we demonstrate that latent world models can be deployed zero-shot as robotic agents, planning toward a single goal image to solve tasks requiring  long-horizon planning on real hardware. We release all model checkpoints together with our training and robot deployment code. To our knowledge, this is the first work to establish scaling laws for multi-embodiment robotic world models trained on real robot data, and RoboJEPA, at 8B parameters, is the largest JEPA predictor model trained to date.

}

\abstract{\paperabstract}
\date{\today}
\correspondence{Artem Zholus at \email{artem.zholus@mila.quebec}, Jeannette Bohg at \email{jeannettebohg@meta.com} }

\metadata[Code]{\url{https://github.com/facebookresearch/robo_jepa}}
\metadata[Website]{\url{https://robojepa.github.io}}

\begin{document}

\maketitle

\noindent
\begin{minipage}{\textwidth}
\centering
\includegraphics[width=\textwidth]{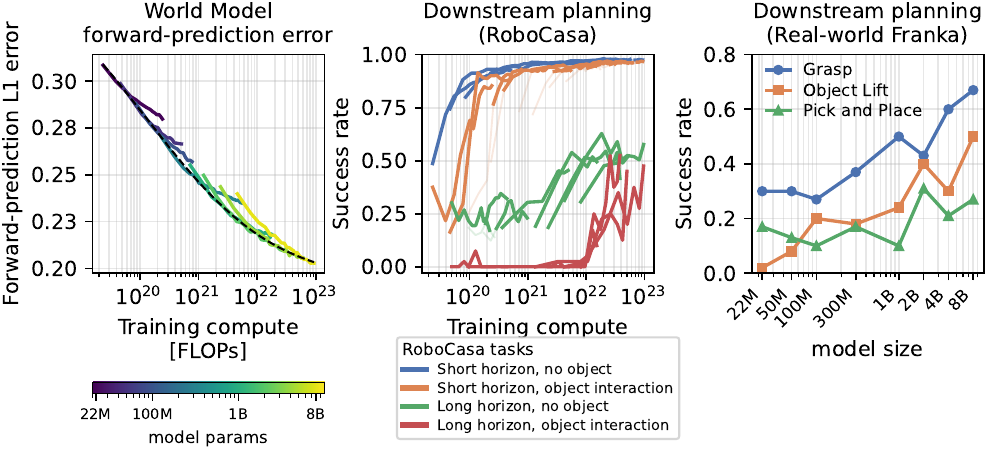}
\captionof{figure}{\textbf{RoboJEPA scales predictably from offline prediction to real-robot control.}
\emph{Left}: the world model's forward-prediction ($\ell_1$) error, averaged over the DROID and RoboCasa held-out sets, follows a scaling law in training compute.
\emph{Middle}: downstream planning success in RoboCasa improves predictably with compute across tasks of increasing difficulty.
\emph{Right}: on the real Franka robot, the success rate improves with model size across Grasp, Object Lift, and Pick and Place.
\textbf{Actions are sampled from a uniform distribution and prioritized using the World Model during planning.}}
\label{fig:visual-abstract}
\end{minipage}
\vspace{6pt}

\section{Introduction}\label{section:intro}

\begin{figure}[t]
\centering
\includegraphics[width=\linewidth]{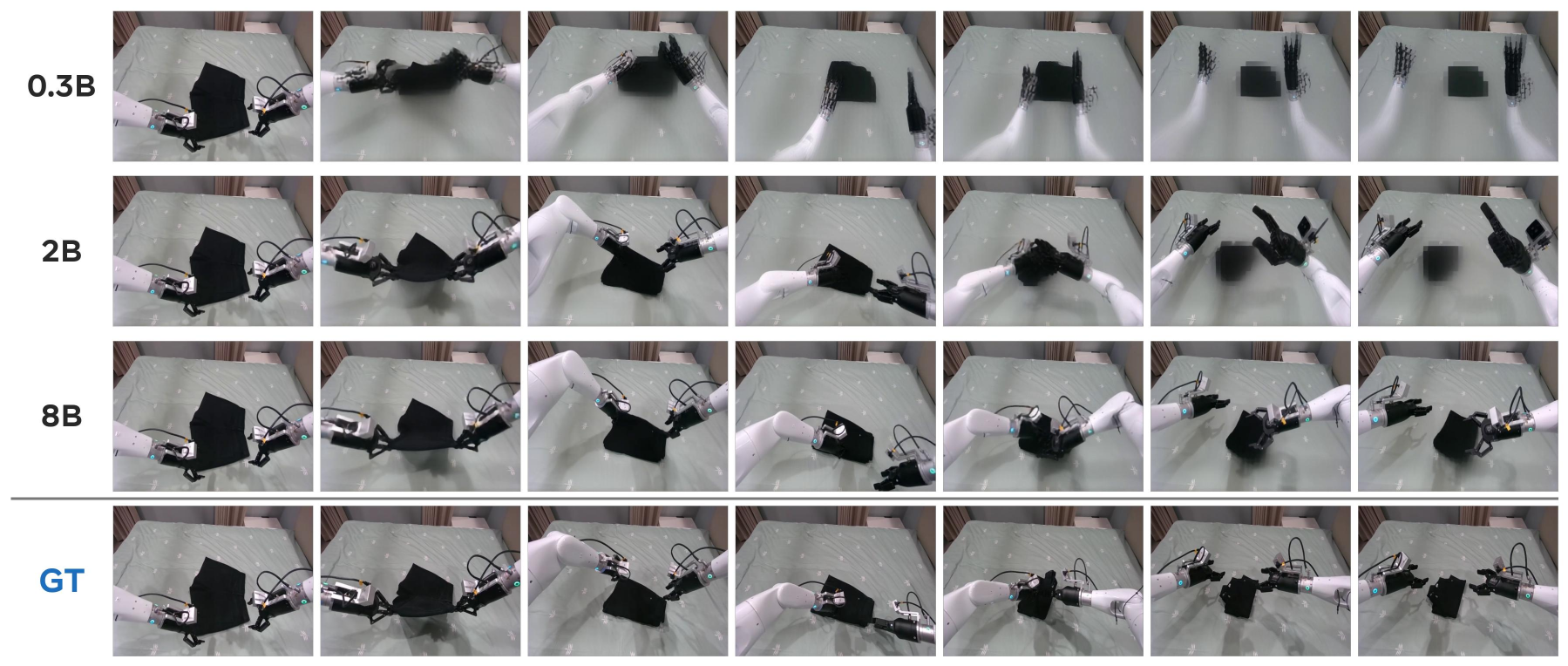}
\caption{\textbf{World-model imagination improves with scale.} Decoded imagination
rollouts on an AgiBot episode for RoboJEPA at $0.3$B, $2$B, and $8$B
parameters, compared against the ground-truth (GT) video (bottom row). Each row
shows the same rollout sampled at equal-time steps; larger models track the true
dynamics and preserve object and contact detail markedly better . All world models share the same encoder (V-JEPA 2.1-G) and decoder. }
\label{fig:imagination_scaling}
\end{figure}

Scaling has been one of the primary engines of recent progress in artificial intelligence.
What makes scaling particularly actionable is \emph{predictability}: empirical scaling laws relate a model's loss to the amount of data, parameters, and compute used to train it, so that the return on an additional order of magnitude of compute can be forecast before it is spent~\citep{kaplan2020scaling,hoffmann2022chinchilla}.  This predictability turns scaling from a gamble into a science, it tells practitioners how large a model to train for a given budget, how much data to pair it with, and when returns will diminish. 
Scaling laws of this kind have been established across language models~\citep{kaplan2020scaling,hoffmann2022chinchilla}, generative modeling of images, video, and other
modalities~\citep{henighan2020scaling}, or semantic
classification of images~\citep{zhai2022scalingvit}.
 Comparatively little is known
about whether, and how, the same  scaling predictability holds for models of the physical world.

We study this question for \emph{world models} in robotics. A world model is a
learned model of environment dynamics that an agent can roll out to imagine
future states and plan~\citep{ha2018worldmodels,hafner2023dreamerv3}. We focus
on latent world models that predict and plan in a latent representation  space~\citep{hafner2020dreamer,lecun2022path, assran2025vjepa2}. Latent world models have shown the ability to plan toward goals specified at test time without the need of a task-specific policy~\citep{assran2025vjepa2}.
Previous latent world models are typically trained at a single scale, and how their quality changes with data, parameters, and compute has not been characterized~\citep{assran2025vjepa2,hafner2025training,zhang2026hierarchical}.
Other studies, focusing on scaling, either target generative video and interactive-environment models outside of robot control~\citep{bruce2024genie,deepmind2024genie2,hu2023gaia1},
or scale \emph{policies} rather than world models~\citep{lin2024datascaling,sartor2024neuralscaling,ai2025embodiment}.
In contrast, our goal is to establish scaling laws for latent world models. This matters particularly in robotics, where high-quality interaction data is scarce and computationally expensive to collect: knowing which capabilities a world model will have at a given training-compute budget lets practitioners decide whether to invest in a larger model, more demonstrations, or more compute, and reveals when a data budget has been saturated.

To this end, this work introduces \textbf{RoboJEPA}, a family of action-conditioned latent joint-embedding predictive world models. We keep the training recipe deliberatly simple to focus on the scaling behavior. \textbf{RoboJEPA} is a feed-forward transformer that predicts the next latent state deterministically in a single forward pass. It is trained
with a plain $\ell_1$ regression loss in the frozen encoder’s latent space. 
\textbf{RoboJEPA} operates in the representation space of a frozen V-JEPA~2.1
encoder~\citep{assran2025vjepa2,mur2026vjepa21} and is trained on a
unified mixture of 23 public manipulation datasets spanning 12 embodiments and 26 different action spaces. We study model scaling of the \emph{world model}, ranging parameter count from 22M up to 8B, while holding the representation encoder fixed, so our laws characterize how a plannable dynamics model scales on
top of a shared representation. We build scaling laws estimating the compute-efficient frontier of the forward-prediction evaluation loss as a function of the training compute~\citep{kaplan2020scaling, pearce2024reconciling}.

Our central finding is that robotic world-model quality scales predictably with
compute. First, the latent forward-prediction loss on out-of-distribution data follows a
scaling law $L(C)=A\,C^{\alpha-\gamma \ln C}+E$, where $C$ is the training compute, $E$ is the irreducible loss, and $A,\alpha,\gamma$ are scaling-law parameters that we estimate alongside $E$.
This holds on both real-robot holdout (DROID~\citep{khazatsky2024droid}) data and simulated holdout (RoboCasa~\citep{nasiriany2026robocasa365}) data.
Second, this improvement on the offline data transfers to downstream planning performance on a robot: on a suite of RoboCasa planning tasks,
downstream success improves predictably with training compute, so that a world
model's latent forward-prediction loss is a strong predictor of its online
planning performance. 
For example, only models trained with more than $10^{22}$ FLOPs of training compute (at sizes 2B, 4B, and 8B) can solve object-interaction tasks in RoboCasa via non-greedy planning, while smaller models trained with less compute cannot (see \Cref{fig:push-success}).

We deploy RoboJEPA on real and simulated robots,
planning toward a single goal image that defines the final state of the environment. Across two robot platforms, we perform planning in over $50{,}000$ evaluation episodes. We keep the planning hyperparameters fixed for each robot, varying only the planning horizon for harder tasks.
We perform planning on a set of real world manipulation tasks that include grasping, lifting, and pick-and-place. In simulation, our planning spans tasks of increasing difficulty from reaching to object pushing.
We observe an improvement trend in the success rate on the real robot as we increase the model size.
Qualitatively, we also observe that the behavior of the planning policies changes as the model grows: for example, in object-manipulation tasks, only models with 2B parameters or more keep the gripper closed after grasping to complete the task, while smaller models often accidentally open it, dropping the object.

Finally, to support open research and reproducibility, we
release all model checkpoints together with the training and robot-deployment
code. To the best of our knowledge, this is the first work to establish
compute-optimal scaling laws for action-conditioned robotic world models
trained on large-scale, multi-embodiment real robot data, and the first scaling law for JEPA with verified extrapolation.
We also train the
largest JEPA \emph{predictor} to date, at 8B parameters.

\section{Method and Data}\label{section:method}

To study the scaling behavior of robotic world models, we require a model family that scales with training compute. 
To this end we introduce RoboJEPA, a family of action-conditioned latent world models . %
The goal of RoboJEPA models is to predict how the latent features of a scene will evolve under the robot's actions. Formally, at timestep $t$, given a history of past visual
features $z_{\le t}$ (where $z_t$ is the latent encoding of a given visual input $x_t$), proprioceptive states $s_{\le t}$, and actions $a_{\le t}$, our models are functions $g_\theta$ that 
deterministically predict the features of the next frame in one neural network forward pass:
\begin{equation}
  \hat{z}_{t+1} \;=\; g_\theta\big([z_{\le t}],\, a_{\le t},\, s_{\le t}\big).
  \label{eq:wm}
\end{equation}
Actions are either state deltas, i.e.\ $s_{t+1}=s_t+a_t$, or absolute robot commands, in which case $s_t=0$.
Feeding predictions back as inputs lets us roll the model out autoregressively over future actions, e.g.
\begin{equation}
  \hat{z}_{t+2} \;=\; g_\theta\big([z_{\le t}, \hat{z}_{t+1}],\, a_{\le t+1},\, s_{\le t+1}\big), \quad
  \hat{z}_{t+3} \;=\; g_\theta\big([z_{\le t}, \hat{z}_{t+1}, \hat{z}_{t+2}],\, a_{\le t+2},\, s_{\le t+2}\big), \quad \dots
  \label{eq:wm-rollout}
\end{equation}
This allows us to autoregressively roll out the world model to predict the effect of a sequence of actions on the latent features of the scene, and consequently to \emph{plan}. 

Planning in RoboJEPA is specified by a goal image encoded once into goal features $z_g$, and at every timestep $t$ we search for the $H$-step action sequence whose imagined latent rollout comes closest to the goal:
\begin{equation}
  a^{\star}_{t:t+H-1} \;=\; \arg\min_{a_{t:t+H-1}}\; \big\lVert \hat{z}_{t+H}(a_{t:t+H-1}) - z_g \big\rVert_1,
  \label{eq:plan}
\end{equation}
where $\hat{z}_{t+H}$ is obtained by rolling $g_\theta$ forward $H$ steps from the current context under the action sequence $a_{t:t+H-1}$. We optimize \Cref{eq:plan} with the cross-entropy method (a concrete instance is given in \Cref{subsec:planning,appx:planning}).

The rest of this section describes the design and training of this family of world models and how to use them for planning, with the goal of subsequently studying their scaling behavior. 

\subsection{RoboJEPA Architecture}\label{subsec:model_architecture}
\begin{figure}[h!]
\centering
\includegraphics[width=\linewidth]{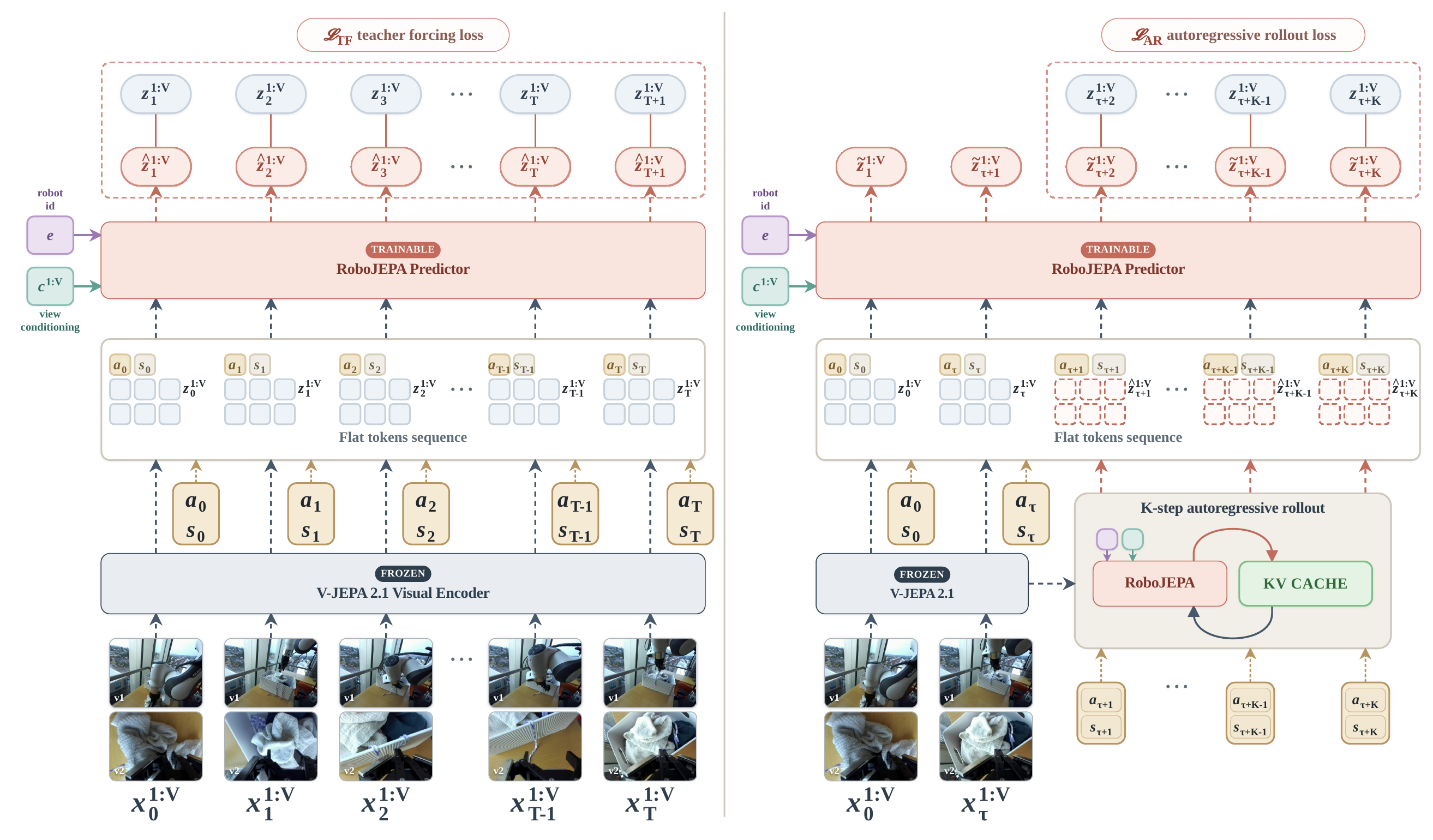}
\caption{RoboJEPA training. The frozen V-JEPA~2.1 encoder $f$~\citep{mur2026vjepa21} maps the input multiview frames $x_t^{(1:V)}$ to a grid of features. The predictor $g_\theta$ consumes per-step sequences of multiview visual ($z_t^{(1:V)}$), action ($a_t$) and proprioceptive state ($s_t$) tokens, conditioned on view and embodiment identifiers, and predicts the next-step features $\hat z_{t+1}$. Training minimises two losses: \textbf{(left)~the teacher-forcing loss}, the per-token $\ell_1$ distance between predicted features and the next-step features from the encoder (\Cref{eq:loss-tf}); and \textbf{(right)~the rollout loss}, which attempts to predict the same target sequence as the teacher-forcing loss, but given fewer input frames. The model is given the full training sequence of actions but only a prefix of the encoded features, and predicts the remaining features autoregressively, with a KV-cache. This way the model learns to make accurate long term predictions.
}
\label{fig:main-method}
\end{figure}

The family of RoboJEPA world models builds on the action-conditioned recipe of
V-JEPA~2~\citep{assran2025vjepa2} and V-JEPA~2.1~\citep{mur2026vjepa21}.
This subsection introduces the architecture of the RoboJEPA predictor. The full list of differences between RoboJEPA and V-JEPA~2-AC is summarised in \Cref{tab:appx-arch-vjepa2ac} and detailed in \Cref{appx:arch}.

Scaling world models across robot datasets is challenging because each dataset comes with a different number of camera views, proprioceptive state size, and controller type and frequency.
Therefore, we aim to build a world model architecture that can work with any combination of views, controller type, and proprioceptive information.
At each timestep $t$, a frozen V-JEPA~2.1-G encoder maps up to $V$ simultaneous camera views at resolution $H \times W$,
each spanning $T$ frames, denoted $x_t^{(1:V)} = (x_t^{(1)}, \dots, x_t^{(V)})$, to the latent features $z_t^{(1:V)} = (z_t^{(1)}, \dots, z_t^{(V)})$, where each $z_t^{(v)}$ is an $H_p \times W_p$ grid of $D_e$-dimensional patch embeddings with $D_e = 1664$ and $H_p = H/16$, $W_p = W/16$.
We use V-JEPA~2.1 as the encoder because early in the work we found that it leads to successful and predictable scaling of the world model, whereas V-JEPA~2 does not (we hypothesize why in Appendix~\ref{appx:encoder-scaling}).
Alongside these features, each timestep provides the action $a_t \in \mathbb{R}^{d_a}$ and proprioceptive state $s_t \in \mathbb{R}^{d_s}$, while each input is associated with a robot type operating in a particular action space, identified by the \textit{robot id} $e \in \mathcal{E}$, and a per-view camera type $c^{(v)} \in \mathcal{C}$. Here, $\mathcal{E}$ and $\mathcal{C}$ are finite sets of known robot ids and camera types.
Next we describe how we tokenize inputs, and the world model architecture.

\begin{table}[t!]
\caption{
    Compressed view of the $23$-dataset, $26$-action-space mixture.
    Video hours sum the durations of all recorded videos.
    Action hours measure the total duration of teleoperation or policy execution,
    counted once regardless of the number of cameras.
    The per-dataset breakdown with resolutions and source FPS is in
    \Cref{tab:appx-datasets-full}; sampling weights are in
    \Cref{tab:appx-mixture-weights}; and semantic-camera mappings
    are in \Cref{tab:appx-data-cameras}. We report dataset sizes
    that we were able to collect and verify, which may be smaller than those
    reported in the original papers.
}
\label{tab:datasets}
\centering
\small
\setlength{\tabcolsep}{4pt}
\resizebox{\textwidth}{!}{%
\begin{tabular}{lrrrl}
\toprule
Dataset group & Episodes & Video hours & Action hours & Embodiments \\
\midrule
DROID~\citep{khazatsky2024droid}                & 92k    & 414  & 138  & Franka \\
Roboset~\citep{bharadhwaj2023roboagent}         & 73k    & 1244 & 380  & Franka (Roboset) \\
RoboMind~\citep{wu2024robomind}                 & 35.3k  & 276  & 98   & Franka, AgileX \\
LeRobot~\citep{cadene2024lerobot}               & 21.5k  & 182  & 93   & SO-101 \\
RoboCasa365~\citep{nasiriany2026robocasa365}          & 32k    & 482  & 482  & Franka + Omron (sim) \\
1X~\citep{1xworldmodel2024}                     & 23.5k  & 90   & 90   & 1X humanoid \\
AgiBot World~\citep{bu2025agibot}               & 167.5k & 8036 & 2734 & AgiBot dual-arm, humanoid \\
Open-X Embodiment~\citep{oxe2023}\footnotemark[1] & 2.4265M & 4298 & 2677 & 8 platforms \\
\midrule
\textbf{Total}                                  & \textbf{2.8713M} & \textbf{15,022} & \textbf{6,692} & 12 platforms / 26 ids \\
\bottomrule
\end{tabular}%
}
\end{table}
\footnotetext[1]{We use 9 datasets of Open-X Embodiment~\citep{oxe2023}: RT-1~\citep{brohan2023rt1}, Bridge~\citep{ebert2021bridge}, Bridge V2~\citep{walke2023bridgev2}, QT-Opt~\citep{kalashnikov2018qtopt}, \mbox{MT-Opt~\citep{kalashnikov2021mtopt}}, RoboNet~\citep{dasari2019robonet}, Language-Table~\citep{lynch2023languagetable}, BC-Z~\citep{jang2022bcz}, FMB~\citep{luo2024fmb}.}

\textbf{Tokenization.} We encode each modality separately and interleave the results into a single
flat sequence. We refer to these encoded multimodal vectors as tokens. The motivation of our tokenization scheme is to let the model learn all temporal dependencies only from a structured sequence of tokens. To
tokenize each $z_t^{(v)}$, we project each
patch from $D_e$ to the world-model width $D$ with a single linear layer, giving a visual token
tensor of shape $[T, V, H_p, W_p, D]$. Actions and proprioceptive states are mapped to $D$ by
linear encoders (specific to robot id) applied to $a_t$ and $s_t$.

We interleave these tokens by timestep,
\begin{equation}
\big(z_1^{(1{:}V)},\,a_1,\,s_1\big),\;\big(z_2^{(1{:}V)},\,a_2,\,s_2\big),\;\dots,\;
\big(z_T^{(1{:}V)},\,a_T,\,s_T\big),
\label{eq:interleave}
\end{equation}
so that each timestep contributes $V H_p W_p$ visual tokens followed by one action and one state
token. Beyond this ordering we
impose no structure: the model consumes a flat sequence and recovers spatio-temporal and view
structure from the position encodings and external conditioning information, such as the robot id and view type.

\textbf{World Model Architecture.} RoboJEPA performs the next embedding prediction via a deterministic model.
RoboJEPA uses the vision transformer~\citep{dosovitskiy2021vit} as the backbone and we parameterize the
prediction such that each spatial input token of $z_{t}^{(1:V)}$ predicts the corresponding
spatial token of $z_{t+1}^{(1:V)}$.
We use the frame-level causal attention mask so that
queries that correspond to timestep $t$ have only access to keys and values at step $t$ or earlier.
The model stacks $L$ blocks that form a standard residual stream of ViT.
The architecture of each block follows the standard recipe for modern variants of ViTs: we use
pre-normalization, Rotary Position Encoding \citep{su2024roformer}, RMSNorm normalization~\citep{zhang2019rmsnorm},
GELU feed-forward block~\citep{hendrycks2016gelu}, and QK-normalization~\citep{dehghani2023vit22b}. We found QK-normalization to be of
particular importance for training stability when scaling the dataset size and the model size.

\textbf{Factorized RoPE and view conditioning.}
The spatio-temporal position information is supplied by a multi-axis Rotary Position
Embedding~\citep{su2024roformer} that splits each head's channels into \emph{four} equal sub-blocks
of width $d_h/4$, one for each of the four axes $(t,i,j,v)$ for time, height, width, and view, where $d_h$ is the head dimension. Our version
of RoPE rotates each independently by its own per-axis frequency (\Cref{appx:arch-rope}). Action and state tokens have a temporal index but no spatial index, i.e.\ the 4D index of the action $a_t$ would be $(t,0,0,0)$.
We combine this 4D RoPE with a per-layer learnable view bias applied after RoPE.
The idea is that the 4D RoPE informs the model that tokens belong to different views, while the view bias informs the model which content each view holds (e.g.\ left camera, wrist camera; \Cref{appx:arch-rope-view}).
This allows us to randomize the order of views and at the same time provide the information which views the model works with.

\textbf{Attention pattern.}
We use joint space-time-view attention and do not factorize it, meaning that each token can have access to
potentially all other tokens in the space-time-view sequence, unless restricted by the attention mask.
However, using no mask would allow past token activations access future information that they need to predict.
For example, token activations of $z_t$ will have access to tokens of $z_{t+1}$ in self-attention.
Therefore, we use a \textit{timestep-causal} mask~\citep{assran2025vjepa2, po2025longcontext, panteam2025pan}.
That is, a query associated with timestep $t$ has access to keys only at timestep $t$ or less.
This includes attention queries and keys associated with different views, actions, and states.
In addition to that, we make the attention \textit{local} by letting queries only access keys that are at most $8$ timesteps in the past~\citep{assran2025vjepa2, po2025longcontext, panteam2025pan}.
This design helps the model extrapolate to predicting longer sequences of embeddings because no query will ever attend to a history longer than the model was trained on, including during long autoregressive prediction.
For implementation details on the attention masking, see \Cref{appx:arch-attention}.

\textbf{Outputs.} After feeding the entire sequence of tokens through the
backbone, we project tokens that correspond to $z^{(1:V)}_{1:T}$
back to the encoder space. We map the output from $D$ dimensions to $D_e$
dimensions and apply layer normalisation. The idea is that since both input
and output features are normalized, this strenghtens the learning signal and
lowers the gap between the training and the autoregressive generation.
We do not use any learnable modules to predict states and we simply compute the proprioceptive state $s_t$ either through integration, i.e.\ $s_{t+1}=s_t+a_t$, or $s_t=0$ (for example when working with commanded actions with non-trivial robot dynamics).

\textbf{Training objective.} We train the world model to predict the next embedding and, to minimize exposure bias, we additionally train it on its own outputs over potentially long horizons.
Our objective therefore has two loss components.
The first component trains the model to predict every next embedding along a sequence of encoded multiview frames.
The second component uses the same objective, but the model is given only a prefix of fewer encoded frames (but the entire sequence of actions); it then predicts the missing embeddings autoregressively (\Cref{eq:wm,eq:wm-rollout}) and is trained on them with the same objective.
We use a key-value cache to speed up this autoregressive prediction (\Cref{subsec:kvcache,appx:kv}).
While the first loss component ensures that the model predicts the representations of a fixed encoder, the second makes the world model robust to its own errors and suppresses error accumulation.
Both components use the same per-token $\ell_1$ objective, and the total loss is their average:
\begin{equation}
\mathcal{L}_{\mathrm{TF}} = \sum_{t,v} \big\lVert \textcolor{tfpred}{\boldsymbol{\hat z}_{t+1}^{(v)}} - z_{t+1}^{(v)} \big\rVert_1,
\quad
\mathcal{L}_{\mathrm{AR}} = \sum_{t,v} \big\lVert \textcolor{arpred}{\boldsymbol{\tilde z}_{t+1}^{(v)}} - z_{t+1}^{(v)} \big\rVert_1,
\quad
\mathcal{L} = \tfrac{1}{2}\mathcal{L}_{\mathrm{TF}} + \tfrac{1}{2}\mathcal{L}_{\mathrm{AR}},
\label{eq:loss-tf}
\end{equation}
where $\textcolor{tfpred}{\boldsymbol{\hat z}_{t+1}^{(v)}}$ is the teacher-forced one-step prediction and $\textcolor{arpred}{\boldsymbol{\tilde z}_{t+1}^{(v)}}$ is the autoregressive prediction rolled out from a shorter prefix, both compared against the encoder feature $z_{t+1}^{(v)}$. The full form of the loss (with the per-frame mask and patch normalization), the rollout schedule, and the optimizer are described in \Cref{subsec:training,appx:training}.

\subsection{Data, Training and Inference}\label{subsec:data}\label{subsec:training}\label{subsec:planning}\label{subsec:kvcache}\label{subsec:inference}

\textbf{Data.} Training a world model that generalizes across robot embodiments and scenes requires both
scale and diversity. We assemble $23$ public manipulation datasets spanning $12$ robot
platforms---single- and dual-arm robots and humanoids---across real and simulated data.
The compiled dataset contains $15{,}022$ hours of video, including $6{,}692$ hours synchronized
with actions; some demonstrations are recorded from multiple cameras.
Some platforms record with only one camera (e.g. most of Open-X Embodiment data),
whereas others use two or three cameras (e.g. DROID and RoboSet).
In addition, the datasets were collected with different robot action spaces, camera poses,
sensor resolutions, and frame rates. Therefore, to build a data-scalable world model, we unify
our training datasets and formulate the architecture and training recipe to be general enough
to work with any robotic dataset. \Cref{tab:datasets} summarizes the mixture, and
\Cref{appx:data} provides further details about its action spaces, unification, and sampling.

\begin{figure}[t!]
\centering
\includegraphics[width=\linewidth]{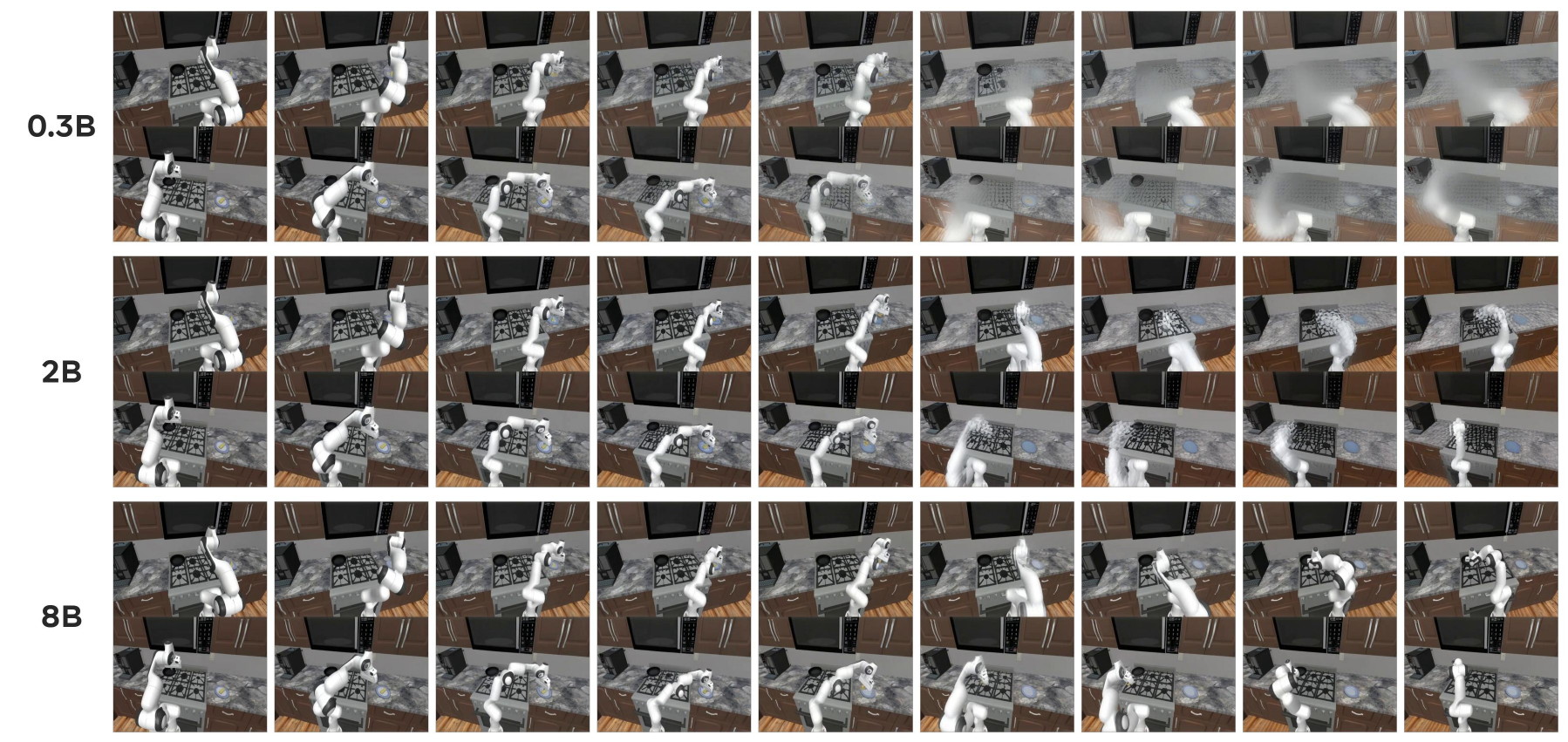}
\caption{\textbf{Two-view imagination scales in simulation.} First $60$\,s of a RoboCasa
rollout (two agent-view cameras, stacked within each cell) imagined by RoboJEPA at $0.3$B,
$2$B, and $8$B parameters. Larger models keep the two views mutually consistent and
preserve manipulation detail markedly deeper into the rollout. All world models share the same encoder (V-JEPA 2.1-G) and decoder.}
\label{fig:imagination_scaling_2view}
\end{figure}

\textbf{Training.} {}
We train a series of RoboJEPA models that range in size from $22$M to $8$B parameters and in training compute from $2{\times}10^{19}$ to $9.5{\times}10^{22}$ FLOPs.
For all models, we use AdamW~\citep{loshchilov2019adamw} with a weight decay of $0.1$ and global gradient-norm clipping at $1.0$.
We linearly warm up the learning rate, hold it constant at a peak of $5\times10^{-5}$ for most of training, and then linearly decay it to zero over a short cooldown phase at the end~\citep{hu2024minicpm}.
This allows us to overlap the flat learning-rate stage across models trained with different amounts of compute: we run the flat stage only once and perform learning-rate annealing from intermediate checkpoints.
For the detailed description of the training configuration and hyperparameters, please refer to Appendix~\ref{appx:training}.

The training recipe of RoboJEPA is designed to facilitate scaling the model on as much data as is available, which requires training on many heterogeneous robotic datasets at once.
A central issue with such heterogeneous data is that some datasets supply only a single view while others provide multiple views, often at different resolutions.
We therefore train the model to work with a variable number of views and a variable resolution.
To achieve this, we partition the GPUs into groups, where all GPUs within a group work with the same number of views and the same resolution.
In practice this often leads to imbalanced step times across groups, which is why we balance these times by using different batch sizes for different groups.
Our initial scaling-law experiments were done at a resolution of $240\times 320$ with only two GPU groups: one working with single-view data and the other with dual-view data.
However, eventually we found it beneficial to include high-resolution GPU groups, resulting in four groups: single-view low-resolution ($240\times 320$), dual-view low-resolution, single-view high-resolution ($480\times 640$), and dual-view high-resolution.
We describe the GPU grouping and batch-size balancing in detail in \Cref{appx:training:batch,appx:training:cooldown}.

\begin{figure}[t!]
\centering
\begin{subfigure}[t]{0.49\linewidth}
\centering
\includegraphics[width=\linewidth]{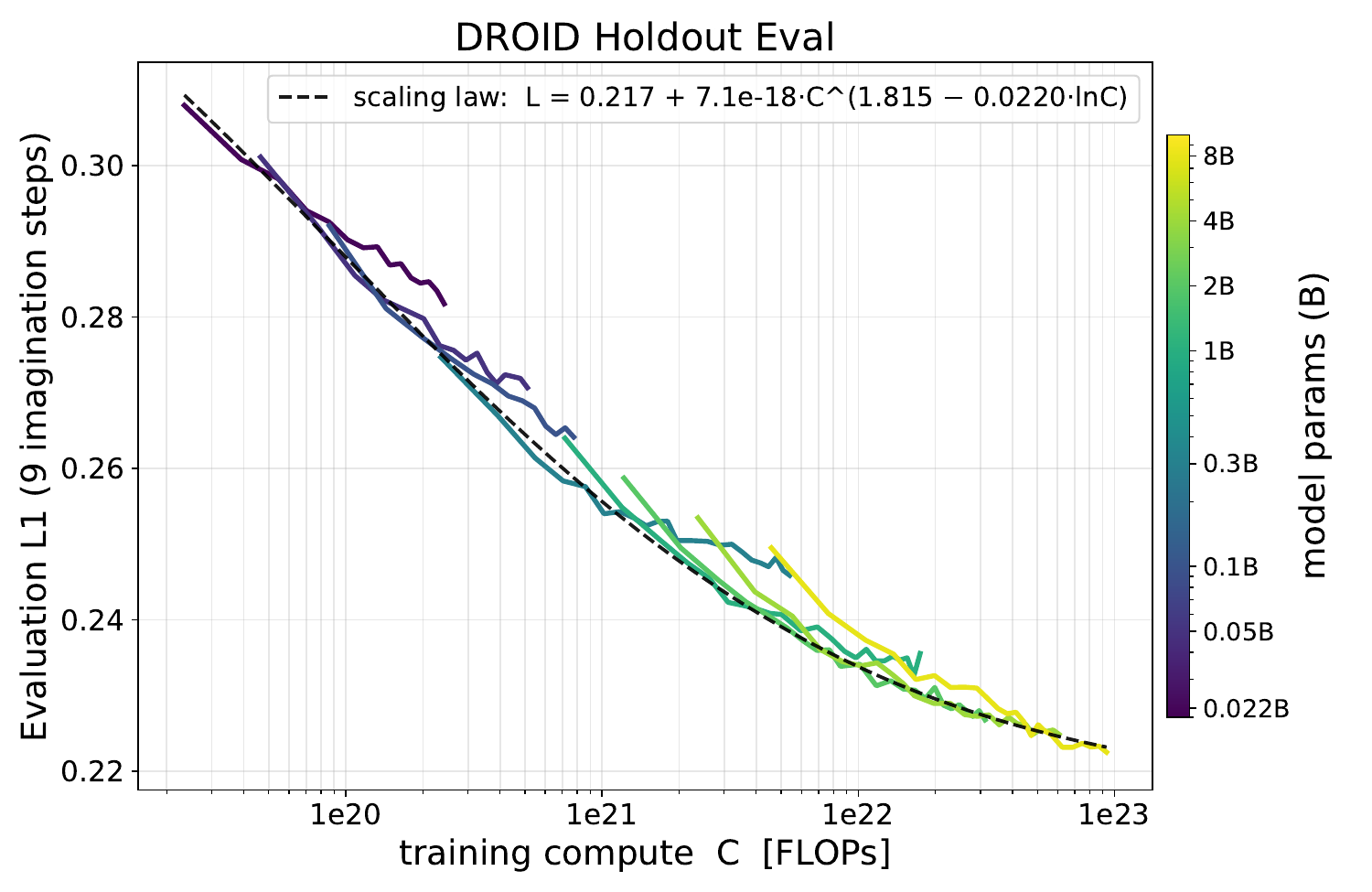}
\caption{DROID holdout (real Franka, unseen scenes).}
\label{fig:scaling-droid}
\end{subfigure}
\hfill
\begin{subfigure}[t]{0.49\linewidth}
\centering
\includegraphics[width=\linewidth]{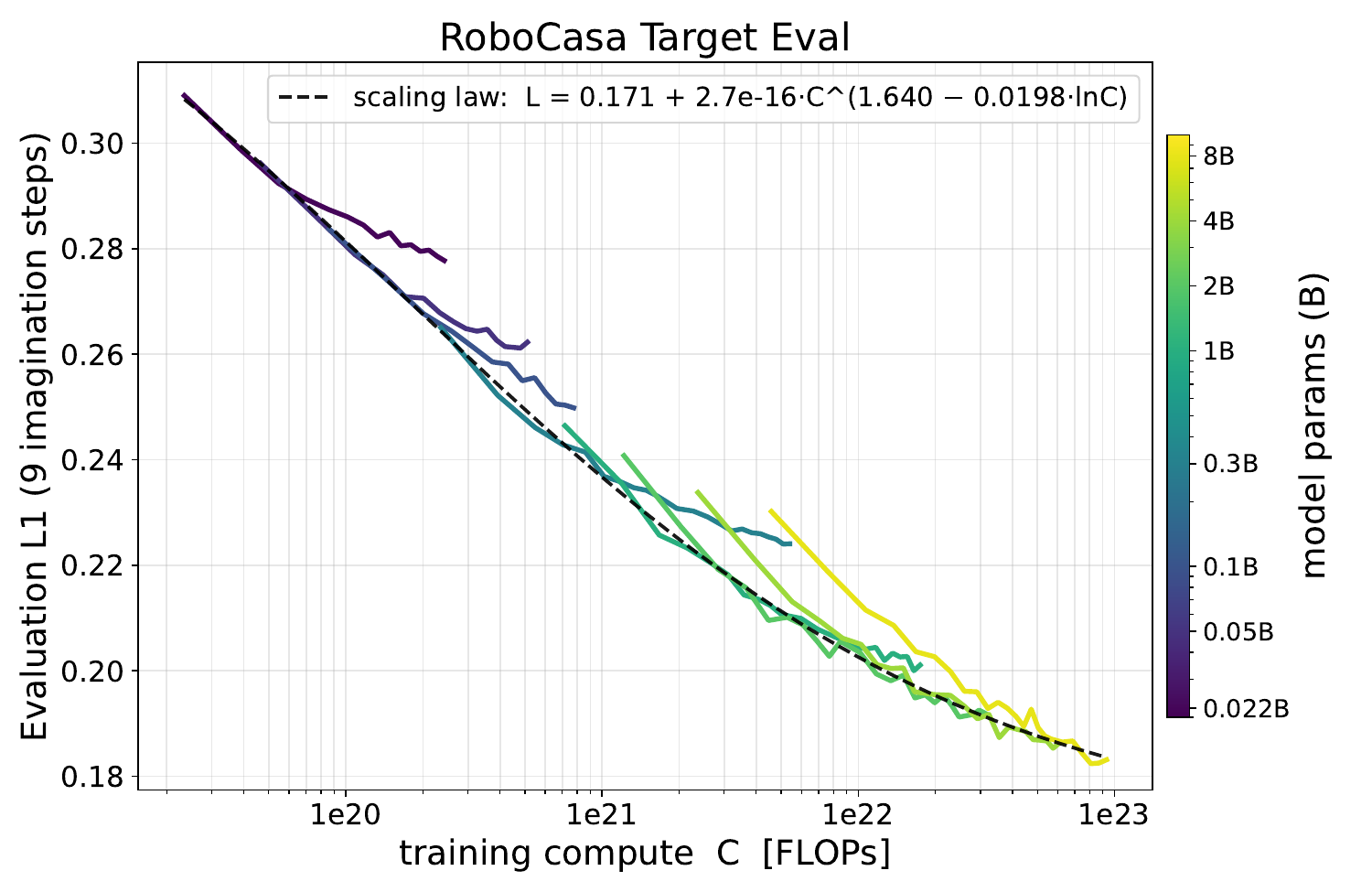}
\caption{RoboCasa target scenes (simulation).}
\label{fig:scaling-robocasa}
\end{subfigure}
\caption{\textbf{World-model scaling laws.} Out-of-distribution forward-prediction $L_1$ error versus
training compute for RoboJEPA models from 22M to 8B parameters (color), evaluated on a real-robot
DROID holdout and simulated RoboCasa scenes. The compute-optimal frontier (lower envelope)
follows the second-order power law $L(C){=}E{+}A\,C^{\alpha-\gamma\ln C}$ (dashed).
}
\label{fig:scaling-laws}
\end{figure}

\textbf{Inference and Planning.} {}
RoboJEPA performs planning by predicting the future in the latent space and searching over the action sequence that brings that future to the desired goal. {}
We employ receding-horizon CEM~\citep{rubinstein1997optimization, deboer2005tutorial} as our planning algorithm with image goals; importantly, we use only a single image goal that represents the final state of the task, and we do not use subgoals. {} {}
Notably, we run all planning with the same set of hyperparameters across embodiments and tasks.
This process requires running thousands of forward passes of the world model, which is prohibitively expensive if the full history is recomputed every step.
Therefore, we employ a rolling KV-cache that reuses the keys and values of past steps in self-attention.
This way, at every step the model only processes a single frame, action, and proprioceptive vector, while the past information is provided by the cache (see \Cref{appx:kv}).
Additionally, we employ PyTorch compilation to further speed up inference.
However, in the typical planning scenario we need to run compiled inference during planning on the real robot, and since first-time compilation is time-consuming, running planning with compilation can become prohibitively slow in practice.
This is why we build a simple inference engine that precompiles the inference computation and saves it to disk, and at planning time simply loads it back (see \Cref{appx:infer:aot}).
Finally, since CEM planning is heavily parallelizable, we connect the robot directly to the GPU cluster and run the planning on multiple GPUs.
This way, we make the planning fast enough to be practical: while it is not real-time, it takes only several seconds to perform a multi-step planning even for an $8$B model.

\subsection{Diffusion Decoder}\label{subsec:decoder}
To qualitatively understand the capabilities of our world models, we train a diffusion decoder that produces pixel outputs from the predicted features; its goal is to visualize the forward predictions of the world model.
The decoder is a diffusion model~\citep{ho2020ddpm, song2021ddim} that denoises the continuous video tokens of the Cosmos tokenizer~\citep{nvidia2025cosmos}\footnote{Cosmos-CV4x8x8.} spatially conditioned on the V-JEPA~2.1 features.
The purpose of using a generative model here is that the decoder does not merely convert the V-JEPA features into Cosmos tokens, but also imagines unseen motion.
During the training of the decoder, we compute V-JEPA features with a frame skip of $4$ frames, whereas the Cosmos tokens compress the same frames without any frame skip.
As a result, the Cosmos tokens contain more information than the original V-JEPA feature frames, and this extra information --- precisely, the $3$ RGB frames between consecutive frames encoded by the V-JEPA encoder --- must be generated.
With such a decoder, we can run forward prediction with the world model at, for example, $5$\,Hz actions, producing latent features at the same rate, and then use the decoder to expand them into an RGB stream at $20$\,FPS that covers exactly the same motion.
Full architecture, training, and sampling details are in \Cref{appx:decoder}.

\section{Experiments}\label{section:experiments}
\subsection{World Model Scaling Laws}\label{subsec:scaling_laws}

Our goal is to empirically find the concrete functional form of the scaling law that allows us to extrapolate predictions about the performance of the RoboJEPA model trained at larger scales. We consider four candidate functional forms. First, the standard power law $L(C)=E+A\,C^{\alpha}$~\citep{henighan2020scaling}, whose exponent $\alpha$ is constant. Second, a generalization whose exponent itself changes linearly with compute, $L(C)=E+A\,C^{\alpha-\gamma\ln C}$; we refer to the former as a \emph{first-order} and to the latter as a \emph{second-order} power law\footnote{It is easy to show that, for the latter, the dependency between $\ln(L-E)$ and $\ln C$ is a quadratic function of $\ln C$.}. Finally, the Broken~\citep{caballero2022bnsl} and Unified~\citep{caballero2026unsl} Neural Scaling Laws. Here, $C$ is the total training compute (in FLOPs) and $L(C)$ is the world model's forward-prediction error, i.e.\ the per-token $\ell_1$ distance between the predicted and the encoder features, averaged over a $9$-step latent rollout on held-out data.
Among these four forms, we find that RoboJEPA predictably follows the second-order power law.

\textbf{Setup.}
We evaluate each model trained at each compute budget by its forward-prediction $L_1$ error, $L(C)$, on held-out data. Importantly, this is the same per-token $L_1$ objective the model is trained on and that the planner minimizes at deployment (\Cref{eq:loss-tf}).
We report $L(C)$ on two embodiments, a DROID holdout dataset collected on a real robot and a simulated RoboCasa suite. Evaluation is always on scenes unseen during training, always uses two views: one side camera (left or right) and one wrist camera, and predicts nine future steps from an initial observation under the recorded action sequence, giving a ten-step sequence in total. Further evaluation details are provided in \Cref{appx:scaling-laws}.
For each model at each training duration we compute the total training compute $C$ exactly, by counting the FLOPs of every operation in one training step and multiplying by the number of steps taken across all stages. We plot the prediction error for every model and training duration against its total training compute. At each compute budget, we retain the lowest error across model sizes. These best-achievable errors form the compute-optimal frontier as a function of the training compute. Finally, we fit each candidate curve to the frontier points of the smaller models (22M--2B) and select as our scaling law the one that best extrapolates to the held-out 4B and 8B models.

\begin{table}[t!]
\caption{Variants of parametric curves as scaling-law candidates for the $L_1$ rollout error of RoboJEPA world models. Curve fits use models sized $22$M--$2$B parameters; we report held-out extrapolation error at 4B and 8B for DROID and RoboCasa. The BNSL and UNSL functional forms and their fitted parameter values are in \Cref{appx:scaling-laws}.}
\label{tab:scaling-extrapolation}
\centering
\resizebox{\textwidth}{!}{%
\begin{tabular}{lcccc}
\toprule
Parametric Curve & \multicolumn{2}{c}{Curve parameters} & \multicolumn{2}{c}{\shortstack{Extrapolation Error \\ (4B and 8B), $\times 10^{-3}$}} \\
\cmidrule(lr){2-3} \cmidrule(lr){4-5}
 & DROID & RoboCasa & DROID & RoboCasa \\
\midrule
$L(C) = E + A\,C^{\alpha}$
 & $E{=}0.182,\ A{=}108.8,\ \alpha{=}{-}0.151$
 & $E{=}0.075,\ A{=}23.2,\ \alpha{=}{-}0.103$
 & $2.0$ & $3.7$ \\
\midrule
$L(C) = E + A\,C^{\alpha - \gamma \ln C}$
 & \shortstack{$E{=}0.218,\ A{=}7.96{\times}10^{-19}$ \\ $\alpha{=}1.91,\ \gamma{=}0.02308$}
 & \shortstack{$E{=}0.1709,\ A{=}1.66{\times}10^{-17}$ \\ $\alpha{=}1.761,\ \gamma{=}0.02105$}
 & $\mathbf{0.6}$ & $\mathbf{1.4}$ \\
\midrule
BNSL~\citep{caballero2022bnsl}, $n{=}1$
 & \multicolumn{2}{c}{\Cref{appx:scaling-laws:bnsl}}
 & $1.0$ & $1.8$ \\
\midrule
UNSL~\citep{caballero2026unsl}, $n{=}1$
 & \multicolumn{2}{c}{\Cref{appx:scaling-laws:unsl}}
 & $1.4$ & $5.7$ \\
\bottomrule
\end{tabular}%
}
\end{table}

\textbf{Results.} \Cref{tab:scaling-extrapolation} compares the four candidate curves by how well a fit on the smaller models (22M--2B) extrapolates to the held-out 4B and 8B models. The second-order power law extrapolates best, with a held-out error of $0.6$ and $1.4\times10^{-3}$ on DROID and RoboCasa respectively, roughly $2$--$3\times$ lower than the standard power law ($2.0$ and $3.7\times10^{-3}$) and below both BNSL and UNSL. We therefore adopt it as our scaling law. \Cref{fig:scaling-laws} plots the forward-prediction error against training compute for all models. The lower envelope across model sizes forms the compute-optimal frontier, and no model size lies on the frontier entirely. In addition, \Cref{fig:scaling-laws} plots the scaling law computed on all models as the dashed line.
Interestingly, the two best candidates, the second-order law and BNSL, estimate nearly the same irreducible error, $\approx 0.2$ on DROID and $\approx 0.17$ on RoboCasa, indicating that RoboJEPA is close to saturation on both evaluations under its current training and data budget.

\subsection{Downstream Planning Scaling}\label{subsec:correlations}

\begin{figure}[t!]
\centering
\begin{subfigure}[t]{0.49\linewidth}
\centering
\includegraphics[width=\linewidth]{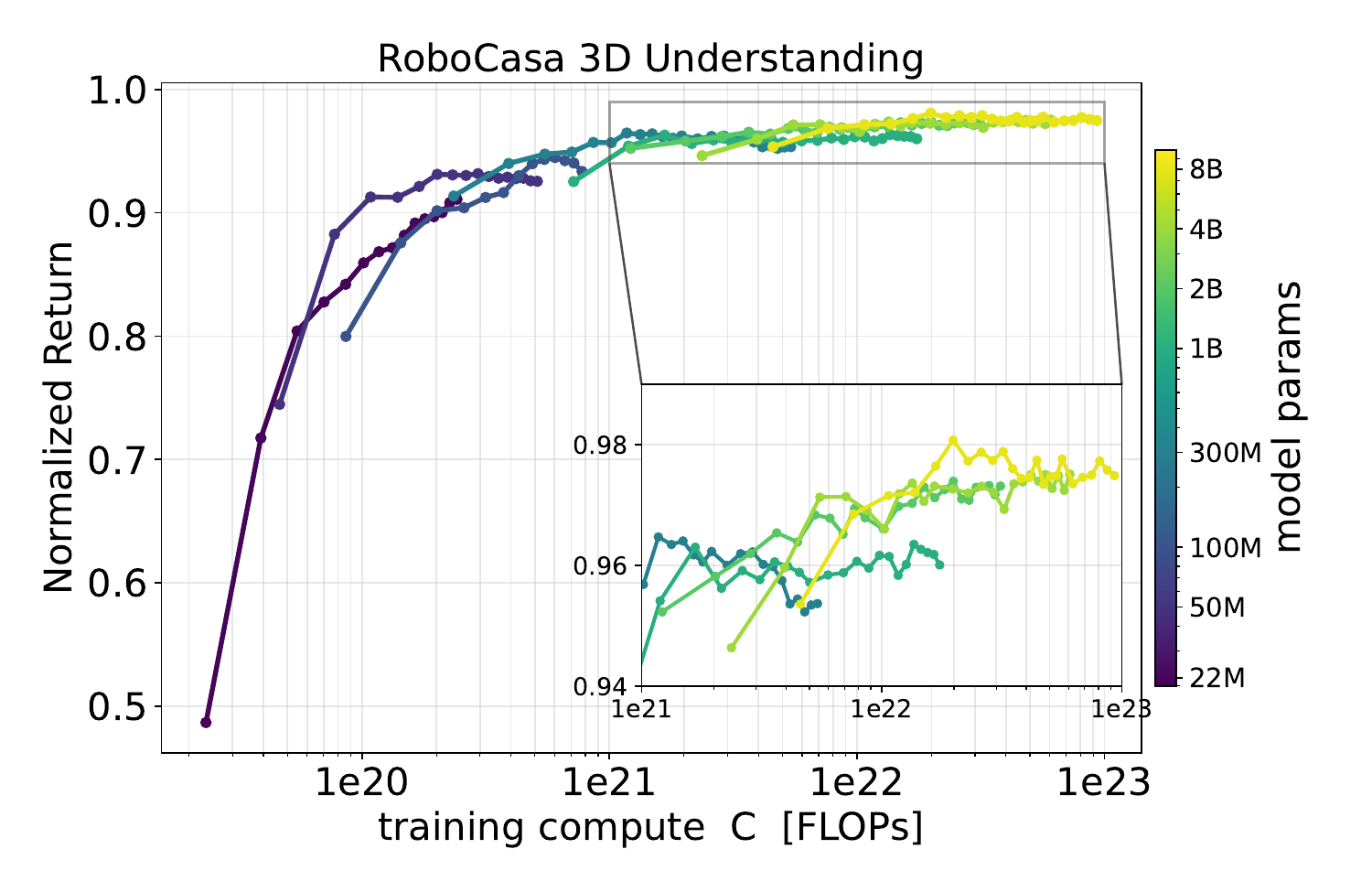}
\caption{The normalized return for the pose reach task in RoboCasa environment evaluated in unseen kitchens.}
\label{fig:reward-scaling-reach-a}
\end{subfigure}
\hfill
\begin{subfigure}[t]{0.49\linewidth}
\centering
\includegraphics[width=\linewidth]{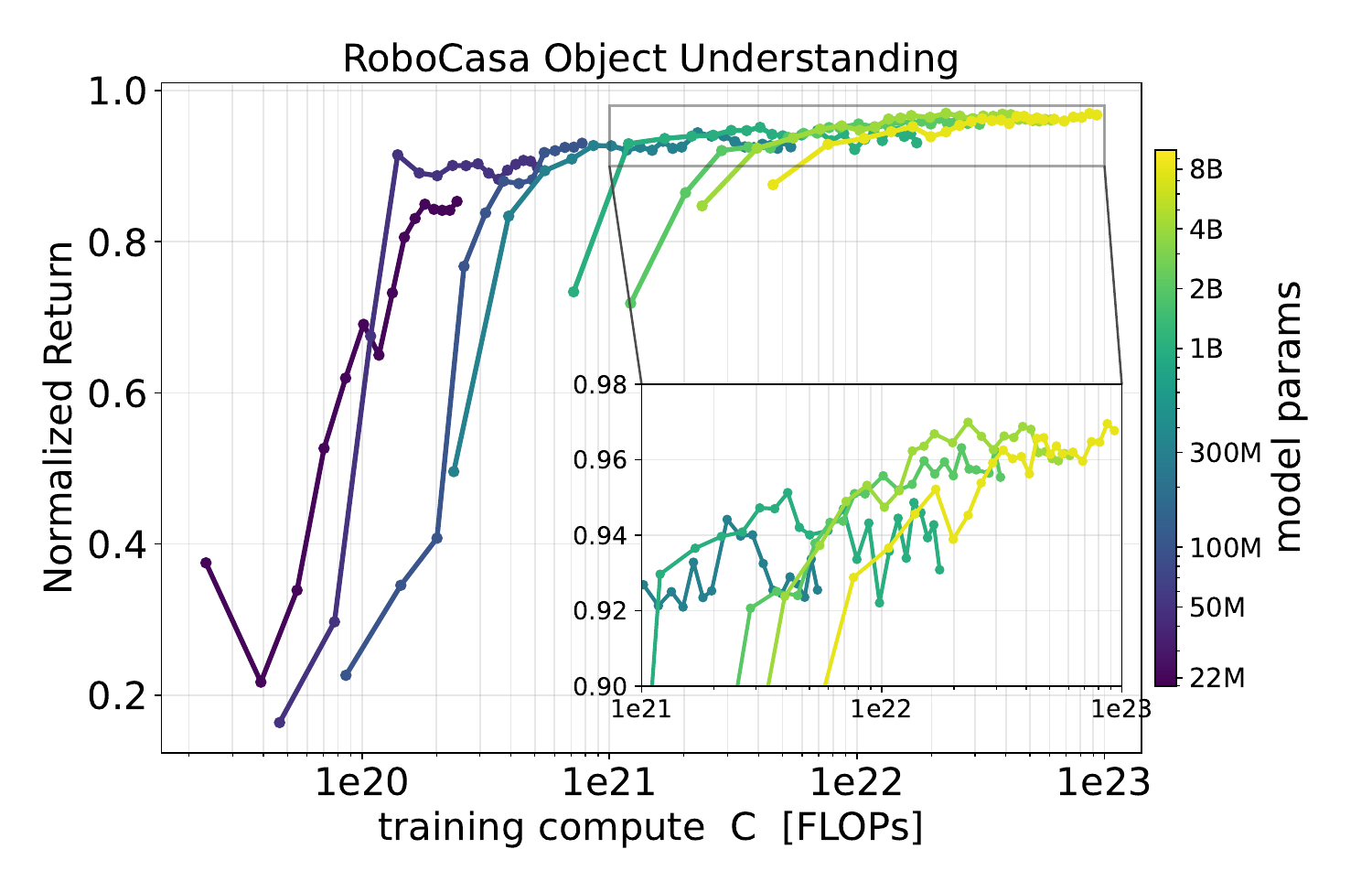}
\caption{The normalized return for the task of pose reach while holding an object.}
\label{fig:reward-scaling-reach-b}
\end{subfigure}
\caption{Downstream planning success scales with training compute. Median success rate on the reach task versus total training compute, following a power-law fit.}
\label{fig:reward-scaling-reach}
\end{figure}

Having identified an improvement trend in offline performance of world models, we hypothesize that this improvement translates to the world model's physical understanding of the new environment in which it operates. For example, is the action conditioning accurately mapped to the correct movements of the robot in the representation space? Are the grasping and object interaction actions correctly represented by the world model in response to incoming action? %
We test this hypothesis via online planning with image goals. \Cref{subsec:scaling_laws} introduced the predictable improvement in offline forward-prediction, and we verify that this improvement translates into a better success rate.

To test this hypothesis, we build a set of interactive environments on top of RoboCasa~365~\citep{nasiriany2026robocasa365}, each isolating one basic capability, and test whether more training compute improves it. In this section we focus on \emph{greedy} tasks, which can be solved by one-step planning where simply moving toward the goal at every step is enough. (i) \emph{3D understanding} (pose reaching): reach a target end-effector pose in as few steps as possible. (ii) \emph{Object understanding} (pose reaching while holding an object): reach a target pose while holding a grasped object, so the planner must additionally search over gripper actions to avoid dropping it. Harder tasks that require non-greedy, longer-horizon planning --- such as moving around an obstacle or pushing an object to a target --- are studied in \Cref{subsec:towards-better-scaling}. Full environment and reward details are in \Cref{appx:robotic-setup:robocasa}.

We use the same range of RoboJEPA checkpoints as in \Cref{subsec:scaling_laws} and run each of them in our RoboCasa environments with closed-loop planning~\citep{mayne2000constrained}, where the model re-plans toward the goal image after every executed action (full protocol in \Cref{appx:robotic-setup:robocasa}). \Cref{fig:reward-scaling-reach} shows the evaluation results, measured by the simulator's normalized cumulative reward on the performed trajectory (\Cref{appx:robotic-setup:robocasa}). Just like with the offline performance, we see a clear improvement trend of the downstream performance as we scale the training compute. For both tasks, models form a clear performance frontier, with larger models surpassing smaller ones as the training compute grows. For example, the simplest end-effector 3D movement capabilities emerge after around $10^{20}$ FLOPs while the basic capability of moving while holding an object emerges at around $3\times10^{20}$ FLOPs of training, both of which correspond to training $50$M--$100$M-parameter models on the full multi-embodiment dataset. Due to the simplicity of these tasks, performance saturates early. Therefore, for building robotic world models, we recommend first building downstream evaluations that reflect the desired capabilities, then estimating the minimal training compute required to achieve them.

\subsection{Better Scaling for Long Horizon Planning}\label{subsec:towards-better-scaling}

With the established transfer from offline performance to online greedy performance in \Cref{subsec:scaling_laws,subsec:correlations}, we now ask whether the same offline-to-online connection holds for \emph{non-greedy}, longer-horizon tasks that cannot be solved in one step. We design two such tasks in RoboCasa. The first is an \emph{obstacle-reach} task: the goal is again to reach a target end-effector position, but we place a large object (a kitchen appliance) as an obstacle so that a direct straight-line motion cannot reach the goal. The second is an \emph{object-push} task, where an object is placed on the table in front of the robot and the goal is to move it toward a target position, which requires the arm to descend to the object, push it, and ascend back. Full task definitions are in \Cref{appx:robotic-setup:robocasa}.

\begin{wrapfigure}{r}{0.5\textwidth}
\centering
\includegraphics[width=\linewidth]{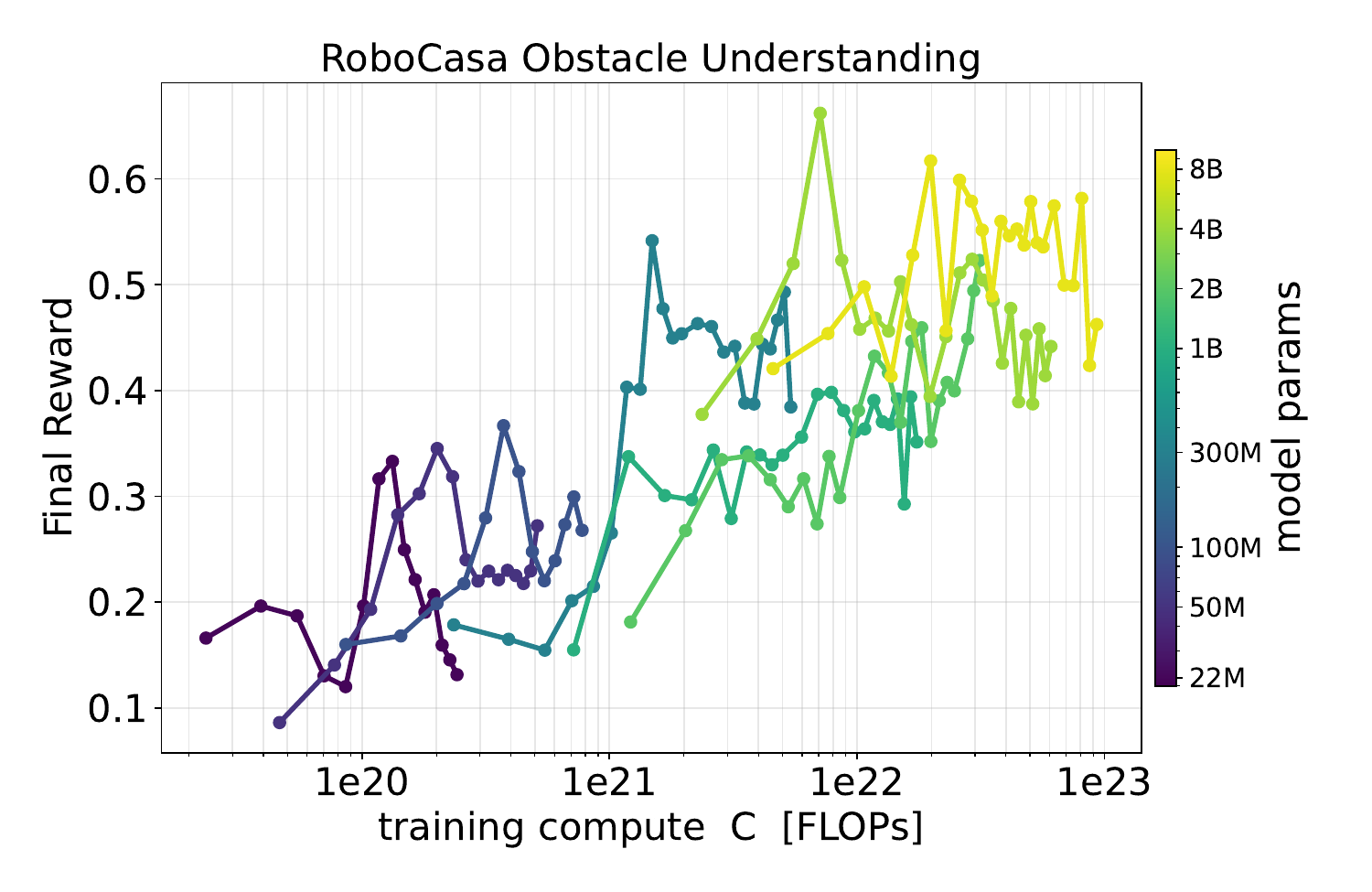}
\caption{\textbf{Short-horizon prediction training on obstacle-reach.} Final reward versus training compute. The reward correlates with compute, but no clean frontier emerges and smaller models often beat larger ones.}
\label{fig:obstacle-reach-reward-a}
\end{wrapfigure}

We first deploy the models trained with the recipe of \Cref{subsec:scaling_laws} on the obstacle-reach task. \Cref{fig:obstacle-reach-reward-a} shows that the final reward correlates with training compute, but the correlation is weak and no clean compute
frontier emerges: smaller models often beat larger ones (for example, the $300$M planner reaches a higher final reward than the $1$B one). Unlike the greedy tasks of \Cref{subsec:correlations}, predictable scaling does not emerge for longer-horizon tasks, which we attribute to a mismatch between training and deployment. Planning on these tasks relies on accurate \emph{long-horizon} prediction, yet the models were trained with only a short autoregressive rollout ($K=2$ steps). We therefore align training with how the model is used at test time and increase the training rollout to $K=10$ steps during the cooldown, together with higher-resolution inputs. We refer to the two cooldown recipes as \emph{short-horizon prediction training} ($K=2$, $240\times 320$) and \emph{long-horizon prediction training} ($K=10$, a mix of $240\times 320$ and $480\times 640$). Retraining every model from scratch this way would be prohibitively computationally expensive, so we run the long-horizon cooldown only from the flat-learning-rate pretraining checkpoints (\Cref{subsec:training}): the long flat-learning-rate stage only needs to learn the coarse dynamics cheaply, and the computationally expensive long-rollout, high-resolution signal is spent where it matters, in the final annealing phase.

\begin{figure}[b!]
\centering
\begin{subfigure}[t]{0.49\linewidth}
\centering
\includegraphics[width=\linewidth]{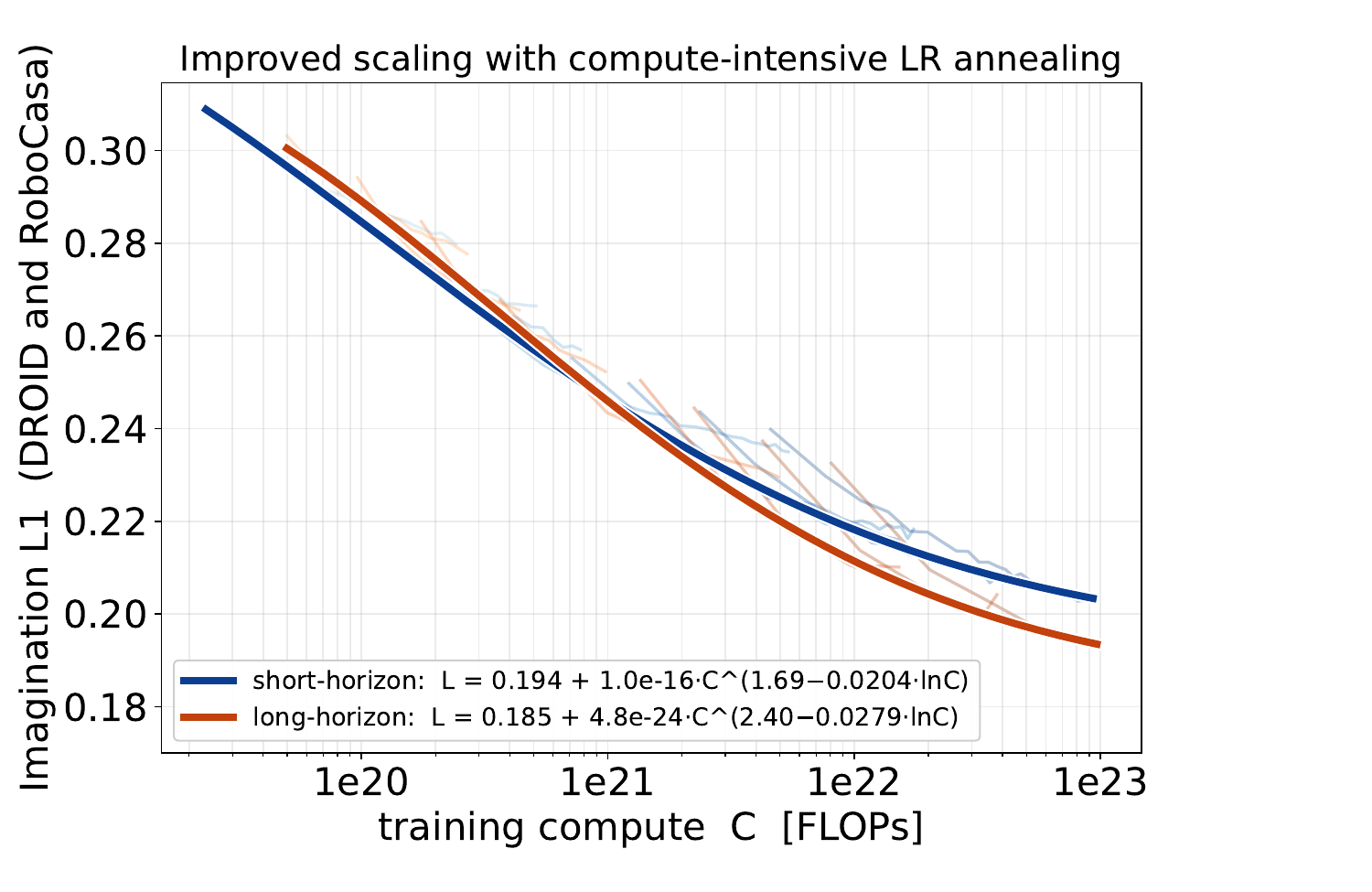}
\caption{Imagination error scaling for short- vs.\ long-horizon prediction training.}
\label{fig:high-low-compute-scaling}
\end{subfigure}
\hfill
\begin{subfigure}[t]{0.49\linewidth}
\centering
\includegraphics[width=\linewidth]{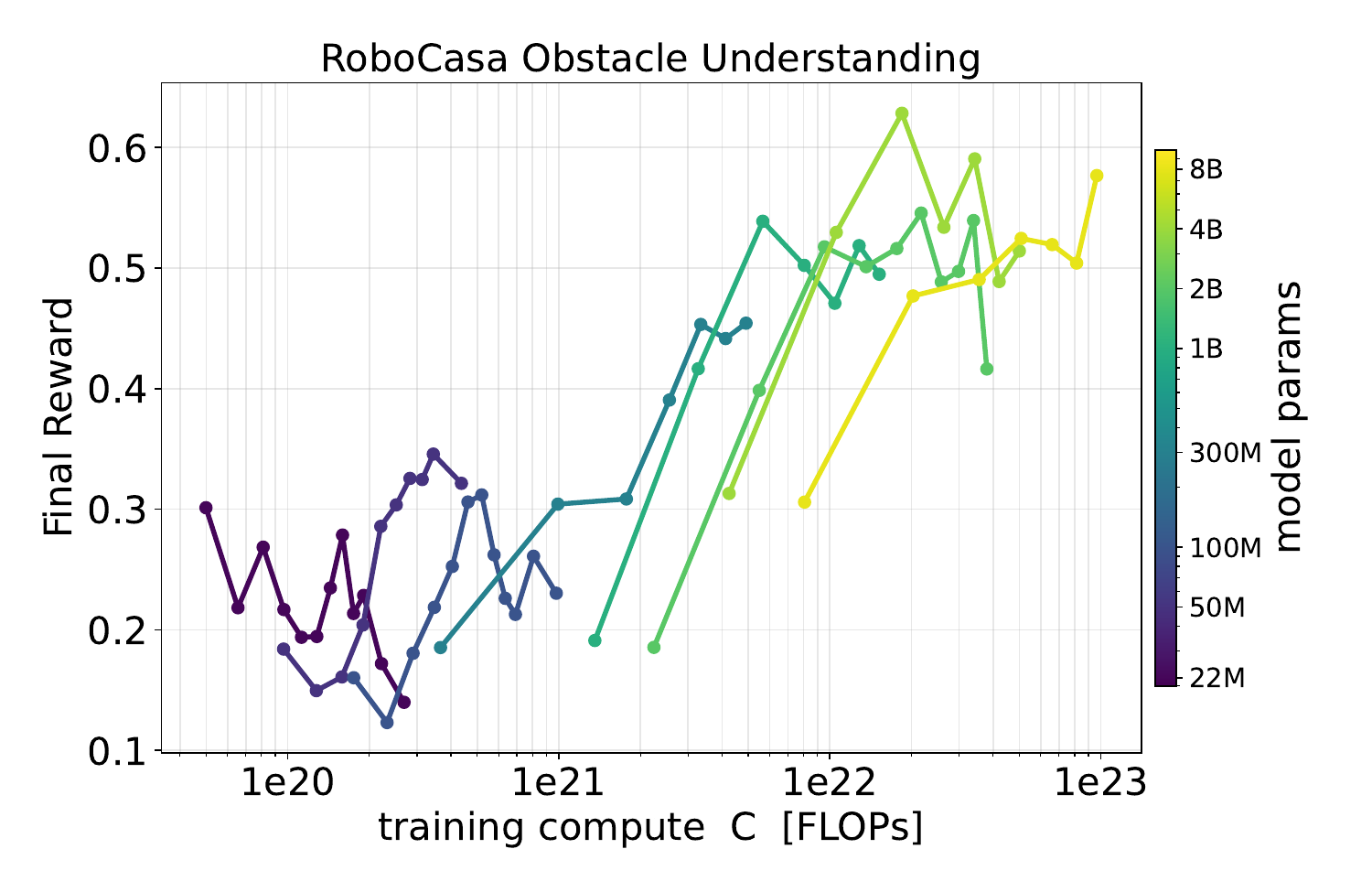}
\caption{Obstacle-reach final reward, long-horizon prediction training.}
\label{fig:obstacle-reach-reward-b}
\end{subfigure}
\caption{\textbf{Longer-horizon prediction training improves scaling and downstream long-horizon planning.} (a)~Second-order scaling laws of the imagination error on DROID and RoboCasa for the two cooldown recipes. (b)~Obstacle-reach final reward versus total training compute for the long-horizon ($K=10$) models.}
\label{fig:obstacle-reach-reward}
\end{figure}
\Cref{fig:high-low-compute-scaling} compares the two recipes by their scaling laws on DROID and RoboCasa. Long-horizon prediction training is more computationally expensive per step, but it becomes more compute-efficient beyond about $10^{21}$ FLOPs of total training compute (around $300$M--$1$B parameters at the compute-optimal frontier) and reaches a substantially lower forward-prediction error. It also lowers the \emph{irreducible} error of the fit, indicating a genuine reduction in error accumulation.

\begin{wrapfigure}{R}{0.5\textwidth}
\centering
\includegraphics[width=\linewidth]{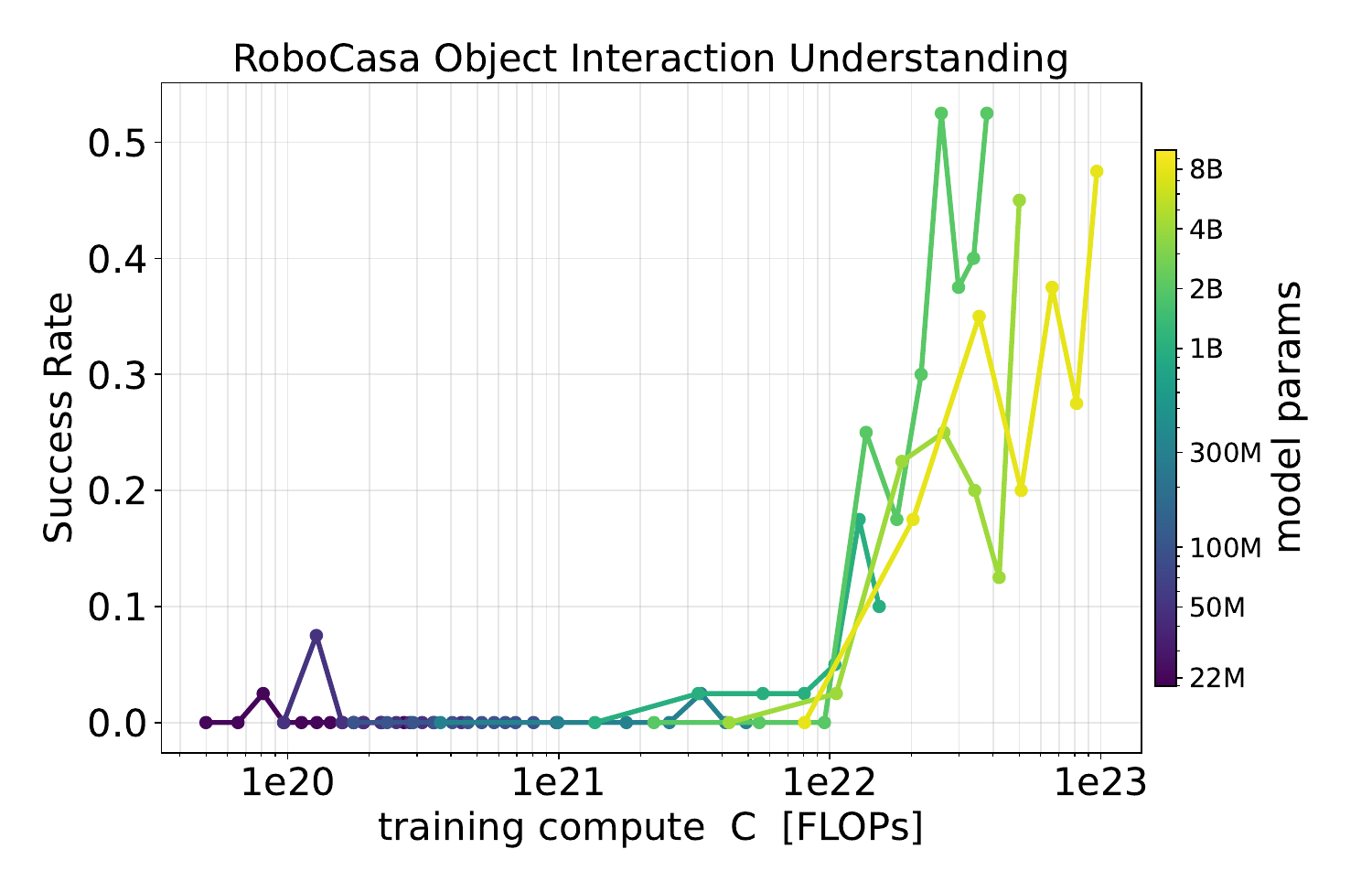}
\caption{Downstream success rate on the RoboCasa push task versus training compute, for the long-horizon ($K=10$) models.}
\label{fig:push-success}
\end{wrapfigure}
Deploying the non-greedy planning models on the obstacle-reach task (\Cref{fig:obstacle-reach-reward-b}) significantly improves the trend: the final reward now strongly follows the predictable scaling pattern. Small models (below $300$M parameters) cap at around $0.3$, which corresponds to the end-effector remaining stuck on the wrong side of the obstacle. The obstacle-avoidance capability then takes off around $10^{21}$ FLOPs and saturates around $10^{22}$ FLOPs. This is much later than the end-effector capabilities of \Cref{subsec:correlations}, which emerged around $10^{20}$--$3\times10^{20}$ FLOPs, indicating that modeling interaction with the static scene geometry requires considerably more training.

Finally, the object-push task assesses the object interaction understanding (through non-greedy planning) by the world model: because the robot's start and goal poses are identical (so the side and wrist views change only through the object's position), the model must understand that pushing the object is what changes the scene. \Cref{fig:push-success} shows the success rate, defined simply as whether the object moved toward its target position (\Cref{appx:robotic-setup:robocasa}). Interestingly, no model achieves a positive success rate until around $10^{22}$ FLOPs, after which all models begin to succeed, suggesting that fine-grained understanding of surrounding objects emerges after certain training computation regardless of the model size. In summary, these results align in a natural order in which capabilities appear with more compute: first the 3D control of the end-effector, then manipulating a grasped object, then the geometry of the static scene, and finally the dynamics of objects in the scene.

\subsection{Real World Robotic Capabilities}

\begin{wrapfigure}{R}{0.5\textwidth}
\centering
\includegraphics[width=\linewidth]{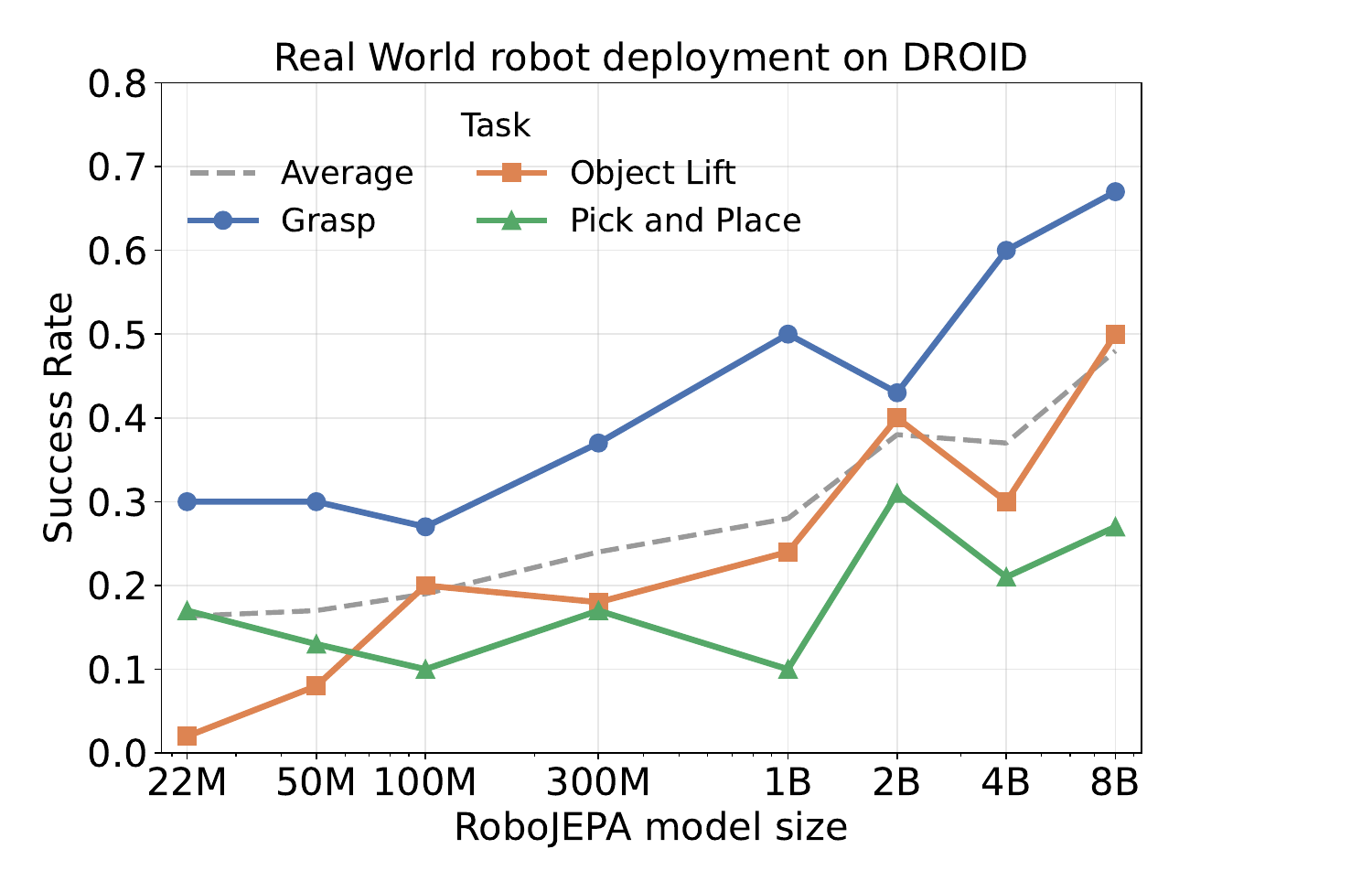}
\caption{\textbf{Real-world deployment on DROID.} Success rates across model sizes for three manipulation tasks; the dashed line shows their average.}
\label{fig:real-world-success-main}
\end{wrapfigure}
Our ultimate goal is to showcase the capabilities of RoboJEPA in real-world settings.
As a step towards this goal, we deploy RoboJEPA models ranging from 22M to 8B parameters
on the DROID~\citep{khazatsky2024droid} platform with a single-arm Franka robot and
two cameras: an external camera mounted to the left of the arm and a wrist camera
using the standard DROID mount. To deploy RoboJEPA via planning, we record an image
goal on the robot by executing a predefined robot motion and manually placing the
object in the grasping position so the robot can interact with it.
We provide only a single final image goal,
requiring longer-horizon, non-greedy planning. We evaluate three tasks: Grasp, Object
Lift, and Pick and Place. We use the same planning hyperparameters across tasks and model sizes,
evaluating the highest-compute checkpoint at each size. Setup, task definitions,
episode counts, scoring criteria, and planning hyperparameters are detailed in
\Crefrange{appx:robotic-setup:franka}{appx:robotic-setup:franka-planning}.

To contextualize the RoboJEPA results, we also evaluate the Vision-Language-Action
(VLA) models $\pi_0$-FAST~\citep{pertsch2025fast} and $\pi_{0.5}$~\citep{black2025pi05}.
However, their deployment differs in goal specification (text instructions versus
a single goal image) and training (DROID fine-tuning for the VLA baselines versus
pretraining on a mixture that includes DROID for RoboJEPA).
Neither RoboJEPA nor the VLA baselines receive task-specific fine-tuning.
These differences make the VLA results contextual references rather than directly
comparable baselines.
\Cref{fig:real-world-success-main} shows the scaling trend.
For each task, all RoboJEPA models, including very small models, demonstrate non-zero
success rates. Overall, success rates show a clear trend of improving with model
size and generally decreasing with task complexity. This picture does not hold for
the reference models, which work best on Pick and Place, the hardest task, but degrade
significantly on the others. For example, $\pi_{0.5}$ achieves a $53\%$ success rate
on Pick and Place but $0\%$ on Object Lift. We hypothesize that this difference may
be an artifact of training, as verbs such as ``lift'' are heavily underrepresented
in the DROID dataset, which focuses mostly on pick-and-place tasks.
Due to the planning nature of the agent and the task specification as an image goal, RoboJEPA
models are not susceptible to this type of degradation.
Qualitatively, behavior transitioned gradually from smooth but less adaptive
execution in smaller models to more reliable recovery from mistakes in larger
models (\Cref{appx:robotic-setup:real-world-discussion}).
\Cref{tab:real-world-capabilities-main} reports task progress and success rates for
the largest RoboJEPA models and VLA references.

\begingroup
\begin{table}[H]
\captionsetup{singlelinecheck=false}
\caption{\textbf{Real-robot task progress and success rates (\%).} Full results are in \Cref{tab:real-world-capabilities}.}
\label{tab:real-world-capabilities-main}
\centering
\small
\setlength{\tabcolsep}{4pt}
\begin{tabular*}{\linewidth}{@{\extracolsep{\fill}}ll*{6}{c}@{}}
\toprule
\multirow{2}{*}{Model} & \multirow{2}{*}{Goal} & \multicolumn{2}{c}{Grasp} & \multicolumn{2}{c}{Object Lift} & \multicolumn{2}{c}{Pick and Place} \\
\cmidrule(lr){3-4}\cmidrule(lr){5-6}\cmidrule(lr){7-8}
 & & Progress & Success & Progress & Success & Progress & Success \\
\midrule
$\pi_0$-FAST & \multirow{2}{*}{Text} & 35 & 22 & 20 & 12 & 55 & 40 \\
$\pi_{0.5}$ & & 25 & 5 & 12 & 0 & 69 & 53 \\
\midrule
RoboJEPA-4B & \multirow{2}{*}{Image} & 60 & 60 & 54 & 30 & 49 & 21 \\
RoboJEPA-8B & & 67 & 67 & 65 & 50 & 42 & 27 \\
\bottomrule
\end{tabular*}
\end{table}
\endgroup

\subsection{Qualitative Analysis}\label{subsec:qualitative-analysis}

\begin{figure}[p]
\centering 
\includegraphics[width=\textwidth, height=0.92\textheight, keepaspectratio]{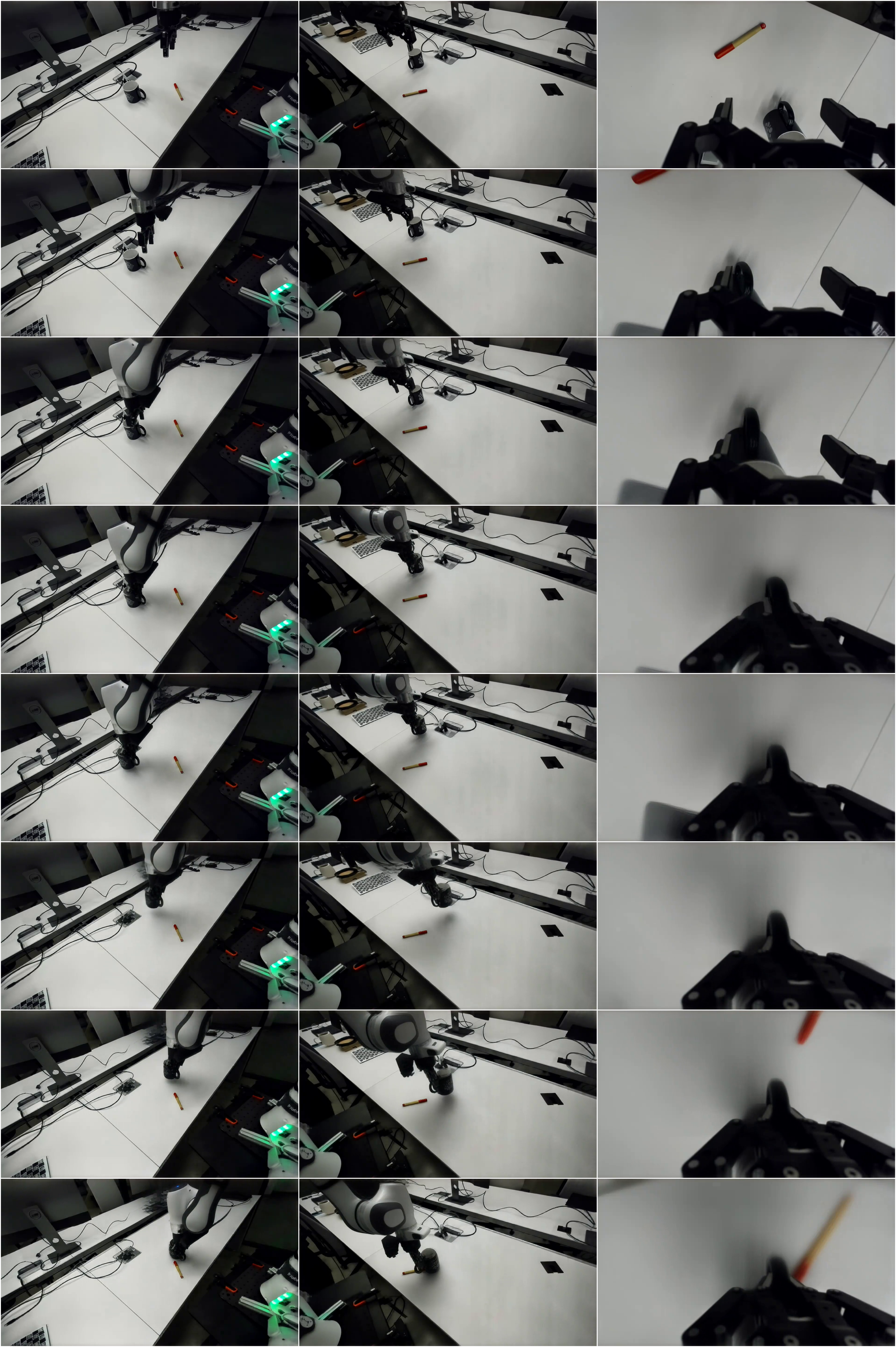} 
\caption{High Resolution 720p World Model Imagination Rollout.} 
\label{fig:demo_720_cup} 
\end{figure}

So far we have quantified the world model through its prediction loss and downstream planning success; we now inspect how the prediction improves qualitatively by decoding its latent rollouts back to pixels. To do that, we use the diffusion decoder introduced in \Cref{subsec:decoder}. The decoder serves purely as a visualization tool and is trained only on the frozen V-JEPA~2.1-G encoder features (see \Cref{appx:decoder}). Note that all visualizations use the same encoder and decoder and vary only the RoboJEPA world model size. We perform forward prediction given a single frame (or a pair of frames) and a sequence of robot actions, and each sequence of predicted embeddings is decoded back to pixels. We run the prediction across a range of settings --- low- and high-resolution, single- and dual-view --- and report them separately (\Cref{fig:imagination_scaling,fig:imagination_scaling_2view}; full set in \Cref{appx:demonstration_figures}); the decoder itself is trained at a single $240{\times}320$ resolution. These visualizations show that the forward-prediction improves with model scale in several concrete ways: larger models track the true dynamics more accurately over the rollout, preserve object identity and fine manipulation detail, and keep multiple camera views mutually consistent. Several behaviors are effectively \emph{emergent} with scale --- gripper--object contact and structured deformation such as folding appear only once the predictor is large enough, whereas smaller models blur these interactions away. Finally, we finetune the $8$B version of RoboJEPA on DROID to imagine all three camera views simultaneously at high ($720$p) resolution (\Cref{appx:training:droid-720p}; \Cref{fig:demo_720_cup}). These rollouts demonstrate that RoboJEPA performs accurate object manipulation even at very high resolution, and they exhibit strong cross-view information transfer, with content from the side (exterior) cameras propagating into the wrist-camera imagination.

\section{Related Work}\label{section:related}

\textbf{Joint-Embedding Predictive Architectures.}
Joint-Embedding Predictive Architectures (JEPAs) were proposed as a path
towards self-supervised models that learn abstract representations of the
world by predicting in a latent space rather than at the pixel level
\citep{lecun2022path}. I-JEPA \citep{assran2023ijepa} instantiated this
recipe for images by predicting the representations of masked image regions
from visible context, avoiding pixel-level reconstruction and the resulting
collapse to low-level texture. V-JEPA extended the framework to video by
predicting masked spatio-temporal feature regions and showed that the
resulting features capture motion and temporal structure
\citep{bardes2024vjepa}. \citet{assran2025vjepa2} scaled feature
prediction to over a million hours of internet video and, crucially for
our setting, post-trained an action-conditioned predictor (V-JEPA~2-AC)
on a small amount of unlabeled robot interaction data, enabling zero-shot
Franka manipulation by latent planning against image goals.
Our work, \emph{RoboJEPA}, is similar to V-JEPA~2-AC in terms of the approach. We keep the V-JEPA encoder frozen and focus on
the action-conditioned predictor as a \emph{world model}. However, compared to V-JEPA 2 AC, we replace the
small post-training corpus with a unified, large-scale mixture of robot
demonstrations, therefore, the methods are not apples-to-apples comparable. To our knowledge, we provide the first systematic recipe
for scaling V-JEPA-style action-conditioned predictors on large-scale,
multi-embodiment robot data, and
we characterize how predictor performance scales with parameters and
demonstration tokens.

\textbf{World Models for Robotics.}
World models compress experience into a learned simulator of the
environment that an agent can train and plan in
\citep{ha2018worldmodels}. The Dreamer family pioneered this paradigm at
scale: PlaNet learned latent dynamics for planning from pixels
\citep{hafner2019planet}, Dreamer used latent imagination to learn
behaviors \citep{hafner2020dreamer}, DreamerV2 introduced discrete latents
to master Atari \citep{hafner2021dreamerv2}, and DreamerV3 demonstrated a
single set of hyperparameters that succeeds across a wide range of domains
including 3D control \citep{hafner2023dreamerv3}. DayDreamer applied
Dreamer directly to physical robots and learned locomotion and manipulation
in the real world \citep{wu2023daydreamer}, while Masked World Models
showed that decoupling visual representation learning from dynamics
modeling improves sample efficiency \citep{seo2023mwm}. More recent work
has scaled video-based world models with transformer and diffusion
backbones: IRIS \citep{micheli2023iris} and iVideoGPT
\citep{wu2024ivideogpt} treat world modeling as autoregressive token
prediction; UniSim \citep{yang2024unisim}, Genie \citep{bruce2024genie},
and GAIA-1 \citep{hu2023gaia1} train large generative simulators of
interactive scenes; and the Cosmos platform releases physical-AI world
foundation models at industrial scale \citep{nvidia2025cosmos}. In contrast
to these pixel-space generators, RoboJEPA predicts in the
\emph{representation space} of a frozen V-JEPA encoder, which is what
allows us to scale the predictor to billions of parameters on robot data
without paying the computational cost of pixel-level reconstruction at every
step.

\textbf{Planning with Learned World Models.}
A central use of a world model is test-time planning. Classical model-based
control methods such as PETS \citep{chua2018pets}, model-based RL with
neural network dynamics \citep{nagabandi2018neural}, and visual foresight
\citep{finn2017deep} optimize action sequences against a learned dynamics
model. Sampling-based shooting methods, in particular the Cross-Entropy
Method \citep{rubinstein1997optimization,deboer2005tutorial} and Model
Predictive Path Integral control \citep{williams2017mppi}, remain the
workhorses for high-dimensional action search. TD-MPC and TD-MPC2 combine
latent dynamics, value learning, and CEM-style trajectory optimization for
robust continuous control \citep{hansen2022tdmpc, hansen2024tdmpc2}.
DINO-WM \citep{zhou2025dinowm} learns task-agnostic dynamics in DINOv2
feature space from offline trajectories using an approximately
$20$M-parameter predictor, and uses CEM to plan toward image goals without
additional task-specific training.
RoboJEPA follows this line: at deployment, we use CEM in the latent space
of the world model to score action sequences by how close their predicted
features land to a goal image's features, and execute the first action of
the best plan in receding-horizon fashion.

Closely related to our work, \citet{terver2026planning} investigate how
architectural, training, and planning choices affect latent world-model
performance, including model and data scaling, albeit at smaller scales.
Their proposed DROID predictor has approximately $229$M parameters and is
trained on an $8{,}000$-trajectory dataset.
We extend this investigation both by scaling to substantially larger world
models (up to $8$B parameters) and datasets (a mixture of approximately $2.87$M trajectories
across $12$ robot platforms) and by explicitly studying their scaling
behavior through scaling laws.

\textbf{Robotic Foundation Models from Demonstrations.}
A complementary line of work scales \emph{policies} directly on robot
demonstrations rather than scaling world models. RT-1
\citep{brohan2023rt1} and RT-2 \citep{brohan2023rt2} train large
transformer- and vision-language-action policies on broad manipulation
data. \citet{oxe2023} aggregate demonstrations across many embodiments
to enable cross-embodiment generalist policies, and a wave of recent
generalist models such as Octo \citep{octo2024}, OpenVLA
\citep{kim2024openvla}, and $\pi_0$ \citep{black2024pi0} push this
behavior-cloning recipe further. RoboJEPA shares the same data substrate
(in particular Open X-Embodiment, DROID \citep{khazatsky2024droid},
RoboMIND \citep{wu2024robomind}, BridgeData V2 \citep{walke2023bridgev2},
LeRobot \citep{cadene2024lerobot}, and AgiBot World \citep{bu2025agibot}),
but trains a goal-agnostic, action-conditioned \emph{predictive} model
instead of a task-conditioned policy. The same world model can then be
used to evaluate or improve any policy, and to plan zero-shot for new
goals at test time.

\textbf{Scaling Laws.}
Power-law scaling of model loss with parameters, data, and compute has
been documented extensively for language models
\citep{kaplan2020scaling}, generative modeling more broadly
\citep{henighan2020scaling}, and vision transformers
\citep{zhai2022scalingvit}, with compute-optimal allocations of model and
data established by Chinchilla \citep{hoffmann2022chinchilla}. In
robotics, however, scaling laws are largely an open question because
high-quality interaction data is scarce. We provide, to our knowledge, the
first systematic scaling study for action-conditioned world models trained
on large-scale, multi-embodiment real robot demonstrations, characterizing how forward-prediction loss
scales with predictor parameters and demonstration tokens, and showing
that this scaling behavior transfers to downstream planning performance.

\textbf{Scaling world models.}
A growing body of work reports that world models improve with scale. In
autonomous driving, the GAIA line of latent-diffusion world models has been
scaled to billions of parameters for controllable simulation and offline
evaluation \citep{russell2025gaia2, wayve2025gaia3, wayve2026gaia4},
though technical details beyond GAIA-2 are not publicly available.
\citet{pearce2024scalingagents} fit power laws for pre-trained agents and
world models and, in their appendix, report scaling laws for a robotic
world model, but their study is otherwise centered on game and simulated
environments rather than large-scale real-world multi-embodiment robot data.
\citet{sato2023scalingwm} study how world-model quality varies with model
size; IRASim \citep{zhu2025irasim} reports empirical scaling trends for a
fine-grained video world model for manipulation; PointWorld
\citep{huang2026pointworld} shows that a 3D point-flow world model improves
with more data and parameters; and DINO-world \citep{baldassarre2025dinoworld}
scales video prediction in the feature space of a frozen DINOv2 encoder
using predictors with up to $1.1$B parameters. Closest in
spirit, $\tau_0$-VLA \citep{cai2026tau0vla} scales \emph{test-time}
computation of a hierarchical policy using a world model to score candidate
subtasks, and Dyna-2 \citep{dyna2026dyna2} scales a \emph{world-action}
model over a million hours of human and robot video and measures
\emph{action}-prediction quality. RoboJEPA differs from these along two
axes: we scale \emph{training} compute of a pure world model and measure
its \emph{future-embedding} (imagination) accuracy, and we fit an explicit
scaling law rather than an empirical trend. This is a very important distinction, as an
empirical scaling trend only establishes that a metric improves as one adds
parameters, data, or compute, but says nothing about \emph{how much} it
will improve, whether returns have begun to saturate, or whether the next
unit of budget is better spent on model size or on data. A scaling law, by
contrast, is \emph{predictive}: it extrapolates performance well beyond the
scale at which it was fit, turning these decisions into quantitative
forecasts. Establishing such predictive laws for action-conditioned robotic
world models---and connecting them to downstream planning---is, to our
knowledge, what sets this work apart.

\section{Conclusion}\label{section:conclusion}

We introduced \textbf{RoboJEPA}, a family of action-conditioned latent world
models trained on a large-scale, multi-embodiment robotic data collection in the
latent space of V-JEPA~2.1. RoboJEPA predicts the next embedding in a single
forward pass, and we train it to predict the future conditioned on past
embeddings and future actions. We presented a recipe for scaling the predictor
that works from 22M to 8B parameters, the latter being the largest JEPA predictor model
trained to date. We then established scaling laws that predict the world-modeling
error as we scale training compute. These take the form of a second-order power
law $L(C)=E+A\,C^{\alpha-\gamma\ln C}$. We established these laws for both a real-world
(DROID) and a simulated (RoboCasa) embodiment. Finally, we demonstrated that
offline prediction scaling laws correlate with the online downstream performance
of world models, evaluated via planning toward a single image goal, which makes
the offline imagination error a reliable proxy for real-robot evaluation and
spares much of its computational cost. We also found that distinct manipulation capabilities
emerge at their own compute thresholds, in a natural order from end-effector
control, to manipulating a grasped object, to reasoning about static scene
geometry, and finally to the dynamics of objects in the scene. On a real robot,
we showed that the policy induced by world-model planning improves in success
rate and in qualitative behavior as the world model scales. We also built a
diffusion decoder to visualize the predicted action-conditioned future in pixel
space, and used it to show that qualitatively better robotic predictions emerge
with scale. We release all model checkpoints together with our training and
robot-deployment code.

Our study also has limitations that mark natural next steps. We scaled the
predictor while keeping the V-JEPA~2.1 encoder frozen, so our laws describe a
dynamics model on top of a fixed representation rather than the encoder and
predictor jointly. Because we train for multiple epochs over a fixed corpus, our
fits already sit close to data saturation, so pushing the frontier further will
likely require more and more diverse interaction data rather than parameters
alone. RoboJEPA also makes no use of text, planning instead toward a single image
goal, and it makes no use of a learned policy proposal during planning, sampling
actions from a uniform distribution and prioritizing them with the world model.

\clearpage
\newpage
\bibliographystyle{assets/plainnat}
\bibliography{paper}

\clearpage
\newpage
\beginappendix

\startcontents[appendix]
\printcontents[appendix]{}{1}{\setcounter{tocdepth}{2}}

\section{Model Architecture: Additional Details}\label{appx:arch}

\emph{Expands \Cref{subsec:model_architecture}.} We collect here the implementation-level details of the predictor block: the channel-pair derivation behind factorised RoPE (\Cref{appx:arch-rope}); the per-block view bias under grouped-query attention (\Cref{appx:arch-rope-view}); the per-embodiment input projection (\Cref{appx:arch-tokenization}); the embodiment-id breakdown that yields $|\mathcal{E}|{=}26$ (\Cref{appx:arch-emb}); and the predictor scaling sweep across all eight scales used in the scaling law of \Cref{section:experiments} (\Cref{appx:arch-scaling}). All notation is shared with \Cref{subsec:model_architecture}; in particular, $\ell$ is reserved for the predictor layer index and $\tau$ for autoregressive rollout steps.

\Cref{tab:appx-arch-vjepa2ac} summarises how the RoboJEPA predictor differs from the action-conditioned V-JEPA~2 predictor (V-JEPA~2-AC)~\citep{assran2025vjepa2} it builds on. The two share the same backbone family and the same next-frame prediction parameterisation; RoboJEPA changes the normalisation, position encoding, and the conditioning that supports cross-view, cross-embodiment world modeling.

\begin{table}[H]
\caption{Architectural comparison between the action-conditioned V-JEPA~2 predictor (V-JEPA~2-AC)~\citep{assran2025vjepa2} and the RoboJEPA predictor. Both share the V-JEPA~2 backbone and the next-frame latent-prediction parameterisation; RoboJEPA adds RMSNorm and QK-norm for training stability at scale, a fourth (view) RoPE axis, and explicit per-embodiment and per-view conditioning for cross-embodiment, multi-view world modeling.}
\label{tab:appx-arch-vjepa2ac}
\centering
\small
\setlength{\tabcolsep}{6pt}
\begin{tabularx}{\linewidth}{>{\raggedright\arraybackslash}p{0.30\linewidth} >{\raggedright\arraybackslash}X >{\raggedright\arraybackslash}X}
\toprule
 & V-JEPA~2-AC~\citep{assran2025vjepa2} & RoboJEPA (ours) \\
\bottomrule
\multicolumn{3}{c}{\textbf{Architecture}} \\
\midrule
Base ViT architecture       & \multicolumn{2}{c}{V-JEPA~2} \\
Training parameterisation    & \multicolumn{2}{c}{\begin{minipage}{0.60\linewidth}\centering Feed-forward next-embedding (frame) prediction with frame-causal mask; trained with an $\ell_1$ latent loss.\end{minipage}} \\
\midrule
Normalisation               & LayerNorm & RMSNorm \\
\midrule
QK-norm                     & No        & Yes \\
\midrule
Position embedding          & 3-D spatio-temporal RoPE & 4-D RoPE (space--time $+$ views) \\
\midrule
Multi-view world modeling   & No, single view        & Yes, single or multi-view without adaptation \\
\midrule
Multi-embodiment conditioning & None    & Learnable per-embodiment bias \\
\midrule
Multi-view conditioning     & None      & Post-RoPE per-block per-view bias \\
\midrule
Attention variant     & ViT self-attention      & GQA \\
\midrule
Action conditioning & Linear action embedding as token & per-embodiment multi-linear action embedding as token \\ 
\midrule
Attention head dim & $64$ & $128$ \\
\bottomrule
\multicolumn{3}{c}{\textbf{Training}} \\
\midrule
Training loss & Teacher-forced L1 + One AR step rollout L1 loss & Teacher-forced L1 + $K$-step AR rollout loss with KV-cache ($K=2$ during pretrain and $K=10$ in cooldown) \\
\midrule
Learning Rate Schedule & Linear warmup $\rightarrow$ cosine LR annealing & Linear warmup $\rightarrow$ flat LR $\rightarrow$ LR cooldown with more per-train-step compute \\
\midrule
Number of context steps for AR rollout loss & 1 & Random \\
\midrule
LR warmup steps & $4500$ & $2000$ \\
Peak LR & $4.25 \times 10^{-4}$ & $5\times 10^{-5}$ \\
Weight Decay & $0.04$ & $0.1$ \\
AdamW $\beta_1$ & $0.9$ & $0.9$ \\
AdamW $\beta_2$ & $0.999$ & $0.95$ \\
Gradient clipping & None & $1.0$, by global $l2$ norm \\
\midrule
Training resolution & $256 \times 256$ & $240 \times 320$ (in pretrain) and $240\times 320$ + $480 \times 640$ (in cooldown) \\
\bottomrule
\multicolumn{3}{c}{\textbf{Data}} \\
\midrule
Training dataset & DROID & RoboJEPA train mixture \\
Training episodes & $\leqslant 73000$ & $2{,}871{,}300$ \\
Training hours & approximately $26$ & $15{,}022$ video; $6{,}692$ action \\
Training cameras & left only & any (external and wrist) \\
Training framerate & $4$ FPS & varies per dataset\\

\end{tabularx}
\end{table}

\subsection{Factorised multi-axis RoPE}\label{appx:arch-rope}

Each per-head channel space $\{0,\dots,d_h-1\}$ partitions into four
contiguous, equal-size sub-blocks of width $d_\mathrm{ax}{=}d_h/4$,
one per spatio-temporal axis. Fix the enumeration
$\mathcal{A}_{\max}\!=\!\{t,i,j,v\}$ for the time, height, width, and
view axes, and let $j_\alpha$ denote the starting channel of axis $\alpha$, so
\begin{equation}
\begin{gathered}
\{0,\dots,d_h-1\}
\;=\;
\bigsqcup_{\alpha\in\mathcal{A}_{\max}}
\big\{o_\alpha,\,o_\alpha+1,\,\dots,\,o_\alpha+d_\mathrm{ax}-1\big\},
\\
o_t = 0,\; o_i = d_\mathrm{ax},\; o_j = 2 d_\mathrm{ax},\; o_v = 3 d_\mathrm{ax}.
\end{gathered}
\label{eq:rope-axis-partition}
\end{equation}
Within axis $\alpha$, the sub-block is partitioned in turn into
$d_\mathrm{ax}/2$ \emph{channel pairs}
$\big(o_\alpha + 2m,\, o_\alpha + 2m+1\big)$ for
$m\!\in\!\{0,\dots,d_\mathrm{ax}/2-1\}$. We use the channel-pair index
$m$ rather than the conventional $k$ to avoid collision with the key
tensor $k$. Each pair is rotated as a 2-D vector by the per-pair angle
\begin{equation}
\theta_\alpha^{(m)}(p)
\;=\;
p \cdot 10000^{-2m/d_\mathrm{ax}},
\qquad
p \;=\; \mathrm{pos}_\alpha \;\in\; \mathbb{N}
\label{eq:rope-angle}
\end{equation}
evaluated at the integer coordinate of the token along axis $\alpha$
(time index for $\alpha\!=\!t$, row for $\alpha\!=\!i$, column for
$\alpha\!=\!j$, view slot for $\alpha\!=\!v$); this matches the
\texttt{lang} frequency convention used by the predictor's
\texttt{RotaryEmbedding} module~\citep{su2024roformer}. Concretely,
for a query channel pair
$\big(q_{o_\alpha+2m},\, q_{o_\alpha+2m+1}\big)$ at axis position
$p\!=\!\mathrm{pos}_\alpha$,
\begin{equation}
\begin{pmatrix} q'_{o_\alpha+2m} \\ q'_{o_\alpha+2m+1} \end{pmatrix}
\;=\;
\begin{pmatrix}
\cos\theta_\alpha^{(m)}(p) & -\sin\theta_\alpha^{(m)}(p) \\
\sin\theta_\alpha^{(m)}(p) & \;\,\cos\theta_\alpha^{(m)}(p)
\end{pmatrix}
\begin{pmatrix} q_{o_\alpha+2m} \\ q_{o_\alpha+2m+1} \end{pmatrix},
\label{eq:rope-2d-rotation}
\end{equation}
and identically for the key channels. Pairs in different axis
sub-blocks rotate independently and at independent frequencies, so the
inner product $\langle q', k'\rangle$ telescopes into a sum of
per-axis, per-pair terms whose differences depend only on relative
positions $\Delta\!\mathrm{pos}_\alpha = \mathrm{pos}_\alpha - \mathrm{pos}'_\alpha$.
This relative-position property lets the predictor generalize
across $V\!\in\!\{1,2\}$ views and across the training and
deployment patch grids of \Cref{subsec:data} without re-learning
positional codes.

Under multi-view training ($V_{\max}{=}2$) we use the full 4-axis
partition $\mathcal{A}\!=\!\{t,i,j,v\}$: the $v$-axis sub-block
contributes its own rotation, and viewpoint identity is
\emph{additionally} supplied by the per-block additive bias of
\Cref{appx:arch-rope-view} (the two are complementary). For purely
single-view runs ($V{=}1$) the $v$-axis sub-block carries no rotation
and is left as the identity. Action and state tokens are detached
from the visual slab before the RoPE rotation is applied: only the
time component of RoPE rotates them (indexed by their step id), so
that a single proprioceptive token per step does not spuriously
inherit a spatial position.

\subsection{Per-block view bias under grouped-query attention}\label{appx:arch-rope-view}

The view information is encoded into the model in two ways. First, the \emph{view position encoding}
disambiguates different views of tokens with the same spatio-temporal position.
This is achieved by rotating the view-axis sub-block ($d_h/4$ channels per head). However,
the view position only informs the model that two different views are
distinct without any encoding of the type of the content
(e.g.\ the model has to know which view represents the side camera looking from the left
and which one is mounted on the wrist). One way of dealing with it is to just let the model
learn this end-to-end; however, we decided to directly inject the view information as
a learnable bias added across the full head width of visual-token queries and keys
after the four-axis RoPE rotation:
\begin{equation}
q_\ell^{(v)} \,\leftarrow\, q_\ell^{(v)} + U_\ell\!\big[c^{(v)}\big],
\qquad
k_\ell^{(v)} \,\leftarrow\, k_\ell^{(v)} + U_\ell\!\big[c^{(v)}\big],
\label{eq:strong-view-cond}
\end{equation}
with $U_\ell\!\in\!\mathbb{R}^{|\mathcal{V}|\times d_h}$ a learnable view embedding (zero-initialised), broadcast across all query and KV heads. This bias supplements, rather than replaces, the view-axis rotation.
A learned global embodiment vector $E_\mathrm{emb}[e]$ and a per-view embedding $E_\mathrm{view}[c^{(v)}]$ are added once at the predictor input.

The strong-view conditioning of \Cref{eq:strong-view-cond} is a
zero-initialised, layer-wise additive bias indexed by the
\emph{semantic} camera id $c^{(v)}\!\in\!\mathcal{V}$ (left, right,
wrist, front, back, head, top), implemented as
$U_\ell\!\in\!\mathbb{R}^{|\mathcal{V}|\times d_h}$ for each
$\ell\!\in\!\{1,\dots,N_{\text{layers}}\}$ and added to all $d_h$
channels of every visual token's queries and keys \emph{after} the
RoPE rotation of \Cref{eq:rope-2d-rotation}. Concretely, let
$Q_\ell\!\in\!\mathbb{R}^{N\times h_q\times d_h}$ and
$K_\ell\!\in\!\mathbb{R}^{N\times h_{kv}\times d_h}$ be the post-RoPE
query and key tensors of layer $\ell$ for a per-step token bag of
length $N$, and let $b_\ell^{(v)}\!:=\!U_\ell[c^{(v)}]\!\in\!\mathbb{R}^{d_h}$
be the bias gathered for view $v$. Under grouped-query
attention~\citep{ainslie2023gqa} with $h_q$ query heads and $h_{kv}$
KV heads sharing the same head dimension $d_h$, the bias is added by
the broadcast rule
\begin{equation}
Q_\ell[n, r, :]\;\mathrel{+}=\;b_\ell^{(v(n))} \quad \forall\,r\!\in\!\{1,\dots,h_q\},
\qquad
K_\ell[n, r, :]\;\mathrel{+}=\;b_\ell^{(v(n))} \quad \forall\,r\!\in\!\{1,\dots,h_{kv}\},
\label{eq:view-bias-broadcast}
\end{equation}
for every visual token index $n$ with view assignment $v(n)$; that
is, the same $d_h$-vector $b_\ell^{(v)}$ is broadcast across
\emph{all} query heads and \emph{all} KV heads of layer $\ell$, so
the table $U_\ell$ has no head dimension and the parameter count is
$|\mathcal{V}|\,d_h$ per layer (for the 1B model:
$7 \cdot 128 \!=\! 896$ parameters per layer, $40 \cdot 896 \!=\! 35{,}840$
in total). The bias is \emph{not} added to action or state tokens,
since these have no view assignment; their post-RoPE queries and keys
propagate through the attention block unchanged.

This per-layer additive bias \emph{supplements} the $v$-axis RoPE
rotation of \Cref{eq:rope-axis-partition}: under our default
strong-view configuration both signals are active, the rotation
contributes a relative-view phase to attention scores, and the bias
contributes a $d_h$-wide per-layer per-camera offset on top of it.
The combination gives a stable handle on viewpoint identity for
multi-view planning (\Cref{subsec:planning}); zero initialisation of
$U_\ell$ makes loading a checkpoint trained without strong view
conditioning a no-op at step~$0$.

\subsection{Attention masking}\label{appx:arch-attention}

We use a similar attention mask to V-JEPA~2-AC~\citep{assran2025vjepa2} but extend it to multiple views:
attention is \emph{timestep-level causal} and restricted to a local temporal window of $8$ timesteps, while
within a step all $V H_p W_p$ visual tokens (plus the action and state tokens) attend omnidirectionally (any-to-any).
More precisely, the rasterized sequence of tokens has $T \times N_{\text{tok}}$ tokens in total and the token at index $k$
can attend to tokens with indices $j$ such that
$\lfloor\frac{k}{N_{\text{tok}}}\rfloor - 7 \leqslant \lfloor\frac{j}{N_{\text{tok}}}\rfloor \leqslant \lfloor\frac{k}{N_{\text{tok}}}\rfloor$.
Schematically, this means that packing the sequence as $T$ frame-blocks of size
$N_{\mathrm{tok}}$ given by \Cref{eq:tokens-per-step}, the attention mask between blocks is
banded lower-triangular: each block of size $N_{\mathrm{tok}}{\times}N_{\mathrm{tok}}$ is fully attended
within the diagonal band of width $W_t$ and zero outside it.
This trick makes the computational cost of attention grow linearly as a function of the step number and it alleviates the problem
of the position extrapolation -- the predicted content quality depends only on the accumulated error, not the
ability of the model to extrapolate, because it never observes a sequence longer than the window size.
This is a classical trick for language models and world models~\citep{assran2025vjepa2, po2025longcontext, panteam2025pan}.
The bounded window keeps the per-step computational cost of attention constant and makes the predictor
KV-cacheable in $\mathcal{O}(W_t \cdot V \cdot H_p \cdot W_p)$ keys per layer; \Cref{subsec:training} uses this at deployment.

\subsection{Per-embodiment input projection}\label{appx:arch-tokenization}

Every embodiment $e$ owns its own input weight matrix $W_a^{(e)}$ in
a single multi-embodiment linear projection. The raw action vector
$a_t\!\in\!\mathbb{R}^{d_a}$ is zero-padded to a common width
$d_a^{\max}$ jointly large enough to absorb the widest action
convention in our mixture (\Cref{subsec:data}; for the 1B run
$d_a^{\max}{=}d_s^{\max}{=}64$, set by the chunked DROID
\texttt{joint\_vel} action-key variant which produces an
$8\!\cdot\!8\!=\!64$-D mega-action) before projection,
\begin{equation}
\tilde a_t \;=\; \big[\,a_t \,;\, \bm{0}_{d_a^{\max}-d_a}\,\big] \;\in\; \mathbb{R}^{d_a^{\max}},
\qquad
\mathrm{ActTok}(a_t,e) \;=\; (W_a^{(e)})^{\!\top}\,\tilde a_t \,+\, b_a^{(e)} \,+\, \mathrm{PE}_t^a,
\label{eq:proprio-embed-appx}
\end{equation}
with $W_a^{(e)}\!\in\!\mathbb{R}^{d_a^{\max}\times D}$, and an
analogous projection $\mathrm{StTok}(s_t,e)$ for state;
$\mathrm{PE}_t^{\{a,s\}}$ are 1-D sin--cos positional embeddings over
time. This lets a single predictor consume the heterogeneous action /
state conventions of \Cref{subsec:data} (chunked DROID controller
commands, multi-D AgiBot World dual-arm states, planar Language-Table
poses, etc.) without per-embodiment branching at the transformer
level: the embodiment identity affects only the input projection.

The action token participates in self-attention as a regular token:
the projected vector $\mathrm{ActTok}(a_t,e)$ is concatenated to the
per-step token bag and attends to (and is attended by) all visual and
state tokens of the same step, and all visual / action / state tokens
within the local temporal window $W_t$. The state token participates
symmetrically. Token-style conditioning, in contrast to additive
action embedding, lets the predictor express step-dependent gating of
action effects (e.g., the gripper acting differently early vs.\ late
in a grasp) without inflating the per-step input dimensionality.

\subsection{Robot-id breakdown}\label{appx:arch-emb}

The training mixture spans $12$ physical robotic platforms (the
\emph{Total} row of \Cref{tab:datasets}). We treat each (platform,
action-source) pair as a distinct robot id, which yields
$|\mathcal{E}|{=}26$ ids, split into $12$ default state-delta ids
(one per platform) plus $14$ extra splits for those datasets that
ship more than one synchronized controller stream or distinct state
convention not collapsed by the action-key pipeline of
\Cref{appx:data-actions}. \Cref{tab:appx-arch-embodiment} lists the
visible breakdown ($\#$ / Platform / Notes); the internal
\texttt{embodiment\_name} strings used by the training config are
collected in the table footnote.

\begin{table}[H]
\caption{Robot-id breakdown used in this work. The first $12$ rows are default state-delta ids, one per physical platform. The next $14$ rows are extra (platform, action-source) or (platform, state-convention) splits.
Total robot ids: $|\mathcal{E}|\!=\!12\!+\!14\!=\!26$.}
\label{tab:appx-arch-embodiment}
\centering
\small
\setlength{\tabcolsep}{6pt}
\begin{tabular}{cll}
\toprule
\# & Platform & Notes \\
\midrule
\multicolumn{3}{l}{\emph{$12$ default state-delta ids (one per physical platform)}} \\
1  & Franka                    & DROID, RoboMind-Franka, Roboset, OXE-FMB (default) \\
2  & Franka + Omron            & RoboCasa (action-key) \\
3  & SO-101                    & LeRobot \\
4  & AgiBot dual-arm (Robotiq) & AgiBot World \\
5  & AgiBot humanoid           & AgiBot World \\
6  & AgileX (dual-arm)         & RoboMind AgileX \\
7  & WidowX (Bridge)           & OXE Bridge, OXE Bridge V2 \\
8  & Google robot              & OXE RT-1, OXE BC-Z \\
9  & Kuka                      & OXE QT-Opt, OXE MT-Opt \\
10 & Planar arm                & OXE Language-Table (sim, real) \\
11 & 1X humanoid               & 1X \\
12 & Baxter                    & OXE RoboNet (Baxter) \\
\midrule
\multicolumn{3}{l}{\emph{$14$ extra (platform, action-source / state-convention) splits}} \\
13 & Franka                    & DROID, joint-state convention \\
14 & Franka                    & DROID, cartesian controller actions \\
15 & Franka                    & DROID, joint controller actions \\
16 & Franka                    & DROID, cartesian-velocity controller actions \\
17 & Franka                    & DROID, joint-velocity controller actions \\
18 & SO-101                    & LeRobot, joint-state convention \\
19 & Franka + Omron            & RoboCasa, state-delta convention \\
20 & Franka                    & RoboMind-Franka, separate id from generic Franka \\
21 & Franka                    & Roboset, separate id from generic Franka \\
22 & Franka                    & OXE RoboNet (Franka split) \\
23 & Franka                    & OXE FMB, separate id from generic Franka \\
24 & Sawyer                    & OXE RoboNet (Sawyer) \\
25 & WidowX                    & OXE RoboNet (WidowX) \\
26 & Kuka                      & OXE RoboNet (Kuka) \\
\bottomrule
\end{tabular}
\end{table}

\subsection{Predictor scaling sweep}\label{appx:arch-scaling}

For the scaling study of \Cref{section:experiments}, we
train $22$M, $50$M, $100$M, $300$M, $1$B, $2$B, $4$B, and $8$B variants with the same block
recipe (pre-norm GQA + GELU MLP, factorised 4-axis RoPE on $(t,i,j,v)$ supplemented
by the per-block view bias of \Cref{eq:strong-view-cond}, encoder, padded action / state widths,
local temporal window, optimizer schedule). The eight variants differ only in
$N_{\text{layers}}$, $D$, and the GQA head counts $(h_q,\,h_{kv})$
(\Cref{tab:appx-arch-scaling}); the head dimension $d_h{=}128$ and the MLP
hidden ratio $4D$ are held constant. The KV head count matches the query head count at the smallest scales and approaches $h_{kv}=h_q/2$ as model size increases.

\begin{table}[H]
\caption{Hyperparameters of the RoboJEPA predictor at eight scales (zero-init view bias). Only depth $N_{\text{layers}}$, width $D$, and head counts $(h_q,h_{kv})$ change across scales; everything else (encoder, factorised RoPE on $(t,i,j,v)$, post-RoPE full-head-width view bias, padded action / state widths, local temporal window $W_t$, attention kernel, output projection, training loss) is held fixed. Predictor parameter counts are approximate $D^2 N_{\text{layers}}$-dominated estimates.}
\label{tab:appx-arch-scaling}
\centering
\small
\setlength{\tabcolsep}{4pt}
\begin{tabularx}{\textwidth}{l *{8}{>{\centering\arraybackslash}X} l}
\toprule
Hyperparameter & 22M & 50M & 100M & 300M & 1B & 2B & 4B & 8B & Notation \\
\midrule
Predictor depth                 & $7$    & $10$    & $14$    & $24$    & $40$    & $40$    & $52$    & $72$    & $N_{\text{layers}}$ \\
Predictor width                 & $512$   & $640$   & $768$   & $1024$  & $1536$  & $2048$  & $2560$  & $3072$  & $D$ \\
Query heads                     & $4$     & $5$     & $6$     & $8$     & $12$    & $16$    & $20$    & $24$    & $h_q$ \\
KV heads (GQA)                  & $4$     & $5$     & $6$     & $8$     & $4$     & $8$     & $10$    & $12$    & $h_{kv}$ \\
Head dim                        & $128$   & $128$   & $128$   & $128$   & $128$   & $128$   & $128$   & $128$   & $d_h$ \\
MLP hidden ratio                & $4$     & $4$     & $4$     & $4$     & $4$     & $4$     & $4$     & $4$     & $r_{\mathrm{ffn}}$ \\
Predictor parameters            & $0.02$\,B & $0.05$\,B & $0.1$\,B & $0.3$\,B & $1.0$\,B & $2.4$\,B & $4.4$\,B & $8.4$\,B & --- \\
\midrule
\multicolumn{10}{l}{\emph{Shared across all scales}} \\
Encoder (frozen)                  & \multicolumn{8}{c}{V-JEPA~2.1 ViT-G \citep{mur2026vjepa21}}                  & $f$ \\
Encoder embed dim                 & \multicolumn{8}{c}{$1664$}                                                    & $D_e$ \\
Encoder patch / tubelet           & \multicolumn{8}{c}{$16{\times}16$ / $1$}                                      & --- \\
Frames per clip                   & \multicolumn{8}{c}{$16$}                                                      & $T$ \\
Spatial grid (per view)           & \multicolumn{8}{c}{$15{\times}20$ at $240{\times}320$}                        & $H_p{\times}W_p$ \\
Views per clip                    & \multicolumn{8}{c}{$1$ or $2$}                                                & $V$ \\
Action / state width (padded)     & \multicolumn{8}{c}{$64$}                                                      & $d_a^{\max}$, $d_s^{\max}$ \\
Action / state tokens per step    & \multicolumn{8}{c}{$1$ / $1$}                                                 & $n_a$, $n_s$ \\
Embodiment / camera ids           & \multicolumn{8}{c}{$26$ / $7$}                                                & $|\mathcal{E}|$, $|\mathcal{V}|$ \\
Position encoding                 & \multicolumn{8}{c}{factorised 4-axis RoPE on $(t,i,j,v)$ \citep{su2024roformer}}        & --- \\
Per-block view conditioning       & \multicolumn{8}{c}{additive $U_\ell[c^{(v)}]$ on $q,k$}                       & $U_\ell$ \\
Local temporal window             & \multicolumn{8}{c}{$8$ frames (causal); spatial $H_p\!\times\!W_p$ and view $V$ unrestricted} & $W_t$ \\
Output projection                 & \multicolumn{8}{c}{$\mathrm{Linear}(D{\to}D_e) + \mathrm{LayerNorm}$}          & $W_\mathrm{out}$ \\
Loss                              & \multicolumn{8}{c}{per-token $\ell_1$ on encoder features (\Cref{eq:loss-tf})} & $\mathcal{L}_{\mathrm{TF}}$ \\
Rollout steps (training)          & \multicolumn{8}{c}{$2$}                                                        & $K_{\mathrm{ar}}$ \\
\bottomrule
\end{tabularx}
\end{table}

\section{Data: Additional Details}\label{appx:data}

\emph{Expands \Cref{subsec:data}.} Heterogeneous robot data become a single token stream by being normalized onto a common temporal grid (\Cref{appx:data-decimation}), a common padded action / state width (\Cref{appx:data-actions}), and a common semantic camera vocabulary (\Cref{appx:data-cameras}). The on-disk format (\Cref{appx:data-format}) and the non-trivial per-dataset conversions needed to reach it (\Cref{appx:data-postproc}) make this routing reproducible, the view-mixed-rank mixture (\Cref{appx:data-mixture}) fixes the sampling, and the full per-dataset breakdown lives in \Cref{appx:data-full}.

\subsection{Dataset mixture and unification}\label{appx:data-unification}

\paragraph{Actions.}
For most datasets, the per-step action is the difference between consecutive states (which we refer to
as the state-delta action): whether it is
an end-effector (EEF) absolute pose, absolute joint angles, a commanded motion, or any
combination of those for multiarm or wheeled robots.
The same demonstration can often be described through several such parameterisations at once (for example joint angle control or EEF pose control), and each one shows the world model the same behavior from a different control perspective, acting as an action-level augmentation.
We exploit this: we found it beneficial to train on different action parameterisations of the same embodiment
whenever they are available. For example, with the DROID dataset arm we cannot only train
on state-delta of the end-effector but also on the velocities of the end-effector or joint angles.
Early in the work, we found that this improves the world model at a fixed training compute budget.

In addition to that, different embodiments provide a much richer potential learning signal
for world models since they expose manipulation through different action spaces,
each following its own unique dynamics and a set of constraints.
Therefore, the world model has more potential to learn the underlying physics and dynamics of
the world than with one embodiment. We treat each (platform, action-space) pair
as a distinct embodiment id,
giving $|\mathcal{E}|{=}26$ over the $12$ physical platforms ($26 = 12$ default state-delta
ids $+ 14$ extra splits; full enumeration in \Cref{tab:appx-arch-embodiment}).

We use a lean per-embodiment linear projection of actions to facilitate learning a shared
latent action tokenization within the world model. The dimensionality of the actions also
varies. For the single-arm end-effector delta poses, we use seven dimensions that
represent a $6$D translation+rotation and gripper closure.
For dual arm platforms (e.g. AgileX in RoboMind or AgiBot World), this gets doubled since we
have to control two arms independently. For wheeled platforms, the action space is extended by
the commanded actions for the wheeled navigation - for example the RoboCasa platform reserves four
actions for navigation atop the standard $7$D EEF control and 1X reserves $2$D for navigation
(see more details in \Cref{tab:appx-datasets-full,appx:data-actions}).

\paragraph{Unification and sampling.}
To facilitate efficient training, we convert all datasets into the unified data format. Since
datasets are collected with different sensor frequencies, loading continuous chunks of data would
often lead to an overly short horizon for prediction. Additionally, different datasets have different
motion velocities due to differences in data collection mechnisms -- often different teleoperated
platforms scale human motion to a faster or slower motion of the robot. Therefore, we employ
different frame sampling frequencies for different datasets which we report in \Cref{tab:appx-datasets-full} (see \Cref{appx:data-decimation} for details).
Each frame is processed via the image embedding of the V-JEPA~2.1 recipe~\citep{mur2026vjepa21}, whose features performed strongly in policy-based evaluations of latent diffusion world models~\citep{nilaksh2026reconstruction} on Bridge~V2~\citep{walke2023bridgev2}.
To correctly represent the total action between sampled frames, we either use the total measured
state delta action or the concatenated
chunk of commanded actions between sampled frames when working with commanded actions.
The world model does not have a well-defined notion of modeling frequency.
Instead, we simply let the world model integrate the total motion represented by the action into the
next predicted latent state.

\subsection{Per-dataset breakdown}\label{appx:data-full}

\Cref{tab:datasets} collapses the OXE, AgiBot World, and RoboMind sub-corpora into single rows. \Cref{tab:appx-datasets-full} gives the full per-dataset breakdown with resolutions and source FPS.

\begin{table}[H]
\caption{Full per-dataset breakdown of the $23$-dataset, $26$-embodiment mixture. ``Act.\ hrs'' counts only frames with synchronized proprioception. The compressed main-text view is \Cref{tab:datasets}; per-rank sampling weights are in \Cref{tab:appx-mixture-weights}.}
\label{tab:appx-datasets-full}
\centering
\small
\setlength{\tabcolsep}{3pt}
\resizebox{\textwidth}{!}{%
\begin{tabular}{lrrrrrl}
\toprule
Dataset & Traj. & Vid.\ hrs & Act.\ hrs & Resolution & FPS & Embodiment \\
\midrule
DROID~\citep{khazatsky2024droid}                & 92k    & 414  & 138  & $720{\times}1280$ & 60 & Franka \\
Roboset~\citep{bharadhwaj2023roboagent}         & 73k    & 1244 & 380  & $240{\times}424$  & 5  & Franka (Roboset) \\
RoboMind Franka~\citep{wu2024robomind}          & 25k    & 92   & 37   & $480{\times}640$  & 30 & Franka \\
RoboMind AgileX~\citep{wu2024robomind}          & 10.3k  & 184  & 61   & $480{\times}640$  & 30 & AgileX \\
LeRobot~\citep{cadene2024lerobot}               & 21.5k  & 182  & 93   & $480{\times}640$  & 30 & SO-101 \\
RoboCasa365~\citep{nasiriany2026robocasa365}       & 32k    & 482  & 482  & $720{\times}1280$ & 20 & Franka + Omron (sim) \\
1X~\citep{1xworldmodel2024}                     & 23.5k  & 90   & 90   & $512{\times}512$  & 30 & 1X humanoid \\
AgiBot World Robotiq~\citep{bu2025agibot}        & 160k   & 7951 & 2649 & $480{\times}640$  & 30 & AgiBot dual-arm \\
AgiBot World Humanoid~\citep{bu2025agibot}       & 7.5k   & 85   & 85   & $480{\times}640$  & 30 & AgiBot humanoid \\
OXE~RT-1~\citep{brohan2023rt1,oxe2023}          & 87k    & 105  & 105  & $256{\times}320$  & 10 & Google robot \\
OXE~Bridge~\citep{ebert2021bridge,oxe2023}      & 25k    & 48   & 48   & $480{\times}640$  & 5  & WidowX \\
OXE~Bridge V2~\citep{walke2023bridgev2,oxe2023} & 60k    & 244  & 121  & $256{\times}256$  & 5  & WidowX \\
OXE~QT-Opt~\citep{kalashnikov2018qtopt,oxe2023} & 580k   & 238  & 238  & $512{\times}640$  & 10 & Kuka \\
OXE~MT-Opt~\citep{kalashnikov2021mtopt,oxe2023} & 920k   & 654  & 654  & $512{\times}640$  & 5  & Kuka \\
OXE~RoboNet Franka~\citep{dasari2019robonet,oxe2023}  & 7.8k & 196 & 63 & $240{\times}320$  & 1 & Franka \\
OXE~RoboNet Baxter~\citep{dasari2019robonet,oxe2023}  & 18k  & 451 & 145 & $240{\times}320$ & 1 & Baxter \\
OXE~RoboNet Sawyer~\citep{dasari2019robonet,oxe2023}  & 50k  & 1254 & 404 & $240{\times}320$ & 1 & Sawyer \\
OXE~RoboNet WidowX~\citep{dasari2019robonet,oxe2023}  & 5k   & 126 & 40 & $240{\times}320$  & 1 & WidowX \\
OXE~RoboNet Kuka~\citep{dasari2019robonet,oxe2023}    & 1.6k & 40  & 12 & $240{\times}320$  & 1 & Kuka \\
OXE~Language-Table sim~\citep{lynch2023languagetable,oxe2023}  & 180k & 259 & 259 & $360{\times}640$ & 5 & Planar arm \\
OXE~Language-Table real~\citep{lynch2023languagetable,oxe2023} & 440k & 392 & 392 & $360{\times}640$ & 5 & Planar arm \\
OXE~BC-Z~\citep{jang2022bcz,oxe2023}            & 43.5k  & 165  & 165  & $171{\times}213$ & 10 & Google robot \\
OXE~FMB~\citep{luo2024fmb,oxe2023}              & 8.6k   & 126  & 31   & $171{\times}213$ & 10 & Franka \\
\midrule
\textbf{Total}                                  & \textbf{2.8713M} & \textbf{15,022} & \textbf{6,692} & --- & --- & 12 platforms \\
\bottomrule
\end{tabular}}
\end{table}

\subsection{Standardised on-disk format}\label{appx:data-format}

Every source dataset is re-packaged into a single object-storage layout so that the dataloader does not need per-source branching beyond a thin adapter that emits the standard tuple consumed by the predictor. Each trajectory exposes three groups: an MP4-encoded camera stream per view (\texttt{cameras/}), HDF5 datasets holding the native proprioceptive vector (\texttt{trajectory/observation/robot\_state/}), and, when the source ships a controller stream, raw controller commands (\texttt{trajectory/action/}). Episode-level metadata (e.g.\ DROID \texttt{building}/\texttt{labs} fields used by the negative filter that drops setup-room recordings) is recorded in a per-dataset \texttt{aggregated\_metadata.json}; a per-dataset index maps episode ids to S3 paths, episode lengths, and per-camera availability so the sampler operates on this index without touching object storage. Per-clip sampling reads $T{=}16$ frames at the world-model rate (\Cref{appx:data-decimation}) and $V$ cameras (\Cref{appx:data-cameras}); intrinsics, extrinsics, and uncalibrated metadata are loaded as NaN-filled placeholders and discarded by the predictor in this configuration.

\subsection{Per-dataset postprocessing}\label{appx:data-postproc}

Beyond the common repackaging of \Cref{appx:data-format}, several datasets require non-trivial numerical, geometric, or video conversions before they can enter the shared token stream. We describe these here grouped by the kind of transformation; the trivial relabelling of fields onto the common schema is omitted.

\paragraph{Video decoding and re-encoding.}
Every source is emitted as one MP4 per view at its native frame rate (\Cref{tab:appx-datasets-full}). Most OXE~\citep{oxe2023} sources ship RLDS/TFDS records with per-step RGB arrays, which we re-encode to MP4 (e.g.\ $1$\,fps for RoboNet~\citep{dasari2019robonet}, $5$\,fps for Bridge~\citep{ebert2021bridge}/MT-Opt~\citep{kalashnikov2021mtopt}/Language-Table~\citep{lynch2023languagetable}, $10$\,fps for RT-1~\citep{brohan2023rt1}/QT-Opt~\citep{kalashnikov2018qtopt}/BC-Z~\citep{jang2022bcz}/FMB~\citep{luo2024fmb}). RoboMind~\citep{wu2024robomind} stores each frame as a JPEG byte-string inside HDF5: we decode every frame individually, falling back to a raw-buffer reshape ($720{\times}1280$ or $480{\times}640$) when the JPEG header is missing, and correct the channel order from BGR to RGB for the real Franka streams (the simulated streams are already RGB). For DROID~\citep{khazatsky2024droid} we discard the stereo/SVO recordings and keep the rectified monocular left, right, and wrist MP4 streams. For sources with multiple nominal cameras we detect and drop all-zero (absent) camera streams so they are not written as empty views (e.g.\ the unused auxiliary cameras in Bridge~V2~\citep{walke2023bridgev2}).

\paragraph{Forward kinematics for joint-only sources.}
Roboset~\citep{bharadhwaj2023roboagent} and LeRobot~\citep{cadene2024lerobot} record only joint encoders, with no end-effector pose. We reconstruct the end-effector Cartesian pose with a differentiable forward-kinematics model of the corresponding arm (a Franka Panda for Roboset, the SO-101 URDF for LeRobot), applied to both the state trajectory and, for Roboset, the commanded-joint trajectory to recover a Cartesian action. LeRobot joint encoders are first converted from degrees to radians. For Roboset we additionally apply constant per-joint offsets (a $+\pi/2$ shift on the wrist joints of both state and action, and a $-\pi/2$ shift on one action joint) so that the joint zero convention matches the URDF used by the kinematics model.

\paragraph{Orientation representation.}
Sources that store the end-effector orientation as a quaternion (RT-1, QT-Opt, MT-Opt, FMB, the simulated RoboMind Franka split) and the forward-kinematics outputs above are all converted to an $xyz$ Euler parameterisation, so that the entire mixture shares a single $3$-D translation $+$ $3$-D Euler orientation convention for end-effector poses.

\paragraph{Gripper normalisation.}
Gripper signals are placed on a common $[0,1]$ closedness scale. Roboset state grippers are divided by their observed open width ($0.8305$) and clipped to $[0,1]$, and its command grippers are affinely mapped from $[-1,1]$ to $[0,1]$. WidowX sources (Bridge and Bridge~V2) invert and clip the raw gripper signal, $g \leftarrow 1-\mathrm{clip}(g,0,1)$, so that larger values consistently denote a more closed gripper.

\paragraph{Multi-arm and humanoid assembly.}
Bimanual and humanoid embodiments are assembled into a single per-step vector. For the RoboMind AgileX dual-arm platform we concatenate the left and right end-effector poses and the left and right arm joints, taking each arm's last joint as its gripper. Bridge~(v1) actions, which are stored as separate translation and rotation-delta fields, are concatenated into a single delta-pose action vector, and BC-Z end-effector poses are assembled from separately stored translation and axis-angle observations.

\subsection{View-mixed-rank training: mixture composition}\label{appx:data-mixture}

The 1B run uses $W_{\mathrm{ddp}}{=}256$ data-parallel ranks (32 nodes $\times$ 8 H100s), partitioned into two equally sized \emph{view-groups}:

\begin{itemize}
\item \textbf{Two-view group} (ranks $0{\dots}127$): each rank loads $V_g{=}2$ camera views per clip with $B_g{=}2$ clips per microbatch and gradient accumulation $K_g{=}1$, contributing $4$ view-clips per rank per step, i.e.\ $512$ view-clips group-wide.
\item \textbf{One-view group} (ranks $128{\dots}255$): each rank loads $V_g{=}1$ view per clip with $B_g{=}2$, $K_g{=}1$, contributing $2$ view-clips per rank per step, i.e.\ $256$ view-clips group-wide.
\end{itemize}

This yields the effective per-step batch of $768$ view-clips ($512$ clips) derived in \Cref{appx:training:batch} (\Cref{eq:appx-effective-batch}). The two groups share predictor parameters and step the optimizer in lockstep.

\Cref{tab:appx-mixture-weights} lists the per-(dataset, action-source) sampling weights for both groups in a single table, together with the per-row total. The top block lists datasets that ship two reliably informative external cameras and feed both groups; the lower block lists single-camera sources that feed only the 1-view group. For datasets appearing in both groups, the 1-view rank uses a contracted camera regex selected so that the rank still sees a representative external view (e.g.\ left/right exterior for DROID excludes the wrist; AgiBot World Robotiq is restricted to its head-mounted RGB stream).

\begin{table}[H]
\caption{Sampling weights for the 1-view and 2-view groups of the 1B run. The top block lists datasets that ship two reliably informative external cameras and feed both groups; the lower block lists single-camera sources that feed only the 1-view group. DROID and RoboCasa appear under multiple (platform, action-source) pairs --- each one is a distinct embodiment id with its own per-embodiment input projection (\Cref{appx:data-actions}) --- but all variants share the same underlying source episodes.}
\label{tab:appx-mixture-weights}
\centering
\small
\setlength{\tabcolsep}{4pt}
\begin{tabular}{lrrrr}
\toprule
Dataset (action-source) & 1-view weight & 2-view weight & Total & \shortstack{Marginal dataset \\ weight} \\
\midrule
\cline{5-5}
DROID (state-delta cartesian)        & $0.0233$ & $0.0306$ & $0.0282$ & \multirow{6}{*}{0.1692} \\
DROID (state-delta joints)           & $0.0233$ & $0.0306$ & $0.0282$ &  \\
DROID (action-key cartesian)         & $0.0233$ & $0.0306$ & $0.0282$ &  \\
DROID (action-key joints)            & $0.0233$ & $0.0306$ & $0.0282$ &  \\
DROID (action-key cartesian-vel)     & $0.0233$ & $0.0306$ & $0.0282$ &  \\
DROID (action-key joint-vel)         & $0.0233$ & $0.0306$ & $0.0282$ &  \\
\cline{5-5}
\cline{5-5}
LeRobot (cartesian)                  & $0.0103$ & $0.0135$ & $0.0124$ & \multirow{2}{*}{0.0248} \\
LeRobot (joints)                     & $0.0103$ & $0.0135$ & $0.0124$ &  \\
\cline{5-5}
AgiBot World Robotiq                 & $0.4478$ & $0.5878$ & $0.5411$ &  \\
RoboMind AgileX                      & $0.0104$ & $0.0136$ & $0.0125$ &  \\
\cline{5-5}
RoboCasa (action-key)                & $0.0271$ & $0.0356$ & $0.0328$ & \multirow{2}{*}{0.0656} \\
RoboCasa (state-delta)               & $0.0271$ & $0.0356$ & $0.0328$ &  \\
\cline{5-5}
Roboset                              & $0.0701$ & $0.0920$ & $0.0847$ &  \\
OXE Bridge V2                        & $0.0137$ & $0.0180$ & $0.0166$ &  \\
RoboMind Franka                      & $0.0052$ & $0.0068$ & $0.0063$ &  \\
\midrule
OXE RT-1                             & $0.0059$ & ---      & $0.0020$ &  \\
OXE Bridge                           & $0.0027$ & ---      & $0.0009$ &  \\
OXE QT-Opt                           & $0.0134$ & ---      & $0.0045$ &  \\
OXE MT-Opt                           & $0.0368$ & ---      & $0.0123$ &  \\
OXE RoboNet Franka                   & $0.0110$ & ---      & $0.0037$ &  \\
OXE RoboNet Baxter                   & $0.0254$ & ---      & $0.0085$ &  \\
OXE RoboNet Sawyer                   & $0.0706$ & ---      & $0.0235$ &  \\
OXE RoboNet WidowX                   & $0.0071$ & ---      & $0.0024$ &  \\
OXE RoboNet Kuka                     & $0.0023$ & ---      & $0.0008$ &  \\
OXE Language-Table sim               & $0.0146$ & ---      & $0.0049$ &  \\
OXE Language-Table real              & $0.0221$ & ---      & $0.0074$ &  \\
OXE BC-Z                             & $0.0093$ & ---      & $0.0031$ &  \\
OXE FMB                              & $0.0071$ & ---      & $0.0024$ &  \\
1X                                   & $0.0051$ & ---      & $0.0017$ &  \\
AgiBot World Humanoid                & $0.0048$ & ---      & $0.0016$ &  \\
\midrule
\textbf{Group total}                 & \textbf{1.0000} & \textbf{1.0000} & --- &  \\
\bottomrule
\end{tabular}
\end{table}

\paragraph{Effective per-step mixture.}
At the clip level the split is $50/50$; at the \emph{token-with-loss} level the run sees $\tfrac{2}{3}$ of the signal at $V{=}2$ (the $V{=}2$ clips contribute two view-clips each). Both view counts therefore appear in every optimizer step, so the per-block view conditioning (\Cref{eq:strong-view-cond}) and the factorised RoPE of \Cref{subsec:model_architecture} are jointly exposed to both regimes rather than separated by curriculum.

\subsection{Frame rate, decimation and the per-frame mask}\label{appx:data-decimation}

Every dataset is normalized to a target world-model rate $f_{\text{wm}}$ that determines the temporal stride between consecutive frames in a clip. Source FPS varies by two orders of magnitude across the mixture (\Cref{tab:appx-datasets-full}): $1$\,Hz for OXE RoboNet, $5$--$10$\,Hz for most OXE sources, $20$--$30$\,Hz for RoboCasa / RoboMind / AgiBot / LeRobot, $60$\,Hz for DROID. The loader decimates by the integer factor $c \;=\; \lceil f_{\text{raw}} / f_{\text{wm}} \rceil$ and indexes frames as \verb|np.arange(start_frame, end_frame, c)|, so one predictor step always corresponds to a fixed $1/f_{\text{wm}}$ of robot wall-clock regardless of source FPS. Trajectories whose effective length falls below $T{=}16$ are right-padded with zero frames, and a per-frame mask
\begin{equation}
m_{t,v} \;=\; \mathbb{1}[\text{frame }(t,v)\text{ is a real, non-padded sample}]
\;\in\;\{0,1\}
\label{eq:appx-data-mask}
\end{equation}
is propagated to the loss --- the same $m_{t,v}$ that gates the masked $\ell_1$ in \Cref{eq:loss-tf} and the rollout objective in \Cref{subsec:training}. Within a single clip the mask is the same across views, $m_{t,v}\!\equiv\!m_t$, because clip-level padding is applied symmetrically to all $V$ streams; we keep the per-view subscript in \Cref{eq:loss-tf} for notational consistency with the $V{=}1$ vs $V{=}2$ rank groups.

\paragraph{Rate bridge $f_{\text{wm}}\!\to\!f_{\text{ctrl}}$ at deployment.}
At deployment the Franka controller of \Cref{subsec:planning} runs at $f_{\text{ctrl}}{=}10$\,Hz while the world model operates at $f_{\text{wm}}{=}5$\,Hz. The planner replans on every other controller tick and the $7$-D end-effector delta returned by the planner is split into $f_{\text{ctrl}}/f_{\text{wm}}{=}2$ equal sub-steps emitted back-to-back; this keeps training and deployment aligned on the same $200$\,ms predictor stride and isolates the controller from the world-model rate.

\subsection{Action conditioning details}\label{appx:data-actions}

The mixture exposes two complementary action-vector pipelines per (dataset, action-source) pair. Each pair is a distinct \emph{embodiment id} from the $|\mathcal{E}|{=}26$ registry consumed by the predictor's per-embodiment input projection (\Cref{subsec:model_architecture}); duplicates of the same physical platform under different action sources (e.g.\ the six DROID embodiments listed in \Cref{tab:appx-mixture-weights}) share encoder inputs but learn distinct $W_a^{(e)}$, $W_s^{(e)}$ slices.

\paragraph{State-delta pipeline (default).}
For most datasets the per-step action is computed as a finite difference of the synchronized state trajectory: $\Delta x_t \in \mathbb{R}^3$ for the end-effector position, $\Delta R_t \in \mathfrak{so}(3)$ for orientation (computed in the Lie algebra so that the magnitude is independent of the rotation representation used by the source), and a scalar $\Delta g_t$ for the gripper. The vector concatenates these into a per-platform native width: $7$-D for the standard Franka / Bridge / Kuka end-effector, $14$-D for AgileX (two end-effectors), $23$-D for AgiBot dual-arm, $25$-D for the 1X humanoid, and so on across the mixture.

\paragraph{Action-key pipeline (high-rate sources).}
For datasets that ship a high-rate controller stream --- DROID (four cartesian / joint, position / velocity variants) and RoboCasa (cartesian-position) --- the loader instead reads the raw controller stream at $f_{\text{raw}}$ and concatenates the $f_{\text{raw}}/f_{\text{wm}}$ sub-step commands within one world-model step into a single per-step vector. For DROID at $60\!\to\!7.5$\,Hz this yields a chunk of $8$ controller commands per predictor step (up to $8{\times}8{=}64$ dimensions); for RoboCasa at $20\!\to\!5$\,Hz it yields a chunk of $4$ commands. This recovers the high-frequency controller signal (which the world model must match if its rollout is to drive a downstream planner --- \Cref{subsec:planning}) without changing the $f_{\text{wm}}$ stride at which video features are produced.

\paragraph{Padding to $d_a^{\max}{=}64$ and per-embodiment projection.}
To batch heterogeneous embodiments together within a single rank's microbatch, every per-step action vector --- whether a single state-delta or an action-key chunk --- is right-zero-padded to the fixed width $d_a^{\max}{=}64$ used by the predictor's per-embodiment input projection. The width $64$ is chosen jointly to absorb the widest native state-delta (the $25$-D 1X humanoid) and the largest action-key chunk emitted by the pipeline above. The per-embodiment projection $W_a^{(e)}\!\in\!\mathbb{R}^{D\times d_a^{\max}}$ is a multi-embodiment linear (a single $|\mathcal{E}|\!\times\!D\!\times\!d_a^{\max}$ weight tensor indexed by embodiment id) that recovers the native action space from the padded slot. The same construction is mirrored on the proprioceptive side ($d_s^{\max}$, $W_s^{(e)}$); the action and state tokens enter the predictor as $n_a{=}n_s{=}1$ tokens per step in \Cref{eq:tokens-per-step}.

At deployment the planner of \Cref{subsec:planning} targets a single Franka-style $7$-D end-effector delta convention, so the action dimensionality the user sees on a real robot is fixed regardless of which subset of the mixture trained the model. The deployment pipeline mirrors the training-time padding and embodiment-id lookup exactly, so the projection $W_a^{(e)}$ used at planning time is the same one trained against the corresponding state-delta variant.

\subsection{Image preprocessing}\label{appx:data-image}

The video tensor returned by the loader has shape $[C, V, T, H_{\text{img}}, W_{\text{img}}]$ with $C{=}3$, $V\!\in\!\{1,2\}$ and $T{=}16$ frames at $f_{\text{wm}}$. Spatial cropping follows the V-JEPA~2.1 input recipe~\citep{mur2026vjepa21}: at training time each clip is randomly resized with aspect ratio sampled from $[1.333, 1.777]$ and area scale sampled from $[0.5, 1.0]$, then center-cropped to $240{\times}320$, while at evaluation we use a deterministic $1.0/1.0$-scale crop of the same size. Random horizontal flips, auto-augmentation, motion shift and random erasing are disabled to avoid distorting end-effector geometry. Frames are normalized with the standard ImageNet statistics~\citep{deng2009imagenet} expected by the V-JEPA~2.1 encoder~\citep{mur2026vjepa21}, and the resulting $240{\times}320$ tensor is fed to the frozen encoder which produces an $H_p{\times}W_p{=}15{\times}20$ patch grid with $D_e{=}1664$ channels per patch.

\subsection{Per-camera enumeration}\label{appx:data-cameras}

Every physical camera in the mixture is mapped at load time to one of $|\mathcal{V}|{=}7$ semantic camera ids by a per-dataset regex map. Cameras with the same semantic role (e.g.\ all left-exterior cameras across DROID, Roboset, RoboMind, RoboCasa, OXE FMB) are mapped to the same id and therefore share the learned per-block view bias of \Cref{eq:strong-view-cond} regardless of which dataset they come from or whether the rank is $V{=}1$ or $V{=}2$. \Cref{tab:appx-data-cameras} reports the mapping in transposed form (semantic ids run across columns, datasets down rows).

\begin{table}[H]
\caption{Mapping from raw per-dataset camera names to the $|\mathcal{V}|{=}7$ semantic camera ids consumed by the predictor's per-block view conditioning (\Cref{eq:strong-view-cond}). Rows are datasets; columns are semantic ids; each cell shows the raw camera name(s) routed to that id by the per-dataset regex. After regex matching, only the semantic id $c\!\in\!\{0,\dots,6\}$ is consumed by the predictor's per-block view bias $U_\ell[c^{(v)}]$; the original camera names are not seen by the model. The \emph{Default / unmapped} bucket (id $4$) collects sources that emit a single generic camera (LeRobot, OXE RoboNet) or a non-canonical camera label (OXE Bridge V2 \texttt{camera\{1,2\}}) for which no semantic role is reliably consistent across the corpus.}
\label{tab:appx-data-cameras}
\centering
\scriptsize
\setlength{\tabcolsep}{3pt}
\resizebox{\textwidth}{!}{%
\begin{tabular}{lccccccc}
\toprule
Semantic id $c$ & $0$ & $1$ & $2$ & $3$ & $4$ & $5$ & $6$ \\
Role            & Left exterior & Right exterior & Wrist / eye-in-hand & Head / front-facing & Default / unmapped & Top-down & Planar-arm overhead \\
\midrule
DROID                & \texttt{left} & \texttt{right} & \texttt{wrist} & --- & --- & --- & --- \\
Roboset              & \texttt{left} & \texttt{right} & \texttt{wrist} & --- & --- & \texttt{top} & --- \\
RoboMind             & \texttt{camera\_left} & \texttt{camera\_right} & --- & \texttt{camera\_front} & --- & \texttt{camera\_top} & --- \\
RoboMind AgileX      & --- & --- & \texttt{camera\_\{left,right\}\_wrist} & \texttt{camera\_front} & --- & --- & --- \\
RoboCasa             & \texttt{robot0\_agentview\_left} & \texttt{robot0\_agentview\_right} & \texttt{robot0\_eye\_in\_hand} & --- & --- & --- & --- \\
OXE FMB              & \texttt{left} & \texttt{right} & --- & --- & --- & --- & --- \\
OXE Bridge           & --- & \texttt{right} & --- & --- & --- & --- & --- \\
OXE Bridge V2        & --- & \texttt{right\_back} & --- & --- & \texttt{camera\{1,2\}} & --- & --- \\
OXE RT-1             & --- & --- & --- & \texttt{back} & --- & --- & --- \\
OXE BC-Z             & --- & --- & --- & \texttt{back} & --- & --- & --- \\
OXE RoboNet          & --- & --- & --- & --- & \texttt{.*} & --- & --- \\
OXE Language-Table   & --- & --- & --- & --- & --- & --- & \texttt{front} \\
LeRobot              & --- & --- & --- & --- & \texttt{.*} & --- & --- \\
AgiBot World          & --- & --- & \texttt{hand\_\{left,right\}\_color} & \texttt{head\_color} & --- & --- & --- \\
1X                   & --- & --- & --- & \texttt{front} & --- & --- & --- \\
\bottomrule
\end{tabular}}
\end{table}

\section{Training: Additional Details}\label{appx:training}

\emph{Expands \Cref{subsec:training}.} We document the per-rank batch decomposition across view-groups (\Cref{appx:training:batch}). The loss is defined once in \Cref{eq:loss-tf}.

We use data parallelism for models up to $2$B parameters and FSDP2 for the $4$B and $8$B models during pretraining and short-horizon prediction training. For long-horizon prediction finetuning, we instead use data parallelism combined with tensor parallelism ($\mathrm{tp}{=}2$) for the $4$B and $8$B models. Sharing each sequence across two tensor-parallel GPUs allows an effective per-GPU batch size below one sequence in the most memory-intensive group (two-view, high-resolution), accommodating the $K{=}10$ autoregressive rollout used during this cooldown.

\subsection{Per-rank batch decomposition and view groups}
\label{appx:training:batch}

The dataloader of \Cref{subsec:data} partitions the $W_{\mathrm{ddp}}{=}256$ DDP ranks into $G_v{=}2$ \emph{view-groups}, indexed by $g\!\in\!\{1,2\}$. Group $g$ owns $R_g{=}128$ ranks, processes clips with $V_g$ cameras ($V_1{=}1$, $V_2{=}2$), and runs $K_g$ gradient-accumulation microbatches per optimizer step ($K_1{=}K_2{=}1$ for the 1B run, but the same code path supports $K_g{>}1$ on memory-constrained configurations). Each microbatch on a rank in group $g$ contains $B_g$ clips ($B_2{=}2$ for $g{=}2$, $B_1{=}2$ for $g{=}1$). The effective number of clip-views consumed per optimizer step is therefore
\begin{equation}
N_{\mathrm{cv}}^{\mathrm{eff}}
\;=\;
\sum_{g=1}^{G_v} R_g \cdot K_g \cdot B_g \cdot V_g
\;=\;
128\!\cdot\!1\!\cdot\!2\!\cdot\!1
\;+\;
128\!\cdot\!1\!\cdot\!2\!\cdot\!2
\;=\;
768,
\label{eq:appx-effective-batch}
\end{equation}
i.e.\ $256$ single-view clips and $256$ two-view clips ($512$ clips, $768$ clip-views) per optimizer step. Because the encoder is run per-view independently, the encoder forward sees the full $768$ clip-views; the predictor processes each clip as a single sequence of $T\!\cdot\!(V_g H_p W_p + n_a + n_s)$ tokens (concretely $16\cdot 302$ for $g{=}1$ and $16\cdot 602$ for $g{=}2$ with the geometry of \Cref{tab:appx-arch-scaling}).

\subsection{Cooldown phase}
\label{appx:training:cooldown}

The schedule of \Cref{subsec:training} is warmup--stable--decay (WSD)~\citep{hu2024minicpm}: warmup linear $0\!\to\!\eta$, then a long stable phase held flat at $\eta{=}5{\times}10^{-5}$, then a \emph{cooldown} (decay) phase. We do not run cooldown inline; instead we resume the final stable-phase checkpoint as a fresh job that (i) swaps the optimizer schedulers to a linear LR cooldown paired with a constant weight-decay schedule, (ii) raises the autoregressive rollout horizon from $K_{\mathrm{ar}}{=}2$ to $K_{\mathrm{ar}}{=}10$, and (iii) re-partitions the ranks into four view$\times$resolution groups. The LR decays as $\eta_t = \eta\,(1 - \min(t, N_{\mathrm{cool}})/N_{\mathrm{cool}})$ with $N_{\mathrm{cool}}{=}5000$ optimizer steps, reaching $0$ at the end; weight decay stays at $0.1$ throughout (the cooldown is a pure LR sweep), and the per-group $\mu$P \texttt{lr\_scale} factors are preserved.

\paragraph{Four view$\times$resolution groups.}
During cooldown the two view-groups of \Cref{appx:training:batch} are each split by input resolution, giving four groups that evenly partition the world ($25\%$ of ranks each on the $128$-rank 1B cooldown): two-view $240{\times}320$, two-view $480{\times}640$, one-view $240{\times}320$, and one-view $480{\times}640$. The predictor and the frozen V-JEPA~2.1 encoder are unchanged: at $480{\times}640$ the encoder's $16{\times}16$ patching yields a $30{\times}40$ grid, i.e.\ $H_p W_p{=}1200$ tokens per view per frame versus $300$ at $240{\times}320$ (a $4\times$ increase), and the predictor's factorised RoPE (\Cref{appx:arch-rope}) interpolates to the larger grid with no new parameters. High-resolution groups use a smaller per-rank batch and matched gradient accumulation to keep per-GPU memory within budget; the gradient-synchronization machinery is unchanged, now aggregating four groups instead of two in the single per-step all-reduce.

\subsection{Training hyperparameters}
\label{appx:training:hparams}

\Cref{tab:training-hparams-appx} lists the loss, optimizer, and cooldown settings for the $1$B predictor, complementing the model configurations in \Cref{tab:appx-arch-scaling}.

\begin{table}[H]
\caption{Additional training hyperparameters for the 1B-predictor RoboJEPA run, covering the loss, optimizer, and cooldown configuration. Settings already listed in \Cref{tab:appx-arch-scaling}, including the loss type and pretraining rollout horizon, are omitted here. The loss is defined in \Cref{eq:loss-tf}, and AdamW follows \citet{loshchilov2019adamw}.}
\label{tab:training-hparams-appx}
\centering
\small
\setlength{\tabcolsep}{6pt}
\begin{tabularx}{\linewidth}{>{\raggedright\arraybackslash}p{0.31\linewidth} l >{\raggedright\arraybackslash}X}
\toprule
Hyperparameter & Symbol & Value \\
\midrule
\multicolumn{3}{l}{\emph{Loss (defined once in \Cref{eq:loss-tf})}} \\
Pre-loss normalization                   & $\mathrm{LN}_0$                  & parameter-free LayerNorm on $D_e$ \\
Rollout step index                       & $\tau\!\in\!\{0,\dots,K_{\mathrm{ar}}\}$ & --- \\
Rollout prefix sampling                  & $\tau_0$                         & uniform on $\{0,\dots,T{-}K_{\mathrm{ar}}\}$ \\
Recurrent rollout input                  & ---                              & detached from autograd graph \\
Rollout / TF mixing                      & ---                              & $\tfrac{1}{2}\!:\!\tfrac{1}{2}$ \\
\midrule
\multicolumn{3}{l}{\emph{Optimiser (AdamW \citep{loshchilov2019adamw})}} \\
$(\beta_1,\beta_2)$                      & ---                              & $(0.9,\,0.95)$ \\
Numerical floor                          & $\varepsilon$                    & $10^{-8}$ \\
Learning rate (peak, held flat)          & $\eta$                           & $5\!\times\!10^{-5}$ \\
Schedule                                 & ---                              & warmup $\to$ flat (stable) $\to$ linear cooldown (WSD \citep{hu2024minicpm}) \\
Warm-up (epochs)                         & $W_{\mathrm{warm}}$              & $10$ \\
Weight decay (constant)                  & ---                              & $0.1$ \\
Decay-group scope                        & ---                              & all $\geq\!2$-D parameter tensors \\
No-decay scope                           & ---                              & biases, $1$-D LayerNorm, view-bias table $U_\ell$, per-embodiment input projections, RoPE buffers \\
Gradient clipping (global $\ell_2$)      & ---                              & $1.0$ \\
\midrule
\multicolumn{3}{l}{\emph{Cooldown phase (\Cref{appx:training:cooldown}), resumed from stable-phase checkpoint}} \\
Cooldown steps                           & $N_{\mathrm{cool}}$              & $5000$ \\
LR schedule (cooldown)                   & ---                              & linear $\eta\!\to\!0$ over $N_{\mathrm{cool}}$ \\
Weight decay (cooldown)                  & ---                              & held constant at $0.1$ \\
Rollout horizon (cooldown)               & $K_{\mathrm{ar}}$               & $10$ \\
View$\times$resolution groups            & ---                              & $4$: $\{1,2\}$-view $\times\,\{240{\times}320,\,480{\times}640\}$ \\
High-res grid / tokens per view-frame    & $H_p{\times}W_p$                 & $30{\times}40 = 1200$ ($4\times$ base $15{\times}20$) \\
\bottomrule
\end{tabularx}
\end{table}

\subsection{DROID 3-view Finetuning at 720$\times$1280}
\label{appx:training:droid-720p}

We additionally finetune a high-resolution DROID world model. It initializes from the $8$B long-horizon
prediction training checkpoint at $300{,}000$ pretraining steps (\Cref{appx:scaling-laws:compute-cooldown}), and the
only things we change relative to the scaling runs are the input resolution and the number of views.
Training uses DROID~\citep{khazatsky2024droid} only, with three fixed views at $720{\times}1280$
resolution over $16$ timesteps. The three views are the exterior-left, exterior-right, and wrist cameras,
selected in this fixed order; this is unlike pretraining, where the views are sampled in a random order.
With patch size $16$, each $720{\times}1280$ view becomes a $45{\times}80{=}3{,}600$-token grid, so a full
sequence is about $170{,}000$ input tokens. At this resolution the model would also generate more than
$10{,}000$ tokens per forward pass, which highlights the utility of a feedforward world model that
predicts all next-frame tokens in one pass.

We keep the same world-model objective as in pretraining but with a $K{=}10$-step autoregressive rollout.
We train for $2{,}000$ optimizer steps on $256$ GPUs (tensor-parallel $\mathrm{tp}{=}4$, giving $64$
data-parallel replicas and a global batch of $64$ sequences) with AdamW (weight decay $0.1$, gradient
clipping), warming the learning rate up to $5{\times}10^{-5}$ over $140$ optimization steps and then
decaying it with a cosine schedule over the remaining steps. The per-step compute of this stage is the
finetune row of \Cref{tab:per-step-compute-8b}.

\section{Diffusion Decoder: Additional Details}\label{appx:decoder}

\emph{Expands \Cref{subsec:decoder}.}

\paragraph{Role and rate hierarchy.}
The world model produces dense V-JEPA features at $f_{\mathrm{wm}}{=}5$\,Hz, i.e.\ one feature frame for every fourth RGB frame; the Cosmos tokenizer compresses the raw video time axis by $4{\times}$, so each Cosmos latent frame aligns with exactly one V-JEPA frame. The decoder therefore does two jobs at once --- \emph{reconstruct} the appearance the V-JEPA features already encode, and \emph{imagine} the three RGB frames of motion between consecutive V-JEPA frames. Because $f_{\mathrm{wm}}$ is low, the inter-frame motion the decoder needs to fabricate is small, which is why a relatively small DiT (${\sim}800$M parameters) suffices for $240{\times}320$ video.

\paragraph{Visualisation-only.}
The decoder $D_\phi$ is used \emph{only for visualization}: it is never invoked inside CEM, and no gradient flows between it and the world model. Its training data is exclusively frozen V-JEPA~2.1 encoder features~\citep{mur2026vjepa21}; the encoder, the predictor of \Cref{subsec:model_architecture}, and both halves of the Cosmos-0.1-Tokenizer-CV4x8x8~\citep{nvidia2025cosmos} are all frozen, and the only trainable parameters are those of $D_\phi$ itself.

\paragraph{Frame alignment between V-JEPA and Cosmos.}
Throughout this appendix a decoder clip spans $T_{\text{dec}}$ V-JEPA steps ($T_{\text{dec}}{=}8$ in our experiments) and $\nu\!\in\!\{0,\dots,T_{\text{dec}}\}$ indexes its frames (we reserve the symbol $f$ for the frozen V-JEPA encoder). To keep the two grids exactly aligned, the dataloader loads video at two frame rates simultaneously: a high-fps stream of $4T_{\text{dec}}{+}1$ RGB frames that the Cosmos encoder compresses into the latents to be denoised, and a low-fps stream of every fourth frame, i.e.\ the $T_{\text{dec}}{+}1$ RGB frames $0, 4, \dots, 4T_{\text{dec}}$, that the V-JEPA encoder turns into conditioning features. Because the Cosmos tokenizer downsamples space by $8\times$ and time by $4\times$, encoding the first frame on its own, the $4T_{\text{dec}}{+}1$ RGB frames yield $T_{\text{dec}}{+}1$ Cosmos latent frames of shape $H_p{\times}W_p$ with 16 latent channels. The frame-rate multiplier equals the tokenizer's $4{\times}$ temporal compression, so each Cosmos latent frame $\nu$ corresponds to exactly one V-JEPA frame $z_\nu$; for $T_{\text{dec}}{=}8$ the clip has $9$ V-JEPA frames (RGB frames $0, 4, \dots, 32$), $33$ RGB frames and $9$ Cosmos latent frames. Operating in this $8{\times}8{\times}4$-compressed latent space, rather than on pixels, is what makes training a transformer denoiser on $240{\times}320$ video tractable on a single node.

\paragraph{DiT denoiser architecture.}
$D_\phi$ is a PixArt-style diffusion transformer~\citep{peebles2023dit, chen2024pixart} with $L{=}32$ blocks of hidden dimension $d{=}1280$. Each block uses grouped-query attention~\citep{ainslie2023gqa} with $20$ query heads and $10$ KV heads, an MLP ratio of $4.0$, and rotary position embeddings~\citep{su2024roformer} on the attention queries and keys. Self-attention is restricted to a \emph{local temporal window} of $W_\tau{=}4$ Cosmos frames (full attention across the spatial $H_p{\times}W_p$ grid), which keeps the computational cost per token linear in the clip length and admits an efficient block-sparse mask while encouraging local temporal coherence. This $W_\tau{=}4$ window acts on Cosmos latents (themselves $4{\times}$ temporally compressed) and is distinct from the predictor's $W_t{=}8$-V-JEPA-frame window of \Cref{subsec:model_architecture}.

\paragraph{Conditioning.}
The V-JEPA features $z_{0:T_{\text{dec}}}$ enter $D_\phi$ through dense spatial concat conditioning. The features are upsampled with nearest-neighbour interpolation to the Cosmos grid (in space; in time the two grids already match one-to-one) and concatenated, along the channel axis, with the patchified noisy Cosmos input; a learned linear $\mathrm{Linear}(d{+}1664)\!\to\!d$ then mixes the two streams into the input token embeddings of the first DiT block. The adaLN modulation~\citep{peebles2023dit} of every DiT block is conditioned only on the diffusion noise level: for each Cosmos frame $\nu$, the embedding of its timestep $t_\nu$ produces six per-block parameters $(\text{shift}_{\text{attn}}, \text{scale}_{\text{attn}}, \text{gate}_{\text{attn}}, \text{shift}_{\text{ffn}}, \text{scale}_{\text{ffn}}, \text{gate}_{\text{ffn}})$ that modulate both the attention and FFN sub-layers for that frame. Concretely, with input tokens $h\in\mathbb{R}^{B\times T\times N\times d}$,
\begin{equation}
h' \;=\; \text{gate}\,\odot\,\text{Sublayer}\!\big(\text{LN}(h)\odot(1{+}\text{scale}) + \text{shift}\big) \;+\; h ,
\label{eq:dec-adaln}
\end{equation}
applied per Cosmos frame before flattening back to the joint $(T\!\cdot\!N)$-token sequence consumed by attention.

\paragraph{Training objective.}
We train $D_\phi$ with a standard DDPM~\citep{ho2020ddpm} forward process applied to the Cosmos latents. Let $t_\nu \!\sim\! \mathcal{U}\{0,\dots,K{-}1\}$ denote the diffusion timestep sampled independently for each frame $\nu$, following diffusion forcing~\citep{chen2024diffusionforcing}. With a noise schedule $\bar\alpha_t$, the per-frame forward corruption is
\begin{equation}
x_{t_\nu}^{(\nu)} \;=\; \sqrt{\bar\alpha_{t_\nu}}\;x_0^{(\nu)} \;+\; \sqrt{1-\bar\alpha_{t_\nu}}\;\epsilon^{(\nu)}, \quad \epsilon^{(\nu)} \sim \mathcal{N}(0, I) .
\label{eq:dec-forward}
\end{equation}
Unlike the original $\epsilon$-parameterisation of~\citet{ho2020ddpm}, the network directly predicts the clean signal $\hat x_0^{(\nu)} = D_\phi(x_{t_\nu}^{(\nu)}, t_\nu, z)$. The training loss sums per-frame $x_0$-MSE terms, with $t_\nu$ promoted to a per-frame random variable inside the expectation:
\begin{equation}
\mathcal{L}_{\text{dec}}(\phi)
\;=\;
\mathbb{E}_{x_0,\;\{t_\nu\overset{\text{iid}}{\sim}\mathcal{U}\{0,\ldots,K-1\}\}_{\nu},\;\epsilon,\;z}
\!\left[\,\sum_{\nu}\big\lVert
D_\phi(x_{t_\nu}^{(\nu)}, t_\nu, z) - x_0^{(\nu)}
\big\rVert_2^2\,\right] .
\label{eq:dec-loss}
\end{equation}
We use unweighted $x_0$-MSE, with no additional timestep-dependent weighting. For equivalent $x_0$ and $\epsilon$ predictions at $0<\bar\alpha_t<1$, the squared errors satisfy $\|\hat x_0-x_0\|^2 = \frac{1-\bar\alpha_t}{\bar\alpha_t}\|\hat\epsilon-\epsilon\|^2$. Thus, expressed as an $\epsilon$-prediction loss, this corresponds to inverse-SNR weighting, emphasizing high-noise rather than low-noise timesteps. The condition $z$ is dropped to the zero vector independently per sample with probability $p_{\text{drop}}{=}0.1$ during training to support classifier-free guidance~\citep{ho2022classifierfree}, although in practice we use a guidance scale of $w{=}1.0$ (no guidance) by default since this is the most faithful for visualization. We use the log-SNR-parameterised cosine schedule of Simple Diffusion~\citep{hoogeboom2023simple} in preference to the cosine schedule of IDDPM~\citep{nichol2021iddpm}: parameterising in log-SNR gives finer control over the signal-to-noise trajectory and avoids the degenerate near-zero-SNR behavior at $t\!\to\!K{-}1$ that the IDDPM schedule exhibits. We use $K{=}1000$ training timesteps and enforce zero terminal SNR following~\citet{lin2024zerosnr}; the per-frame independent $t_\nu$-sampling above is what makes the trained denoiser robust to mixed noise levels and, in turn, what enables prefix-conditioned generation at inference time.

\paragraph{Sampling.}
Despite training with $K{=}1000$ DDPM steps, at inference we run only \textbf{2} DDIM steps~\citep{song2021ddim} from $x_{K-1}\!\sim\!\mathcal{N}(0,I)$ per Cosmos frame and pipe the result through the frozen Cosmos decoder to pixels. We attribute this to the strength of the V-JEPA conditioning: dense V-JEPA~2.1 features already pin down most of the appearance, so the denoiser's effective conditional distribution is nearly unimodal and its sampling path nearly straight. For streaming generation we keep the same local-window KV cache as the predictor (\Cref{appx:kv}) and write each new frame's keys/values only on the last DDIM step so the cache never holds noisy activations; the streaming sampler runs at $4$ DDIM steps per new frame, the small overhead buying robustness to the cache turnover. AOT compilation of $D_\phi$ for offline visualization is described alongside the predictor's compilation pipeline in \Cref{appx:infer}; all decoder hyperparameters are collected in \Cref{tab:decoder-hparams}.

\begin{table}[H]
\caption{Diffusion-decoder hyperparameters. The decoder is trained on frozen V-JEPA~2.1 encoder features only; the world model, V-JEPA encoder, and Cosmos tokenizer~\citep{nvidia2025cosmos} are frozen, with no gradient flow between the decoder and the world model. The local temporal window $W_\tau{=}4$ acts on Cosmos latents (themselves $4{\times}$ temporally compressed) and is distinct from the predictor's $W_t{=}8$-frame window over V-JEPA features (\Cref{subsec:model_architecture}).}
\label{tab:decoder-hparams}
\centering
\small
\begin{tabularx}{\linewidth}{>{\raggedright\arraybackslash}X l >{\raggedright\arraybackslash}X}
\toprule
Hyperparameter & Symbol & Value \\
\midrule
\multicolumn{3}{l}{\emph{Architecture}} \\
DiT depth                                          & $L$           & 32 \\
Hidden dim                                         & $d$           & 1280 \\
Query / KV heads                                   & --            & 20 / 10 (GQA, \citep{ainslie2023gqa}) \\
MLP ratio                                          & --            & 4.0 \\
Cosmos latent channels                             & --            & 16 \\
V-JEPA~2.1 condition dim                           & $D_e$         & 1664 \\
Local temporal attention window                    & $W_\tau$      & 4 Cosmos frames \\
Position encoding                                  & --            & RoPE \citep{su2024roformer} \\
Total trainable parameters                         & --            & $\sim$800\,M \\
\midrule
\multicolumn{3}{l}{\emph{Diffusion training}} \\
V-JEPA steps per decoder clip                      & $T_{\text{dec}}$ & 8 ($9$ V-JEPA / $9$ Cosmos frames) \\
Diffusion steps (training)                         & $K$           & 1000 \\
Noise schedule                                     & --            & cosine simple-diffusion \citep{hoogeboom2023simple} \\
Terminal SNR                                       & --            & enforced to $0$ \citep{lin2024zerosnr} \\
Per-frame timesteps                                & --            & diffusion forcing \citep{chen2024diffusionforcing} \\
Parameterisation                                   & --            & $x_0$-prediction \\
Condition dropout (CFG training)                   & $p_{\text{drop}}$ & $0.1$ \\
\midrule
\multicolumn{3}{l}{\emph{Optimisation}} \\
Optimiser                                          & --            & AdamW, $\beta{=}(0.9, 0.95)$ \\
Learning rate                                      & $\eta$        & $10^{-4}$ (constant) \\
Grad clip                                          & --            & $1.0$ \\
Precision                                          & --            & bfloat16 mixed \\
\midrule
\multicolumn{3}{l}{\emph{Sampling}} \\
DDIM steps (inference, default / streaming KV-cached) & --         & $\mathbf{2}$ / $4$ \\
Temporal upsampling (V-JEPA $\to$ RGB)            & --            & $4{\times}$ \\
Guidance scale                                     & $w$           & $1.0$ (default) \\
\bottomrule
\end{tabularx}
\end{table}

\clearpage

\section{Qualitative World-Model Imagination Examples}\label{appx:demonstration_figures}

\emph{Model-size comparison filmstrips.} Each figure seeds RoboJEPA with the
initial frame(s) of a held-out episode, replays the recorded actions, and shows
the decoded imagination for every model size ($22$M--$8$B) alongside the
ground-truth (GT) video. Rows are ordered smallest-to-largest with \textbf{GT at
the bottom}, directly beneath the $8$B model for the most direct comparison; each
row is sampled at $k{=}5$ equal-time columns spanning the first $20$\,s of the
rollout. \textit{Big} figures are decoded at the $480{\times}640$ encoder
resolution and \textit{small} at $240{\times}320$; the number of camera views is
fixed per embodiment.

\newcommand{\demofig}[1]{figures/appendix_demonstration_figures/#1}

\subsection{Large resolution ($480{\times}640$)}\label{appx:demo_large}

\begin{figure}[H]
\centering
\includegraphics[width=\textwidth, height=0.85\textheight, keepaspectratio]{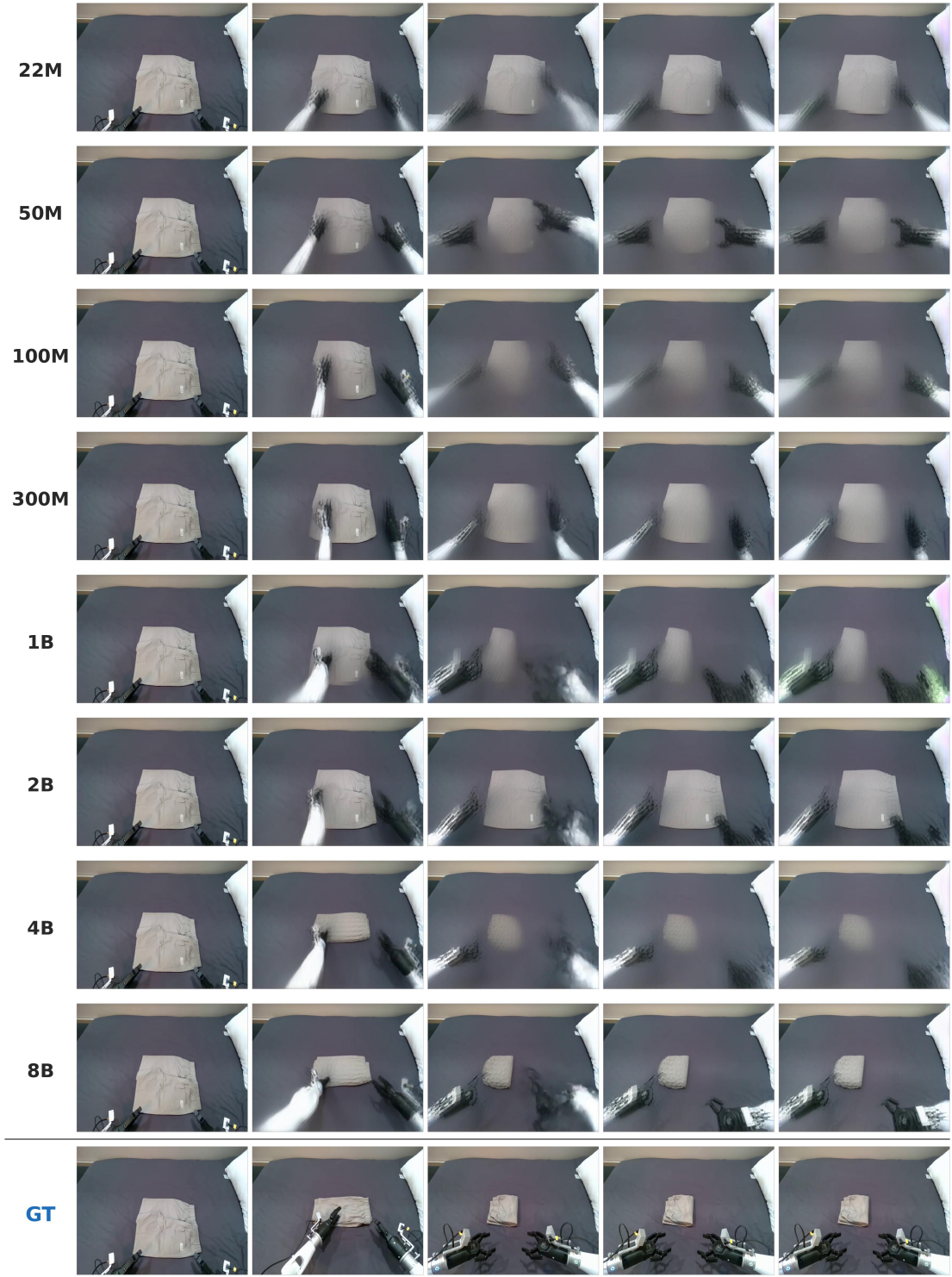}
\caption{AgiBot (single view), $480{\times}640$. Model-size comparison of world-model
imaginations against ground truth; folding structure emerges only at larger scales.}
\label{fig:demo_agibot_1006_v1}
\end{figure}

\begin{figure}[H]
\centering
\includegraphics[width=\textwidth, height=0.85\textheight, keepaspectratio]{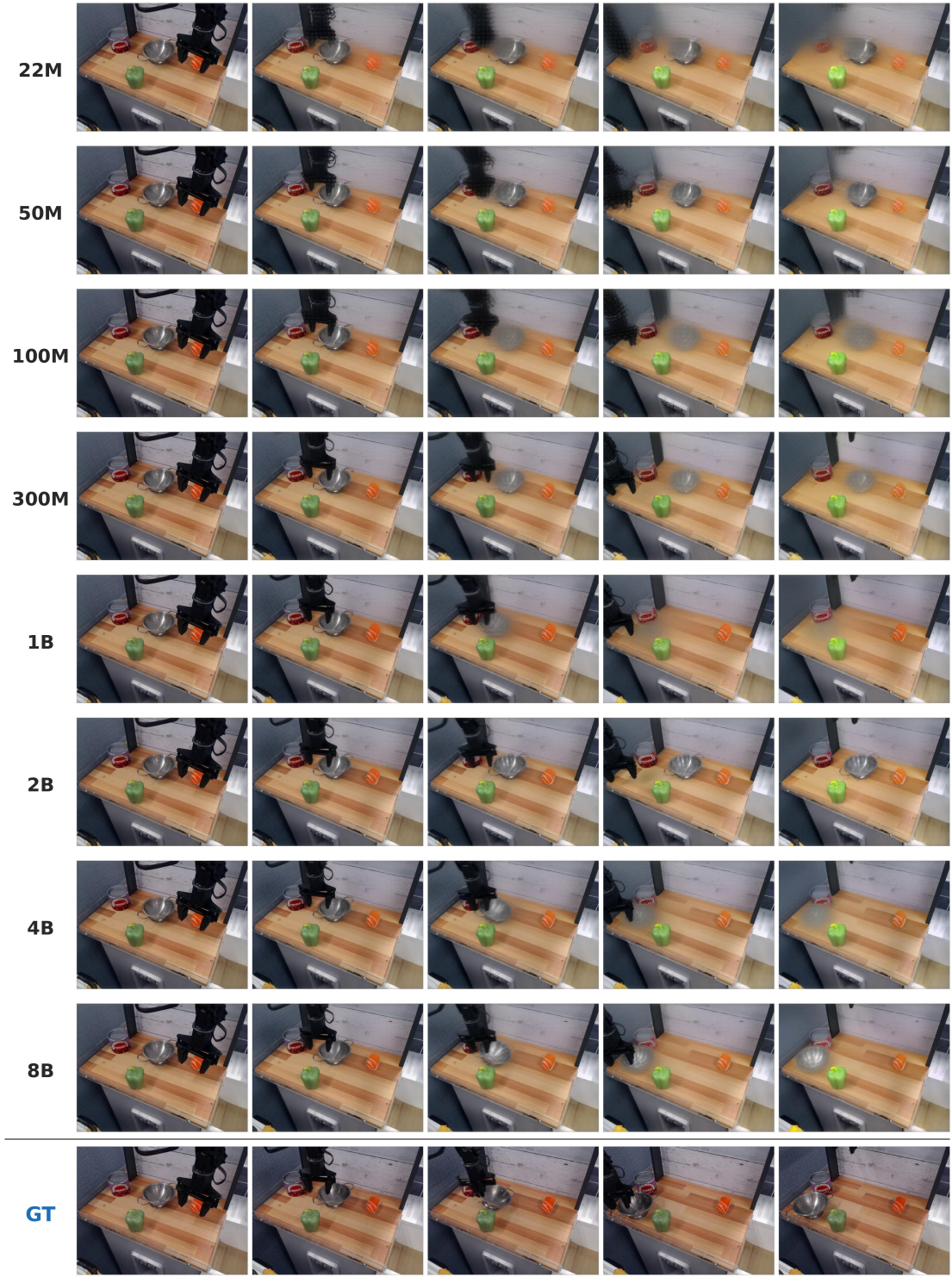}
\caption{Open X-Embodiment Bridge (single view), $480{\times}640$. Physical contact
between the gripper and the object develops only for the larger models.}
\label{fig:demo_oxe_bridge_1002_v1}
\end{figure}

\begin{figure}[H]
\centering
\includegraphics[width=\textwidth, height=0.85\textheight, keepaspectratio]{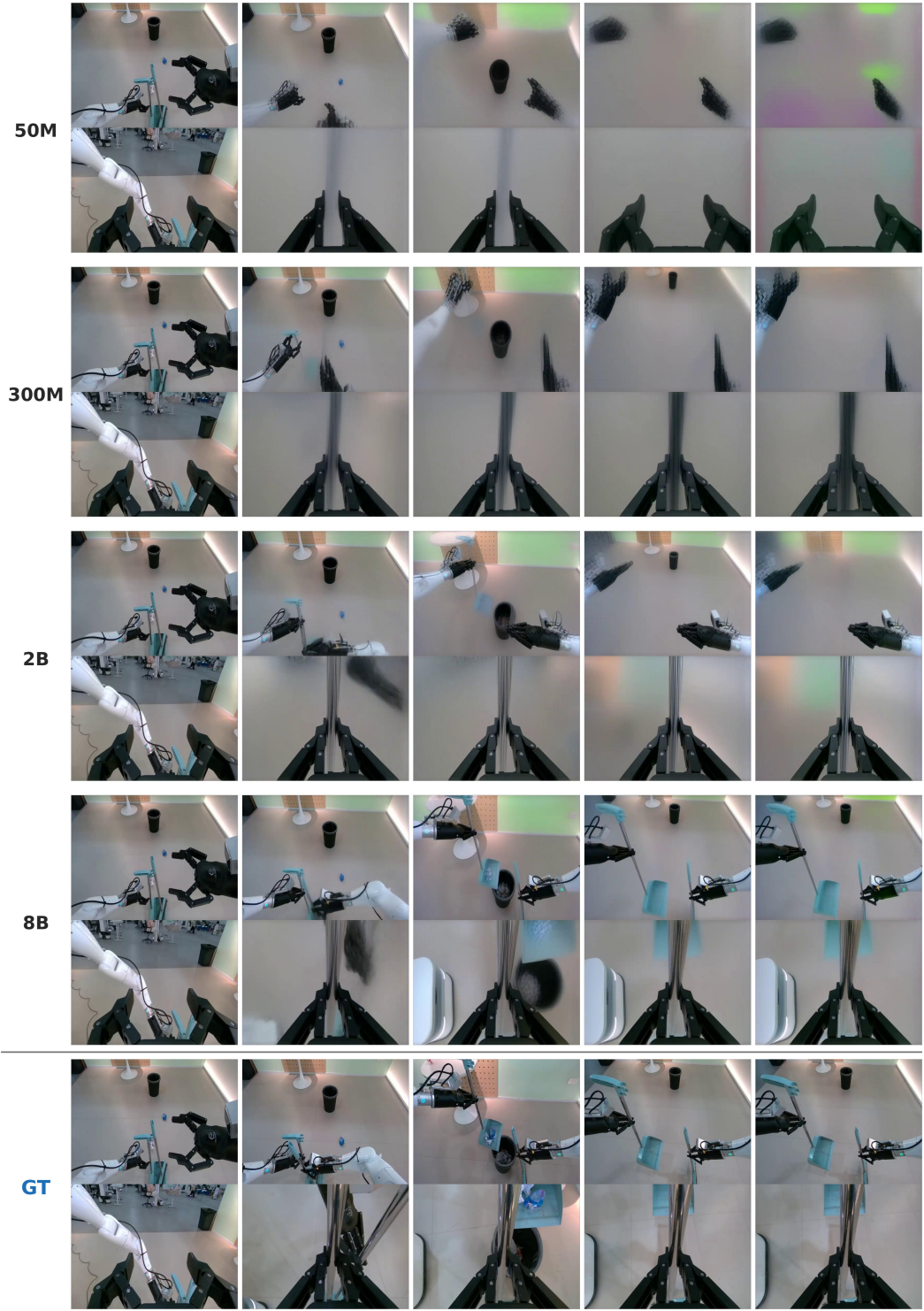}
\caption{AgiBot (two views), $480{\times}640$. Bimanual manipulation rollout across
model sizes; finer details are preserved as scale increases.}
\label{fig:demo_agibot_1008_v2}
\end{figure}

\begin{figure}[H]
\centering
\includegraphics[width=\textwidth, height=0.85\textheight, keepaspectratio]{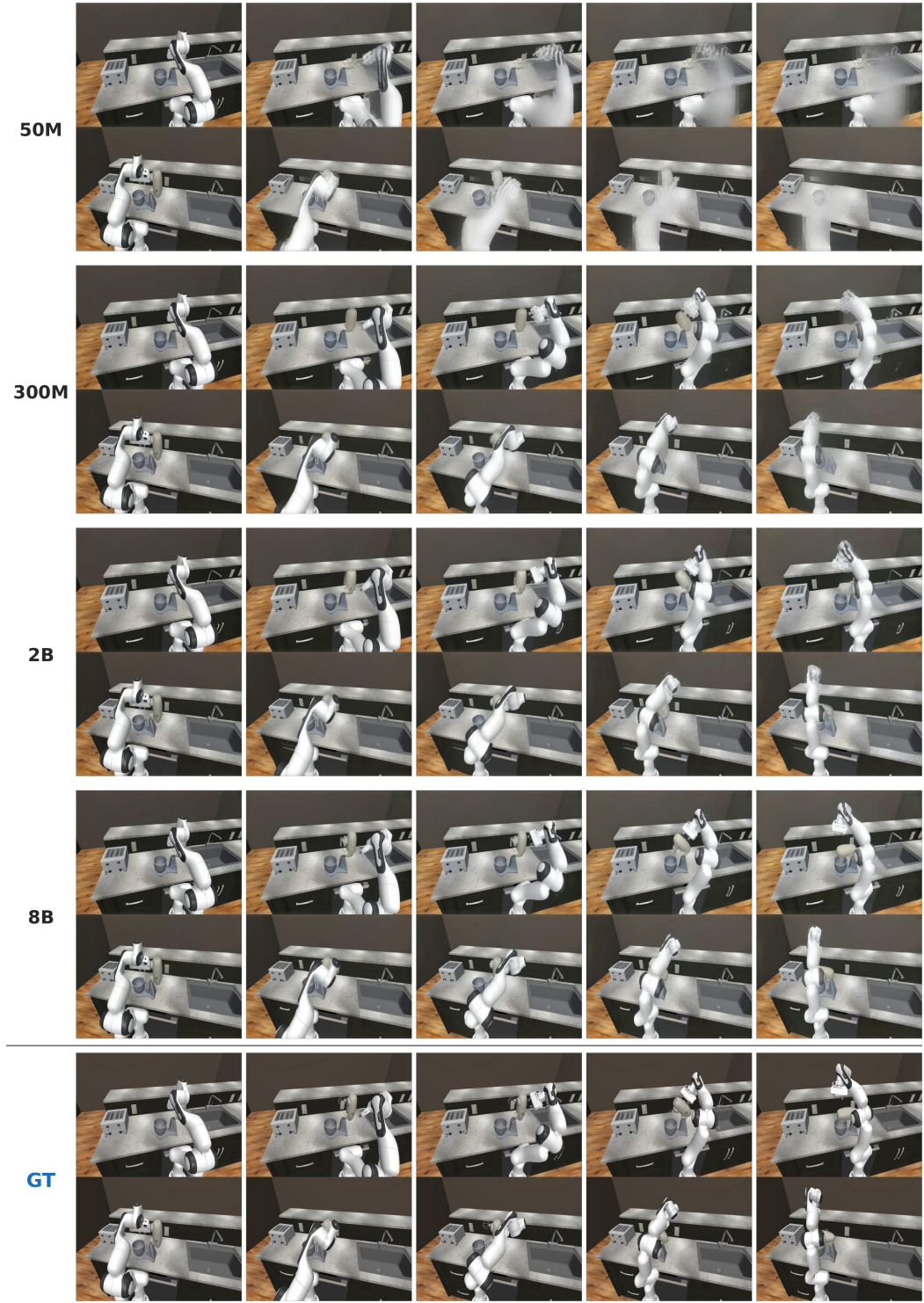}
\caption{RoboCasa (two views), $480{\times}640$. Simulated kitchen manipulation
rollout across model sizes.}
\label{fig:demo_robocasa_1002_v2}
\end{figure}

\Cref{fig:demo_agibot_1006_v1,fig:demo_oxe_bridge_1002_v1,fig:demo_agibot_1008_v2,fig:demo_robocasa_1002_v2} show high-resolution rollouts on AgiBot, Bridge, and RoboCasa with one or two camera views.

\clearpage

\clearpage

\clearpage

\clearpage

\subsection{Small resolution ($240{\times}320$)}\label{appx:demo_small}

\begin{figure}[H]
\centering
\includegraphics[width=\textwidth, height=0.85\textheight, keepaspectratio]{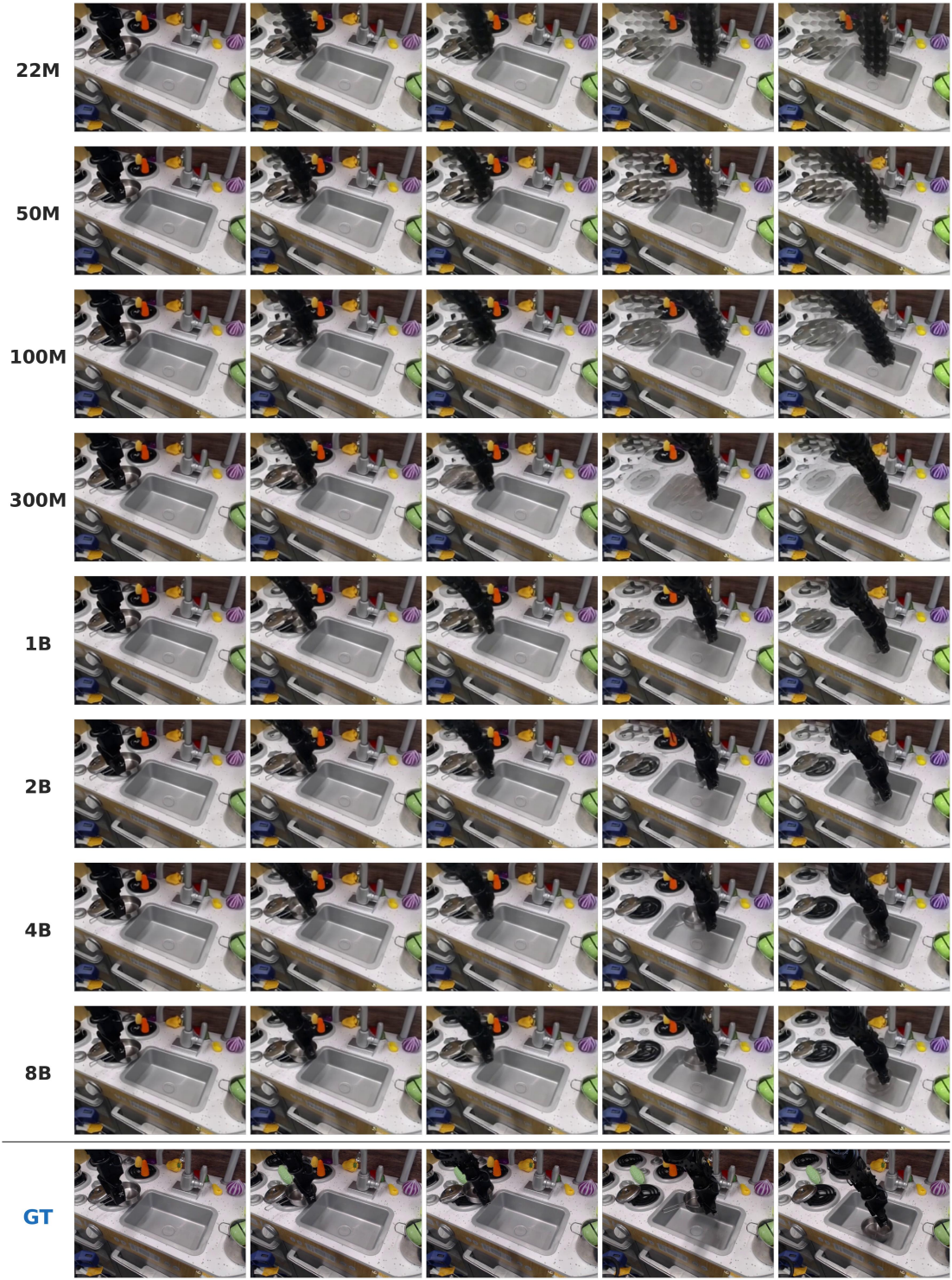}
\caption{Open X-Embodiment Bridge (single view), $240{\times}320$. Contact with the
manipulated object develops only in the larger models.}
\label{fig:demo_oxe_bridge_1006_v1}
\end{figure}

\begin{figure}[H]
\centering
\includegraphics[width=\textwidth, height=0.85\textheight, keepaspectratio]{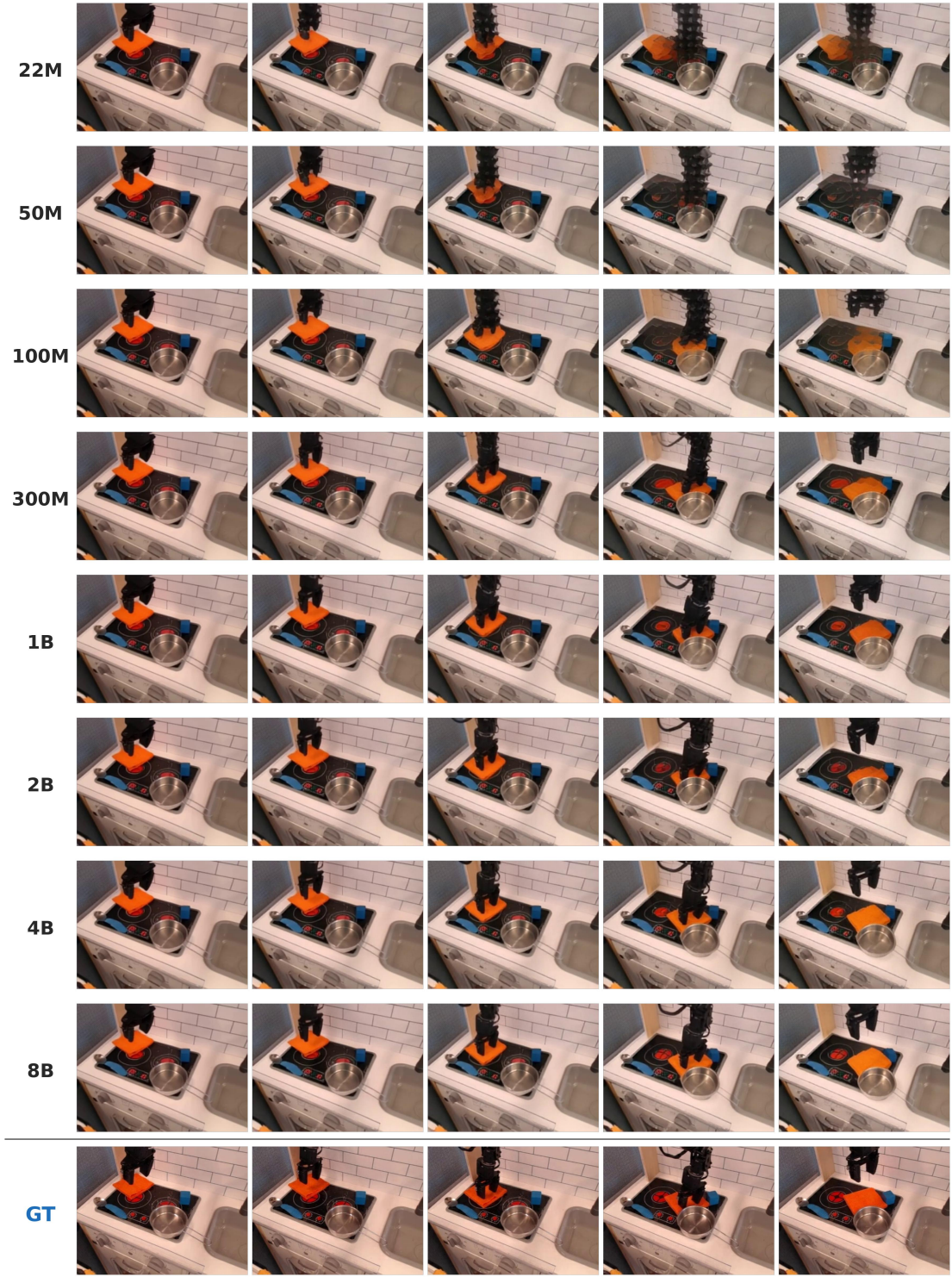}
\caption{Bridge-V2 (single view), $240{\times}320$. Clear progression of imagination
quality with model scale.}
\label{fig:demo_bridge2_1002_v1}
\end{figure}

\begin{figure}[H]
\centering
\includegraphics[width=\textwidth, height=0.85\textheight, keepaspectratio]{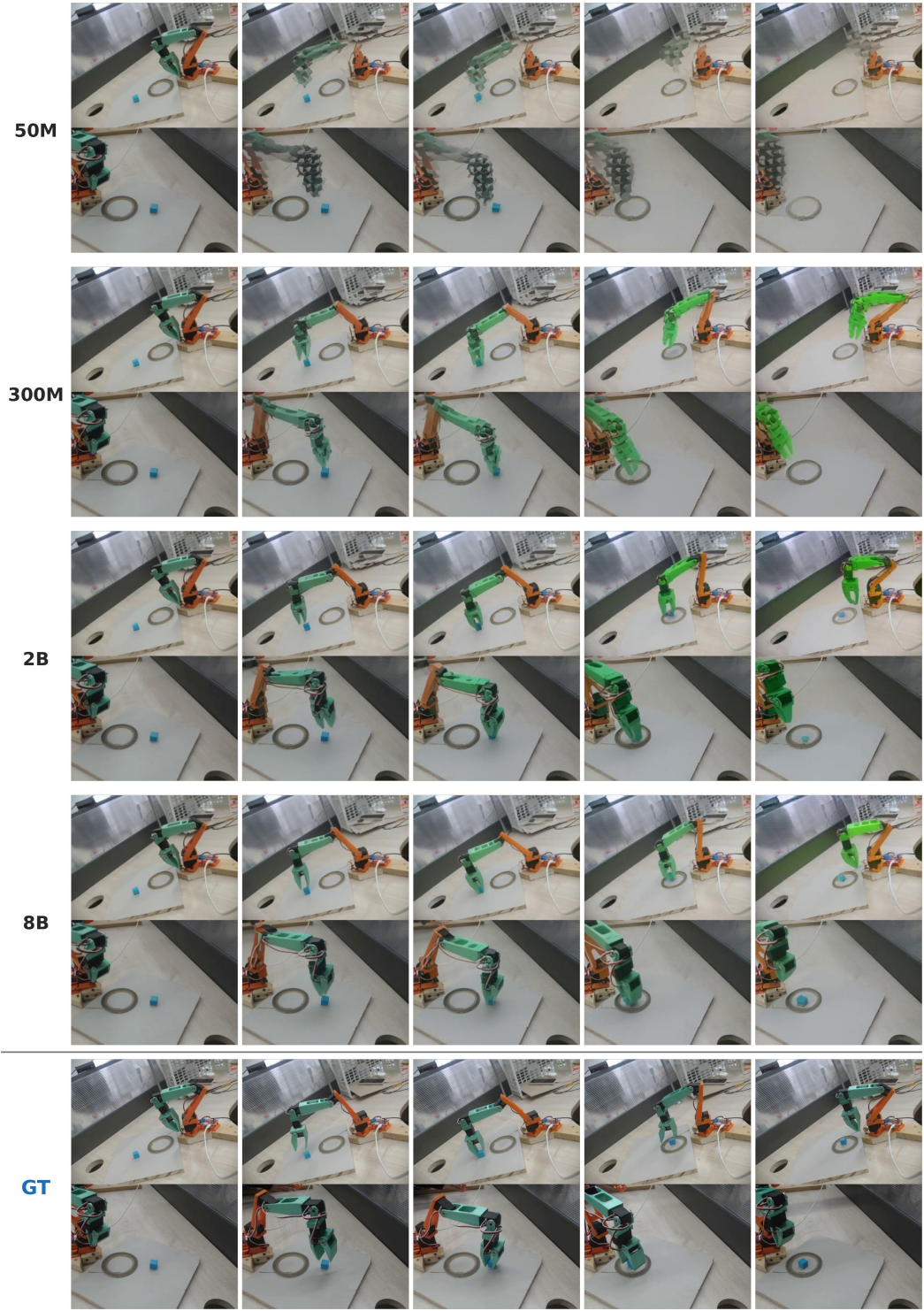}
\caption{LeRobot (two views), $240{\times}320$. Two-view manipulation rollout across
model sizes.}
\label{fig:demo_lerobot_1011_v2}
\end{figure}

\begin{figure}[H]
\centering
\includegraphics[width=\textwidth, height=0.85\textheight, keepaspectratio]{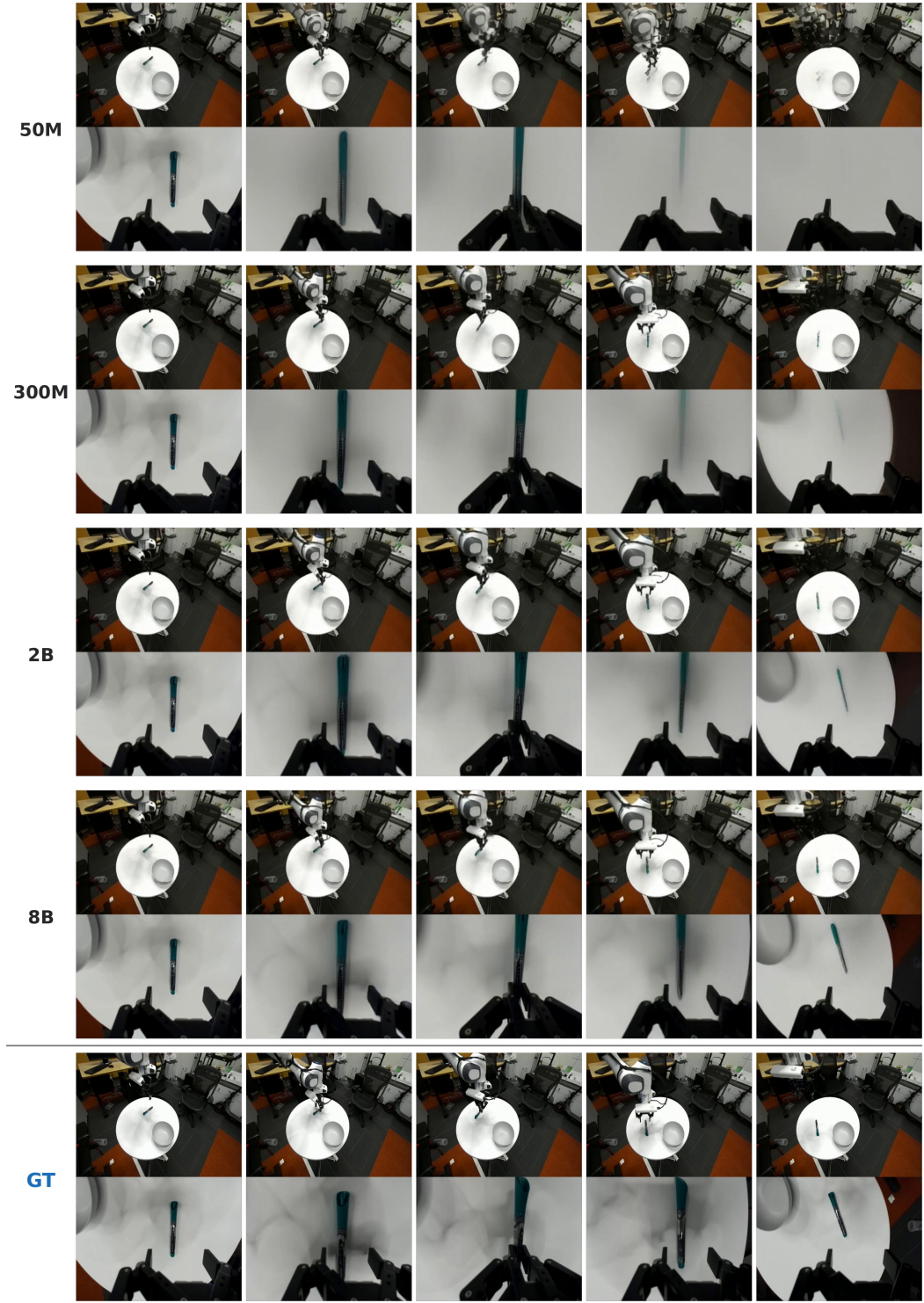}
\caption{Franka / DROID (two views), $240{\times}320$, joint-space actions. Two-view
manipulation rollout across model sizes.}
\label{fig:demo_franka_1021_joint_ak}
\end{figure}

\Cref{fig:demo_oxe_bridge_1006_v1,fig:demo_bridge2_1002_v1,fig:demo_lerobot_1011_v2,fig:demo_franka_1021_joint_ak} show low-resolution rollouts on Bridge, Bridge V2, LeRobot, and DROID.

\clearpage

\clearpage

\clearpage

\clearpage

\section{Planning: Additional Details}\label{appx:planning}

\emph{Expands \Cref{subsec:planning}.} We give the full algorithm box (\Cref{appx:planning-alg}), the data-parallel elite selection (\Cref{appx:planning-distributed}), the spherical translation-sampling distribution (\Cref{appx:planning-spherical}), the $\ell_2$-ball projection (\Cref{appx:planning-ball}), and multi-view goal handling (\Cref{appx:planning-multiview}). The computational cost of \Cref{eq:plan} matches the per-token training loss (\Cref{subsec:training}, \Cref{eq:loss-tf}) and the controller-rate bridge (\Cref{appx:data-decimation}) is what carries the planner output to the robot.

\subsection{Full CEM algorithm}\label{appx:planning-alg}

\Cref{alg:cem-full} restates one call to CEM with all loop details the main text suppressed. It exposes the per-iteration micro-batch loop ($M$ micro-batches per GPU), the cross-rank \textsc{AllGather}, the explicit per-step return accumulation $R^{(n)} = \sum_h r^{(n)}_h$, and the momentum-smoothed Gaussian refit. Across consecutive environment steps the planner is warm-started by shifting the previous mean by $L_{\mathrm{exec}}$ steps and re-using its tail, so it converges in only $I_{\mathrm{step}}$ refinement iterations after the cold-start computational cost of $I_0$.

\begin{algorithm}[t]
\caption{Receding-horizon CEM planning in V-JEPA feature space (full version).}
\label{alg:cem-full}
\begin{algorithmic}[1]
\Require predictor $g_\theta$, goal features $\{z^{(v)}_g\}$, current
context $z_{t-L_{\mathrm{ctx}}+1:t}$, horizon $H$, samples per
iteration per GPU $N_{\mathrm{samp}}$, micro-batches $M$, iterations
$I$, elite count $k$, momenta $(\beta_\mu, \beta_\sigma)$, init std
$\sigma_0$, displacement bound $\rho_{\max}$, distance
$d(\cdot,\cdot)$, world size $G$, per-view weights $\{w_v\}$
\State $\mu_{0:H-1} \gets$ warm start (shift previous plan by
$L_{\mathrm{exec}}$); $\sigma_{0:H-1} \gets \sigma_0$
\For{$i = 1,\dots,I$}
  \State $\mathcal{S} \gets \emptyset$
  \For{$m = 1, \dots, M$} \Comment{micro-batches per GPU; sharded across world}
    \State Draw $N_{\mathrm{samp}}/M$ translation samples
    $u^{(n)}_h \in \mathbb{R}^3$ via the spherical
    parameterisation of \Cref{appx:planning-spherical}
    \State Draw rotation/gripper coords $\sim \mathcal{N}(0,I)$
    when active; concatenate into $\tilde a^{(n)}_h \in \mathbb{R}^{d_a}$
    \State $a^{(n)}_h \gets \mu_h + \sigma_h \odot \tilde a^{(n)}_h$;
    project translation block onto $\ell_2$-ball of radius
    $\rho_{\max}$ (\Cref{appx:planning-ball})
    \State $\mathrm{state} \gets \mathrm{repeat}(z_{t-L_{\mathrm{ctx}}+1:t}, N_{\mathrm{samp}}/M)$ \Comment{share KV cache}
    \For{$h = 0,\dots,H-1$}
      \State $\hat z^{(n)}_{t+h+1}, \mathrm{state} \gets g_\theta(\mathrm{state}, a^{(n)}_h)$ for all $n$
    \EndFor
    \State $R^{(n)} \gets -\sum_v w_v\, d\!\big(\hat z^{(v,n)}_{t+H},\, z^{(v)}_g\big)$ \Comment{final-step goal distance, \Cref{eq:plan}}
    \State $\mathcal{S} \gets \mathcal{S} \cup \{(a^{(n)}, R^{(n)})\}$
  \EndFor
  \If{$G>1$}
    \State \Call{AllGather}{$\mathcal{S}$} across $G$ ranks, yielding pool of size $N_{\mathrm{samp}}\times G$
  \EndIf
  \State Select elites $\mathcal{E} \subset \mathcal{S}$: top-$k$ by $R$
  \State $\hat\mu_h, \hat\sigma_h \gets \mathrm{mean}, \mathrm{std}$ of $\mathcal{E}$ at step $h$
  \State $\mu_h \gets (1\!-\!\beta_\mu)\hat\mu_h + \beta_\mu \mu_h;\;
         \sigma_h \gets (1\!-\!\beta_\sigma)\hat\sigma_h + \beta_\sigma \sigma_h$
\EndFor
\State \Return $a^\star_{0:H-1} \gets \mu_{0:H-1}$
\end{algorithmic}
\end{algorithm}

\subsection{Distributed elite selection}\label{appx:planning-distributed}

The planner is sharded data-parallel across $G$ GPUs: every rank draws its own $N_{\mathrm{samp}}$ samples per refinement iteration and rolls them through its replica of $g_\theta$, using the same shared KV cache trick along the sample axis as on a single GPU. The per-rank rewards and the matching action tensors are then \textsc{AllGather}-ed (line~17 of \Cref{alg:cem-full}) so that every rank sees the same global pool of size $N_{\mathrm{samp}}\times G$ and can pick the top-$k$ elites independently and consistently. Because this elite set is selected from the global pool, the elite count $k$ is decoupled from the per-GPU sample count: in our DROID configuration we use $N_{\mathrm{samp}}{=}16$ samples per GPU but $k{=}20$ elites, which would be ill-defined on a single rank but is in fact a $k{=}20$ out of a pool of $N_{\mathrm{samp}}\!\cdot\!G$ ($\geq 256$ in our deployments).

\subsection{Spherical translation sampling}\label{appx:planning-spherical}

Within each refinement iteration the proposal mean and standard deviation are per-step Gaussian, but we deviate from a vanilla isotropic Gaussian for the 3-D translation block. Following V-JEPA~2-AC \citep{assran2025vjepa2}, we draw the unit translation sample $\tilde u \in \mathbb{R}^3$ from the joint density
\begin{equation}
\label{eq:spherical-sampling}
r \sim \mathrm{Unif}[0,1], \quad
\theta_1 \sim \mathrm{Unif}[0, 2\pi), \quad
\theta_2 \sim \mathrm{Unif}[0, 2\pi),
\end{equation}
\begin{equation}
\tilde u_x = r \cos\theta_1 \cos\theta_2, \;
\tilde u_y = r \cos\theta_1 \sin\theta_2, \;
\tilde u_z = r \sin\theta_1,
\end{equation}
which is then scaled by the per-axis std and shifted by the per-step mean before the $\ell_2$-ball projection of \Cref{appx:planning-ball}. The motivation is twofold. First, with $r$ uniform on $[0,1]$ and angles uniform on $[0,2\pi)$, the distribution puts noticeable mass at small $\|\tilde u\|$ (because the radial coordinate is uniform rather than thrice-folded), which corresponds to ``stay close to the warm-started mean'' and stabilizes CEM in regimes where the previous plan is already near-optimal. Second, decoupling radius from angles makes the proposal density easy to clip in norm without distorting the angular distribution: the subsequent projection onto an $\ell_2$-ball only renormalizes samples whose radius exceeds $\rho_{\max}$, leaving the angular content untouched. Rotation and gripper dimensions, when active, are drawn from a standard isotropic Gaussian and treated identically in the subsequent affine warp by $(\mu_h, \sigma_h)$.

\subsection{Translation $\ell_2$-ball projection}\label{appx:planning-ball}

After the affine warp $a^{(n)}_h = \mu_h + \sigma_h \odot \tilde a^{(n)}_h$, the translation block is projected onto an $\ell_2$-ball of radius $\rho_{\max}$ that bounds the per-step end-effector displacement:
\begin{equation}
\label{eq:l2-ball-clip}
a^{(n)}_h{:}3 \;\leftarrow\;
\begin{cases}
a^{(n)}_h{:}3 & \text{if } \|a^{(n)}_h{:}3\|_2 \le \rho_{\max}, \\
\rho_{\max}\,\dfrac{a^{(n)}_h{:}3}{\|a^{(n)}_h{:}3\|_2} & \text{otherwise}.
\end{cases}
\end{equation}
The projection enforces a per-step end-effector displacement bound that respects the robot's velocity limits, regardless of how aggressively an unlucky elite refit inflated the CEM covariance. We use $\rho_{\max}{=}0.05$\,m on DROID and $\rho_{\max}{=}0.10$\,m on RoboCasa, matching the per-step delta scales of the corresponding demonstration corpora. Because the projection only fires outside the ball, it leaves the (warm-started) mean direction unbiased and, combined with \Cref{eq:spherical-sampling}, keeps the angular content close to uniform after clipping. The gripper dimension, when active, is element-wise clipped to $[-\rho_{\mathrm{gripper}},\rho_{\mathrm{gripper}}]$ rather than projected, since it is a scalar with a different physical meaning.

\subsection{Multi-view goal handling}\label{appx:planning-multiview}

When the world model is conditioned on multiple cameras, the computational cost in \Cref{eq:plan} is summed across views with configurable per-view weights $w_v$, re-using the per-block view conditioning of \Cref{appx:arch-rope-view} so that goal patches are compared at the correct viewpoint. Goal images that are unavailable for some views are handled by zeroing the corresponding $w_v$ at runtime, which lets us re-use the same multi-view checkpoint for single-camera DROID episodes without any retraining.

\section{Robotic Setup}\label{appx:robotic-setup}

\subsection{Franka: DROID Platform}\label{appx:robotic-setup:franka}
Our real-robot experiments run on a Franka Panda arm with a parallel-jaw gripper, controlled through
the original DROID software stack~\citep{khazatsky2024droid}. Deployment is split across two machines.
A \emph{robot-control server} runs on the robot PC and exposes the arm and cameras through a simple
TCP/ZMQ request--reply interface; the \emph{planner client} runs on a GPU cluster and connects to this
server remotely. This separation lets the planner marshal a large number of GPUs for parallel
world-model inference --- the CEM optimizer samples and scores thousands of candidate rollouts per
step across the cluster --- while a single client process is the only one that exchanges observations
and actions with the physical robot (\Cref{appx:robotic-setup:franka-planning}).

The scene is observed by two cameras: one ZED camera mounted on the wrist and one Intel RealSense
camera placed externally to the left of the workspace (unlike the multi-ZED upstream DROID rig, we do
not use a third camera). The wrist ZED records at $720{\times}1280$ and the external RealSense at
$480{\times}640$; both streams are converted to RGB and resized to each agent's native input
resolution ($240{\times}320$ for the RoboJEPA models) before encoding. At deployment RoboJEPA uses the
resulting two views --- external (left) and wrist.

For RoboJEPA we drive the robot with \emph{blocking end-effector control} rather than a fixed-rate
stream: each environment step commands an end-effector delta pose and then waits until the arm and
gripper reach the commanded target before the next observation is returned, so the loop is paced by
motion completion rather than a fixed control frequency. Although the underlying environment supports
the full DROID action interface (7-DoF Cartesian and 8-DoF joint spaces, in velocity or position
mode), our planner acts in a reduced space: it plans only $3$D end-effector translation together with
a $1$D gripper command, holding the end-effector orientation fixed. The proprioceptive state we read
back is the end-effector pose and gripper opening (with joint angles also available). At each step the
planner encodes the current views and state into the world-model latent space, chooses an action by
imagining short rollouts and scoring them against a goal image (\Cref{subsec:planning}), executes it,
and repeats; each episode begins by homing the arm and recording the goal observation that defines the
task. The full planning configuration is given in \Cref{appx:robotic-setup:franka-planning}.

For baseline policies such as $\pi_{0.5}$~\citep{black2025pi05} we instead use the standard DROID
setup: non-blocking control at $15$\,Hz in the joint-velocity action space, matching the interface
these models were trained for.

\begin{figure}[H]
\centering
\setlength{\tabcolsep}{2pt}
\begin{tabular}{@{}c c @{\hspace{2.5em}} c@{}}
 & \textbf{Goal} & \textbf{Start (initial)} \\[3pt]
\rotatebox{90}{\parbox{0.28\linewidth}{\centering\small Side camera}} &
  \includegraphics[width=0.36\linewidth]{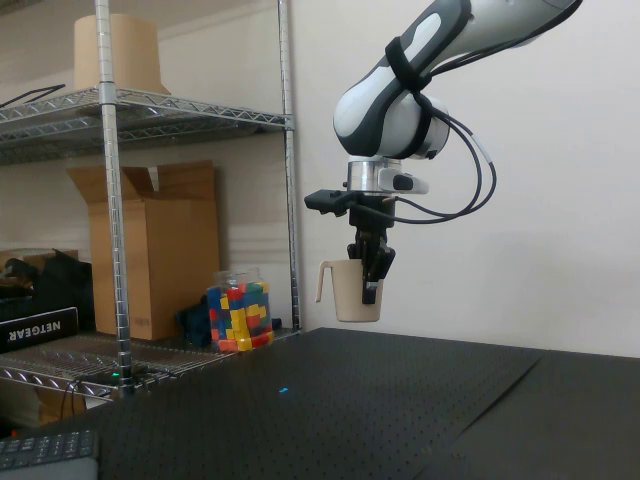} &
  \includegraphics[width=0.36\linewidth]{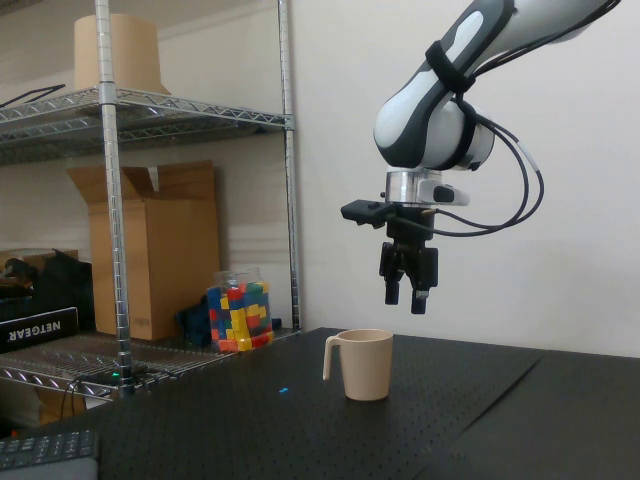} \\[4pt]
\rotatebox{90}{\parbox{0.20\linewidth}{\centering\small Wrist camera}} &
  \includegraphics[width=0.36\linewidth]{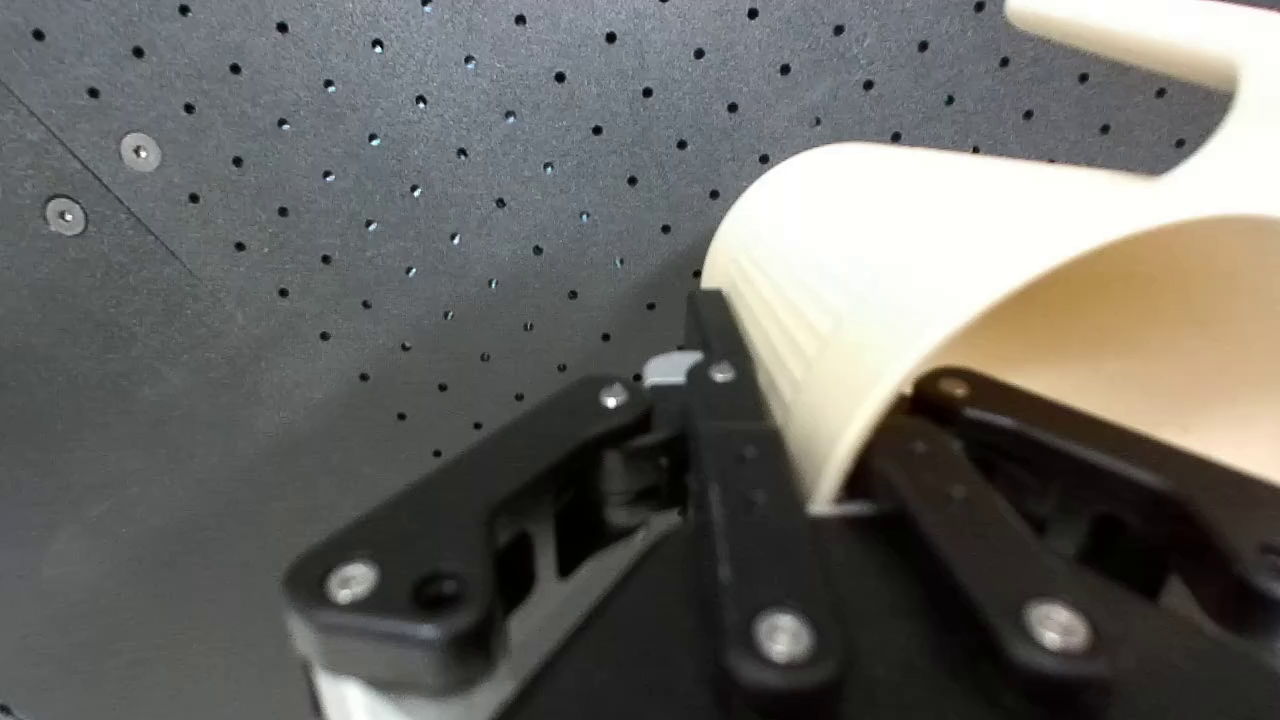} &
  \includegraphics[width=0.36\linewidth]{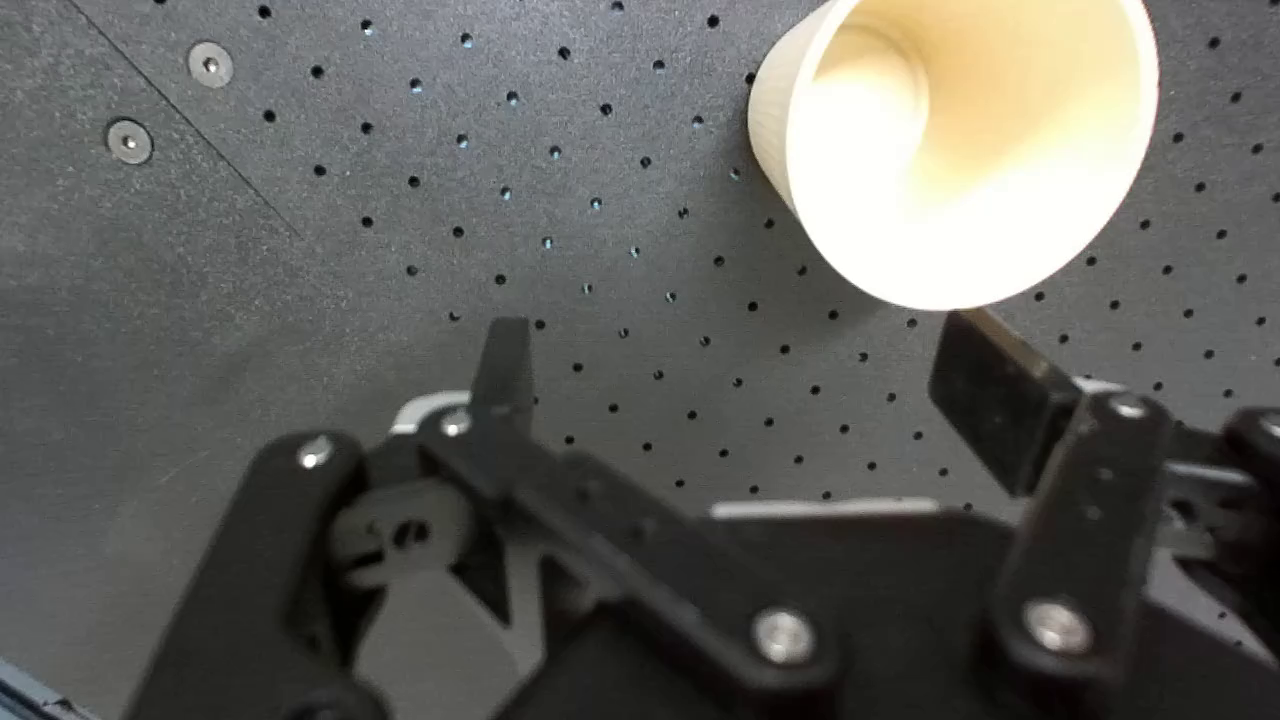} \\
\end{tabular}
\caption{\textbf{Real-world Franka \emph{lift the cup} task.} A goal image (left) and the initial
observation (right) for one episode on the DROID/Franka platform, shown from both cameras: the
external side camera (top) and the wrist-mounted camera (bottom). The task is specified purely by the
goal image, in which the cup has been lifted off the table; from the start state the planner drives the
world model toward these goal features. The two camera views are the inputs the world model is
conditioned on at deployment.}
\label{fig:droid-lift}
\end{figure}

\subsection{Real-robot tasks and evaluation protocol}\label{appx:robotic-setup:franka-tasks}

\paragraph{Tasks and success criteria.}
We evaluate three tasks on the Franka/DROID platform: Grasp, Object Lift, and Pick and Place.
Grasp requires the robot to grasp the object. Object Lift requires the robot to hold the object
lifted in the air until the end of the episode. Pick and Place requires the object to end up at
the target position, regardless of how many times it was dropped during the episode. These
completion criteria do not imply one another: success on Pick and Place does not require
satisfying the Object Lift criterion. Goal and initial observations for all three tasks are shown in
\Cref{appx:robotic-setup:deployment-examples}.

\paragraph{Evaluation protocol.}
For each model--task combination, we run $30$ episodes for RoboJEPA and $50$ episodes for each
VLA baseline. We evaluate RoboJEPA models ranging from $22$M to $8$B parameters, using the
checkpoint trained with the most compute at each size. All RoboJEPA models use the same planning
hyperparameters across tasks, including the planning horizon. Hardware and controller details are provided in
\Cref{appx:robotic-setup:franka}, and the planning configuration is given in
\Cref{appx:robotic-setup:franka-planning}.

For RoboJEPA, we generate a goal image by executing a predefined robot motion and manually
placing the object in the grasping position so the robot can interact with it. Only the final
image is provided as the goal for closed-loop planning; no intermediate image subgoals are used.
The VLA baselines receive text instructions. Neither RoboJEPA nor the baselines receive
task-specific fine-tuning.

\paragraph{Scoring and reported metrics.}
We manually score each episode according to the task-specific completion criteria. For each
model--task combination, we report task progress as the mean episode score and success rate as
the fraction of episodes with $100\%$ progress. The per-model values are reported in
\Cref{tab:real-world-capabilities}.

\begin{table}[h]
\caption{Real-world robotic capabilities on the Franka/DROID platform. Task progress and success rate on the grasp, object-lift, and pick-and-place tasks. VLA baselines ($\pi_0$-FAST and $\pi_{0.5}$) are conditioned on a text instruction, whereas all RoboJEPA models are conditioned on a single image goal. }
\label{tab:real-world-capabilities}
\centering
\small
\resizebox{\textwidth}{!}{%
\begin{tabular}{ll cc cc cc}
\toprule
\multirow{2}{*}{Model} & \multirow{2}{*}{Goal specification} & \multicolumn{2}{c}{Grasp} & \multicolumn{2}{c}{Object Lift} & \multicolumn{2}{c}{Pick and Place} \\
\cmidrule(lr){3-4}\cmidrule(lr){5-6}\cmidrule(lr){7-8}
 & & Task Progress & Success Rate & Task Progress & Success Rate & Task Progress & Success Rate \\
\midrule
\multicolumn{8}{c}{\textit{Model finetuned on DROID, no in-domain finetuning}} \\
\midrule
$\pi_0$-FAST & \multirow{2}{*}{text instruction} & $35\%$ & $22\%$ & $20\%$ & $12\%$ & $55\%$ & $40\%$ \\
$\pi_{0.5}$  &                                            & $25\%$ & $5\%$ & $12\%$ & $0\%$ & $69\%$ & $53\%$ \\
\midrule
\multicolumn{8}{c}{\textit{Pretrained models, no DROID finetuning, no in-domain finetuning, a unified planner for all tasks}} \\
\midrule
RoboJEPA-22M  & \multirow{8}{*}{single goal image} & $30\%$ & $30\%$ & $34\%$ & $2\%$ & $30\%$ & $17\%$ \\
RoboJEPA-50M  &                                            & $30\%$ & $30\%$ & $51\%$ & $8\%$ & $27\%$ & $13\%$ \\
RoboJEPA-100M &                                            & $27\%$ & $27\%$ & $72\%$ & $20\%$ & $20\%$ & $10\%$ \\
RoboJEPA-300M &                                            & $37\%$ & $37\%$ & $52\%$ & $18\%$ & $43\%$ & $17\%$ \\
RoboJEPA-1B   &                                            & $50\%$ & $50\%$ & $53\%$ & $24\%$ & $38\%$ & $10\%$ \\
RoboJEPA-2B   &                                            & $43\%$ & $43\%$ & $60\%$ & $40\%$ & $47\%$ & $31\%$ \\
RoboJEPA-4B   &                                            & $60\%$ & $60\%$ & $54\%$ & $30\%$ & $49\%$ & $21\%$ \\
RoboJEPA-8B   &                                            & $67\%$ & $67\%$ & $65\%$ & $50\%$ & $42\%$ & $27\%$ \\
\bottomrule
\end{tabular}%
}
\end{table}

\subsection{Real robot planning hyperparameters}\label{appx:robotic-setup:franka-planning}
On the DROID/Franka platform the world model is deployed with the closed-loop CEM planner of
\Cref{subsec:planning} (full algorithm in \Cref{alg:cem-full}). The planner searches only over $3$D
end-effector translation and a $1$D gripper command (rotation frozen), scores imagined rollouts by
per-token $\ell_1$ distance to the goal features in $\mathrm{LN}_0$ space, and applies
$L_{\mathrm{exec}}$ actions per replan before re-observing. \Cref{tab:droid-planning-hparams} lists the
configuration used for all real-robot experiments.

\begin{table}[H]
\caption{Closed-loop CEM planning hyperparameters for the DROID/Franka real-robot deployment
(\Cref{subsec:planning}, \Cref{alg:cem-full}). Elites are selected from the pooled
$N_{\mathrm{samp}}\!\cdot\!G$ samples gathered across the $G$ planner GPUs, so $k$ may exceed the
per-GPU sample count. The planner acts in a $3$D-translation $+$ $1$D-gripper space with the
end-effector orientation held fixed.}
\label{tab:droid-planning-hparams}
\centering
\small
\begin{tabular}{lll}
\toprule
Hyperparameter & Symbol & Value \\
\midrule
\multicolumn{3}{l}{\emph{Horizon and closed loop}} \\
Planning horizon (predictor steps)          & $H$                          & $5$ \\
Context length (encoder steps)              & $L_{\mathrm{ctx}}$           & $2$ \\
Execution prefix (actions per replan)       & $L_{\mathrm{exec}}$          & $2$ \\
Environment steps per episode               & ---                          & $20$ \\
\midrule
\multicolumn{3}{l}{\emph{CEM search}} \\
CEM iterations (cold / warm)                & $I_0 / I_{\mathrm{step}}$    & $15$ / $10$ \\
Samples per iteration (global)              & $N_{\mathrm{samp}}\!\cdot\!M\!\cdot\!G$ & $4096$ \\
Elite count, top-$k$ (post-AllGather)       & $k$                          & $20$ (pool $N_{\mathrm{samp}}\!\cdot\!G$) \\
Computational cost                          & $d(\cdot,\cdot)$             & per-token $\ell_1$ in $\mathrm{LN}_0$ space \\
\midrule
\multicolumn{3}{l}{\emph{Translation search (3D)}} \\
Init std                                    & $\sigma_0$                   & $0.15$ \\
Max-norm ($\ell_2$-ball radius)             & $\rho_{\max}$                & $0.15$ \\
Mean / std momentum                         & $\beta_\mu / \beta_\sigma$   & $0.15$ / $0.70$ \\
\midrule
\multicolumn{3}{l}{\emph{Gripper search (1D)}} \\
Init std                                    & ---                          & $0.80$ \\
Max magnitude                               & $\rho_{\mathrm{gripper}}$    & $0.99$ \\
Mean / std momentum                         & ---                          & $0.15$ / $0.45$ \\
\midrule
\multicolumn{3}{l}{\emph{Observations and execution}} \\
Views fed to the world model                & ---                          & left ($0$), wrist ($2$) \\
Per-view reward weights                     & $\{w_v\}$                    & $\{1.0,\, 0.0\}$ (goal scored on the left view) \\
Input resolution                            & ---                          & $240{\times}320$ \\
Action substeps                             & ---                          & $1$ \\
Predictor precision                         & ---                          & AOT-compiled bf16 \\
\bottomrule
\end{tabular}
\end{table}

\subsection{RoboCasa Custom Environments}\label{appx:robotic-setup:robocasa}
\newcommand{\robocasataskfig}[3]{%
\begin{figure}[htbp]
\centering
\setlength{\tabcolsep}{2pt}
\begin{tabular}{@{}c c @{\hspace{2.5em}} c@{}}
 & \textbf{Goal} & \textbf{Start (initial)} \\[3pt]
\rotatebox{90}{\parbox{0.25\linewidth}{\centering\small Side camera}} &
  \includegraphics[width=0.34\linewidth]{figures/robocasa_tasks/#1/goal_side.png} &
  \includegraphics[width=0.34\linewidth]{figures/robocasa_tasks/#1/start_side.png} \\[4pt]
\rotatebox{90}{\parbox{0.25\linewidth}{\centering\small Wrist camera}} &
  \includegraphics[width=0.34\linewidth]{figures/robocasa_tasks/#1/goal_wrist.png} &
  \includegraphics[width=0.34\linewidth]{figures/robocasa_tasks/#1/start_wrist.png} \\
\end{tabular}
\caption{#2}
\label{#3}
\end{figure}%
}
Our simulated evaluations use four custom tasks built on top of RoboCasa kitchen
scenes~\citep{nasiriany2024robocasa, nasiriany2026robocasa365} with the mobile-base \texttt{PandaOmron} (Franka Panda) robot.
The four tasks share the same common controller. We drive the robot with a blocking controller that
we implement on top of the simulator: each planner action specifies a target end-effector offset,
which the environment servos toward with operational-space control until it converges (or a step cap
is reached) before returning the next observation. We further extend the controller to drive the
commanded end-effector \emph{rotation} to convergence: rather than the default operational-space
controller, which does not guarantee the orientation reaches its target, our controller feeds back the
world-frame rotation error at every sub-step and only terminates once both the position and rotation
errors fall below tolerance. We found this rotation convergence to be important for correctly matching
the wrist-camera view during planning, since the planner scores candidates by a distance induced by
both the side-view and the wrist-view predictions and the wrist camera is rigidly mounted to the
end-effector, so any residual orientation drift would misalign the wrist view. RoboCasa exposes a
large collection of kitchens
split into training and evaluation scenes; the world model is trained on the training kitchens and all
four tasks are run in the held-out evaluation kitchens. Because the environment is fully simulated, we
can define both the goal image and the reward directly inside it. Every episode is instantiated from
up to ten held-out target kitchens (one per layout/style pair), with non-structural countertop appliances
near the workspace hidden to reduce clutter. Observations are rendered at $480{\times}640$ from the
agent-view and, when enabled, wrist cameras. In every task the reward is the normalized progress toward
the goal, $r_t = \max\!\big(0,\, 1 - \lVert x_t - x_{\mathrm{goal}} \rVert / d_0\big)$, where $d_0$ is
the initial goal distance.

\paragraph{Reach.} The robot must move its end-effector from a randomly sampled start pose to a
randomly sampled goal pose in the kitchen. We remove all objects from the scene so that nothing
occludes the path from the start to the goal. Start and goal are drawn uniformly within a small ball
around the robot's home pose and rejection-sampled so that they lie a fixed distance apart
($0.25$--$0.35$\,m) and are reachable without a large reorientation of the end-effector. This task
isolates the world model's ability to predict free-space end-effector motion across kitchens.
\Cref{fig:robocasa-reach} shows example goal and initial observations.

\robocasataskfig{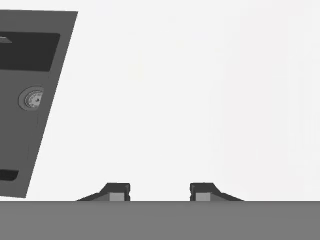}{\textbf{RoboCasa Reach.} Goal and initial observations for one Reach episode,
from the external side camera (top) and the wrist camera (bottom). The goal image (left) specifies the
target end-effector pose in an emptied kitchen; the planner drives the arm from the start (right)
toward the goal.}{fig:robocasa-reach}

\paragraph{ObstacleReach.} A harder variant of Reach: one appliance (by default a coffee machine) is
kept in the scene and repositioned directly in front of the robot, and the start and goal are
constrained to lie on opposite sides of it. The end-effector must therefore arc \emph{around} the
obstacle rather than travel in a straight line, and the reachable distance band is widened
($0.25$--$0.50$\,m). It tests whether the planner can find collision-free trajectories through the
world model. \Cref{fig:robocasa-obstacle} shows an example episode.

\robocasataskfig{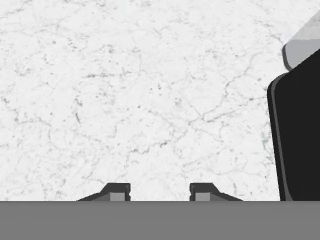}{\textbf{RoboCasa ObstacleReach.} Goal and initial observations for one
ObstacleReach episode. An appliance placed directly in front of the robot forces the end-effector to
arc around it, so the straight-line path from the start (right) to the goal (left) is blocked.}{fig:robocasa-obstacle}

\paragraph{ObjectReach.} Here the robot must perform a reach while holding an object. To set this up,
we first grasp the object with a heuristic scripted agent: a small Python controller that executes a
top-down grasp (approach, descend, close, and lift) inside the simulator. Running the grasp in
simulation ensures the object is held in a physically correct way before planning begins. The world
model then has to move the held object to a randomly sampled goal position, and the reward is the
normalized progress of the \emph{object} toward its goal rather than of the end-effector. The object is
one of a banana, a can, or a mug, which differ in shape and physical properties (friction and grasp
offsets are tuned per object). \Cref{fig:robocasa-object-banana,fig:robocasa-object-can,fig:robocasa-object-mug} show examples for the three objects.

\robocasataskfig{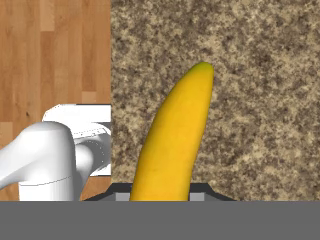}{\textbf{RoboCasa ObjectReach (banana).} Goal and initial observations
after the scripted grasp: from the start (right) the robot must carry the held banana to the goal pose
(left), shown from the side (top) and wrist (bottom) cameras.}{fig:robocasa-object-banana}

\robocasataskfig{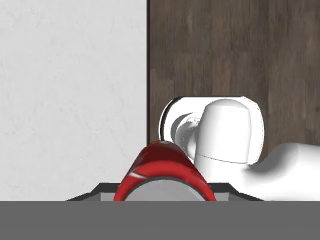}{\textbf{RoboCasa ObjectReach (can).} As in \Cref{fig:robocasa-object-banana}
for the can object.}{fig:robocasa-object-can}

\robocasataskfig{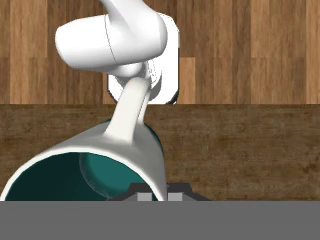}{\textbf{RoboCasa ObjectReach (mug).} As in \Cref{fig:robocasa-object-banana}
for the mug object.}{fig:robocasa-object-mug}

\paragraph{ObjectPush.} The robot must push the object across the counter by
contact from a randomly sampled start position to a randomly sampled goal position. The gripper is
held closed so that it acts as a solid pusher, and the arm is driven by translation deltas, so the
object moves only through contact rather than being grasped. Start and goal are sampled a fixed
distance apart ($0.10$--$0.15$\,m) within a small disk on the counter; because forward pushes (along
the robot's reach direction) are far more reliable than lateral ones, the push direction is restricted
to a $60^\circ$ cone around the forward axis. As in ObjectReach, the reward is the normalized progress
of the \emph{object} toward its goal. This task probes whether the world model captures non-prehensile,
contact-rich dynamics. \Cref{fig:robocasa-push} shows example goal and initial observations.

\robocasataskfig{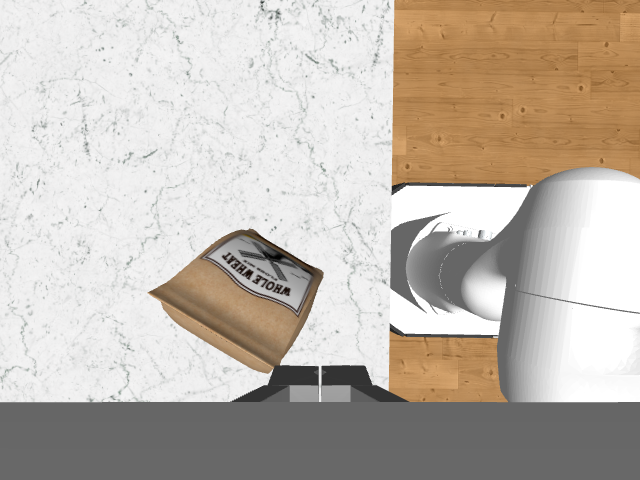}{\textbf{RoboCasa Push.} Goal and initial observations for one object-push
episode, shown from the side (top) and wrist (bottom) cameras. The robot must push the object across
the counter from its initial position (right) to the goal position (left).
Note that the goal pose and the initial pose of the robot are the same, the only thing that changes is the object position.}{fig:robocasa-push}

\subsection{RoboCasa planning hyperparameters}\label{appx:robotic-setup:robocasa-planning}
All four RoboCasa tasks are solved with the same closed-loop CEM planner (\Cref{subsec:planning},
\Cref{alg:cem-full}). Across the four tasks the only CEM hyperparameter we vary is the predictor
horizon $H$; every other CEM setting is shared (\Cref{tab:robocasa-planning-hparams}, top). Push uses the same planning configuration as ObstacleReach. As in the
DROID configuration (\Cref{tab:droid-planning-hparams}) the planner scores rollouts by per-token
$\ell_1$ distance to the goal features and re-plans after every executed action. Reach and
ObstacleReach and Push search over $3$D end-effector translation, while ObjectReach
first grasps an object with a scripted agent and then additionally searches the
gripper. The tasks otherwise differ in their environment configuration
(\Cref{tab:robocasa-planning-hparams}, bottom).

\begin{table}[H]
\caption{Planning and environment configuration for the four RoboCasa tasks
(\Cref{appx:robotic-setup:robocasa}). The table is split into the planning hyperparameters (top) and
the environment parameters (bottom). Among the planning hyperparameters, only the predictor horizon
$H$ varies across tasks; the rest are shared. The effective CEM sample budget is $1024$ candidates per
iteration for every task (distributed across the planner GPUs), from which the top-$k$ elites are
selected.}
\label{tab:robocasa-planning-hparams}
\centering
\small
\resizebox{\textwidth}{!}{%
\begin{tabular}{llcccc}
\toprule
 & Symbol & Reach & ObstacleReach & ObjectReach & Push \\
\midrule
\multicolumn{6}{l}{\emph{Planning hyperparameters}} \\
Predictor horizon                           & $H$                          & $1$               & $5$                    & $1$ & $5$ \\
CEM samples per iteration (total)           & ---                          & \multicolumn{4}{c}{$1024$} \\
CEM iterations (cold / warm)                & $I_0 / I_{\mathrm{step}}$    & \multicolumn{4}{c}{$15$ / $10$} \\
Elite count, top-$k$                         & $k$                          & \multicolumn{4}{c}{$20$} \\
Computational cost                          & $d(\cdot,\cdot)$             & \multicolumn{4}{c}{per-token $\ell_1$ in $\mathrm{LN}_0$ space} \\
Context length (encoder steps)              & $L_{\mathrm{ctx}}$           & \multicolumn{4}{c}{$2$} \\
Execution prefix (actions per replan)       & $L_{\mathrm{exec}}$          & \multicolumn{4}{c}{$1$} \\
Action substeps                             & ---                          & \multicolumn{4}{c}{$1$} \\
Translation init std                        & $\sigma_0$                   & \multicolumn{4}{c}{$0.30$} \\
Translation max-norm ($\ell_2$-ball radius) & $\rho_{\max}$                & \multicolumn{4}{c}{$0.30$} \\
Mean / std momentum                         & $\beta_\mu / \beta_\sigma$   & \multicolumn{4}{c}{$0.15$ / $0.70$} \\
Views fed to the world model                & ---                          & \multicolumn{4}{c}{left ($0$), wrist ($2$)} \\
Per-view reward weights                     & $\{w_v\}$                    & \multicolumn{4}{c}{$\{0.5,\, 0.5\}$} \\
Input resolution                            & ---                          & \multicolumn{4}{c}{$240{\times}320$} \\
Gripper init std                            & ---                          & \multicolumn{4}{c}{$0.25$} \\
Gripper max magnitude                       & $\rho_{\mathrm{gripper}}$    & \multicolumn{4}{c}{$0.3$} \\
Predictor precision                         & ---                          & bf16 (compiled)   & bf16 (eager)           & bf16 (compiled) & bf16 (eager) \\
\midrule
\multicolumn{6}{l}{\emph{Environment parameters}} \\
Environment steps per episode               & ---                          & $5$               & $18$                   & $5$ & $18$ \\
Start--goal distance                        & ---                          & $0.25$--$0.35$\,m & $0.25$--$0.50$\,m      & $0.11$--$0.26$\,m & $0.10$--$0.15$\,m \\
Manipulated object                          & ---                          & ---               & ---                    & banana / can / mug & flour bag \\
\bottomrule
\end{tabular}%
}
\end{table}

\subsection{Real-Robot Task Examples}\label{appx:robotic-setup:deployment-examples}
\begin{figure}[p]
\centering
\setlength{\tabcolsep}{2pt}
\renewcommand{\arraystretch}{0.5}
\begin{tabular}{cccc}
\textbf{Goal (side)} & \textbf{Goal (wrist)} & \textbf{Start (side)} & \textbf{Start (wrist)} \\[2pt]
\includegraphics[height=0.073\textheight]{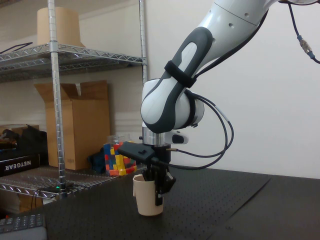} & \includegraphics[height=0.073\textheight]{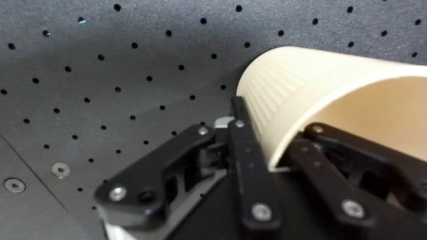} & \includegraphics[height=0.073\textheight]{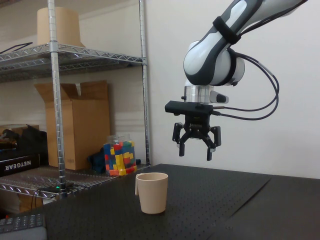} & \includegraphics[height=0.073\textheight]{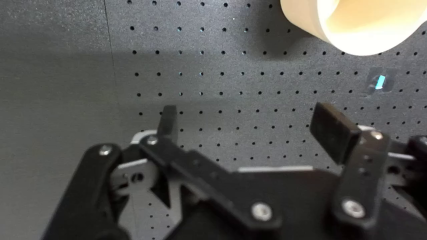} \\
\includegraphics[height=0.073\textheight]{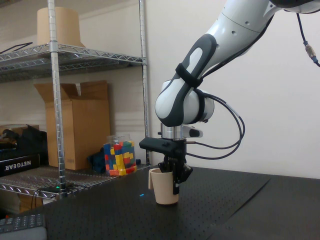} & \includegraphics[height=0.073\textheight]{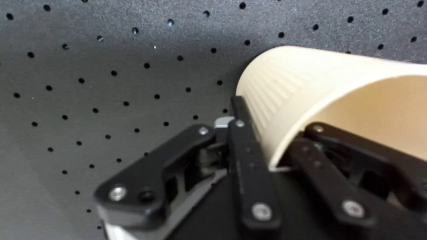} & \includegraphics[height=0.073\textheight]{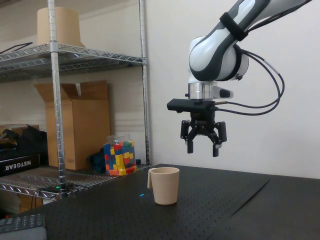} & \includegraphics[height=0.073\textheight]{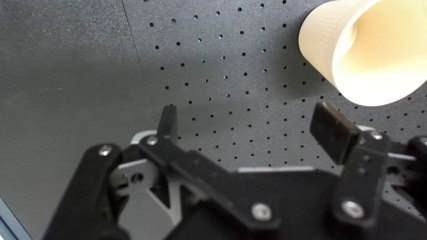} \\
\includegraphics[height=0.073\textheight]{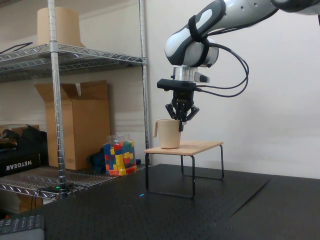} & \includegraphics[height=0.073\textheight]{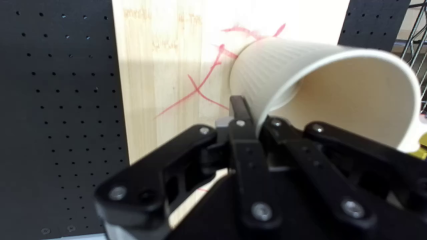} & \includegraphics[height=0.073\textheight]{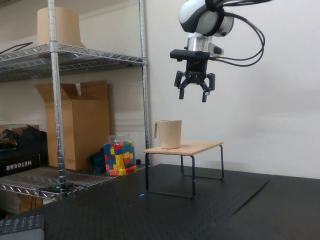} & \includegraphics[height=0.073\textheight]{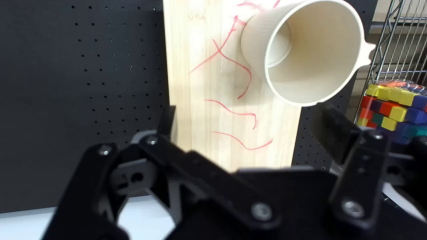} \\
\includegraphics[height=0.073\textheight]{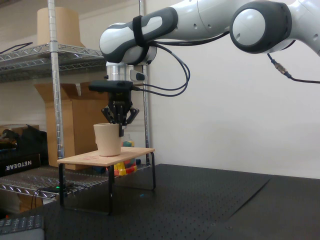} & \includegraphics[height=0.073\textheight]{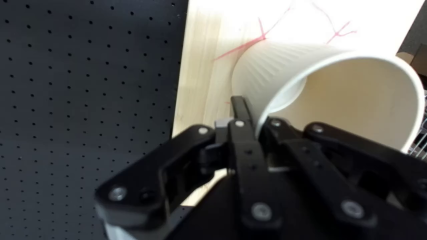} & \includegraphics[height=0.073\textheight]{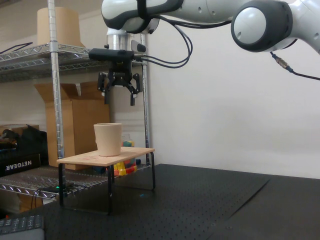} & \includegraphics[height=0.073\textheight]{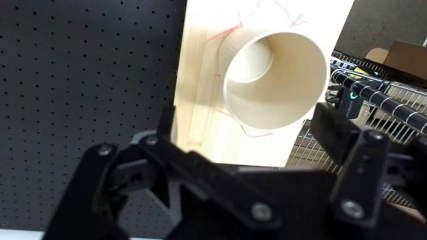} \\
\includegraphics[height=0.073\textheight]{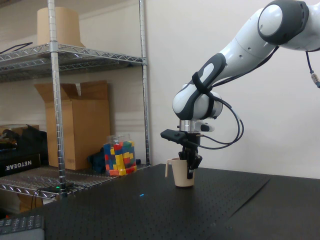} & \includegraphics[height=0.073\textheight]{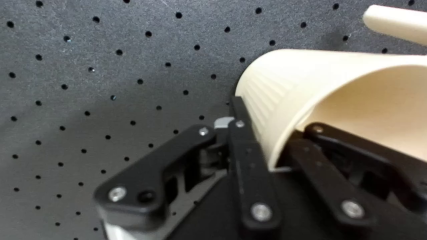} & \includegraphics[height=0.073\textheight]{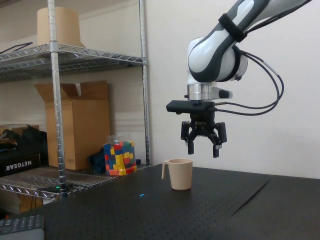} & \includegraphics[height=0.073\textheight]{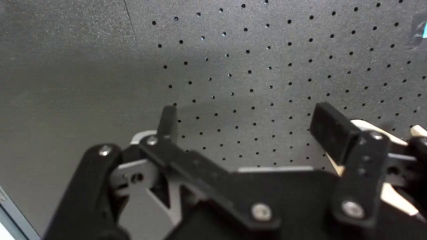} \\
\includegraphics[height=0.073\textheight]{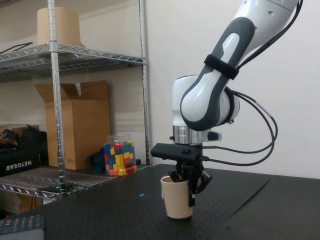} & \includegraphics[height=0.073\textheight]{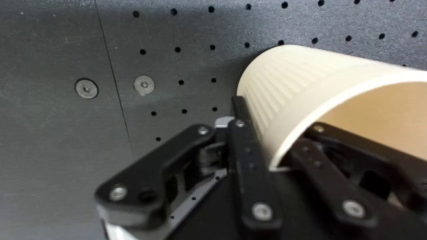} & \includegraphics[height=0.073\textheight]{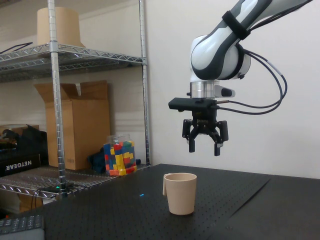} & \includegraphics[height=0.073\textheight]{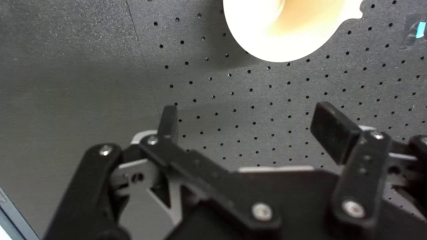} \\
\includegraphics[height=0.073\textheight]{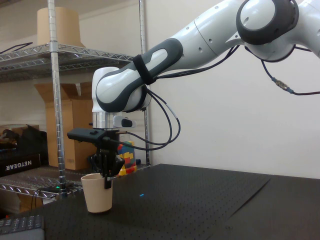} & \includegraphics[height=0.073\textheight]{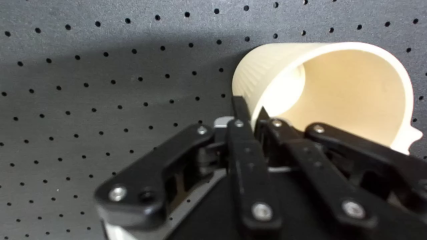} & \includegraphics[height=0.073\textheight]{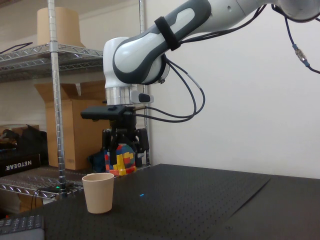} & \includegraphics[height=0.073\textheight]{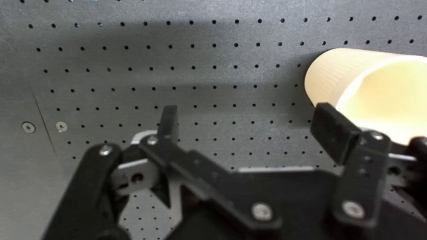} \\
\includegraphics[height=0.073\textheight]{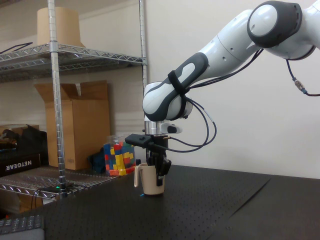} & \includegraphics[height=0.073\textheight]{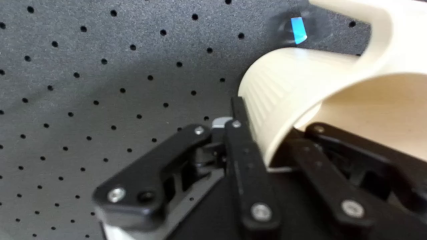} & \includegraphics[height=0.073\textheight]{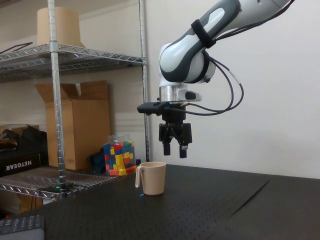} & \includegraphics[height=0.073\textheight]{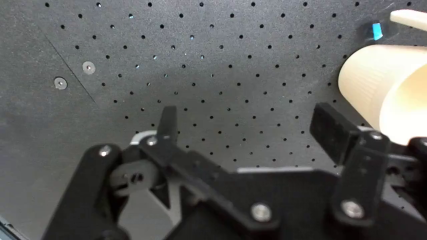} \\
\includegraphics[height=0.073\textheight]{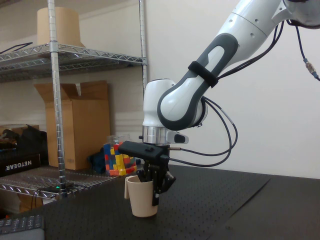} & \includegraphics[height=0.073\textheight]{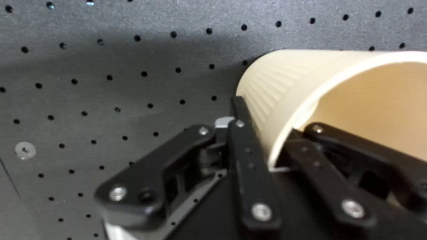} & \includegraphics[height=0.073\textheight]{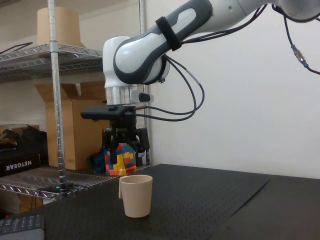} & \includegraphics[height=0.073\textheight]{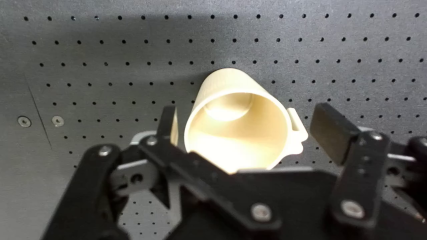} \\
\includegraphics[height=0.073\textheight]{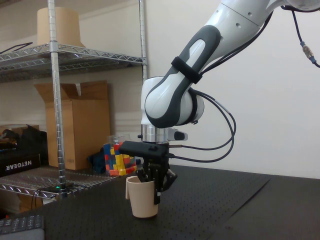} & \includegraphics[height=0.073\textheight]{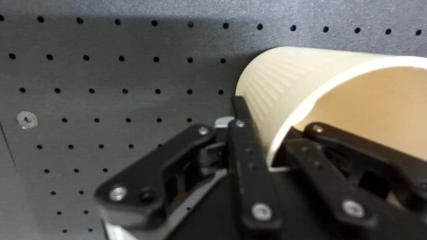} & \includegraphics[height=0.073\textheight]{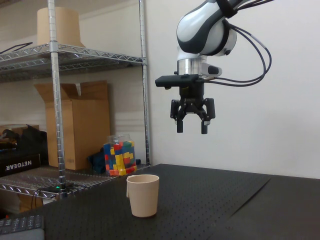} & \includegraphics[height=0.073\textheight]{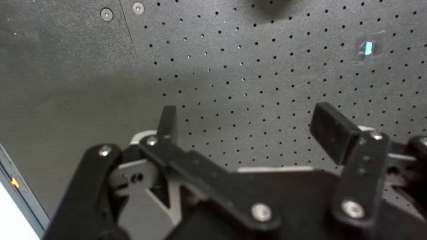} \\
\end{tabular}
\caption{\textbf{Real-robot Grasp task: goal and initial observations.} Ten episodes (rows) of the
Grasp task on the DROID/Franka platform. For each episode we show the goal image (left pair) and the
initial state (right pair), each from the external side camera and the wrist camera. The goal is
specified purely by these images; the planner drives the robot from the initial state toward the goal.}
\label{fig:deploy-grasp}
\end{figure}

\begin{figure}[p]
\centering
\setlength{\tabcolsep}{2pt}
\renewcommand{\arraystretch}{0.5}
\begin{tabular}{cccc}
\textbf{Goal (side)} & \textbf{Goal (wrist)} & \textbf{Start (side)} & \textbf{Start (wrist)} \\[2pt]
\includegraphics[height=0.073\textheight]{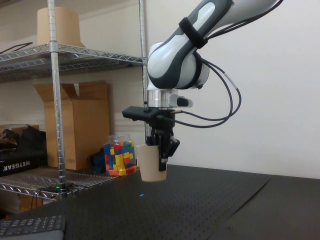} & \includegraphics[height=0.073\textheight]{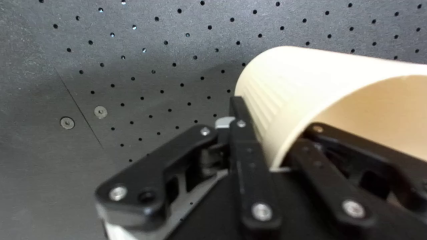} & \includegraphics[height=0.073\textheight]{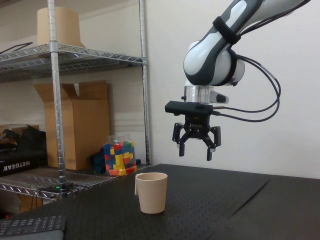} & \includegraphics[height=0.073\textheight]{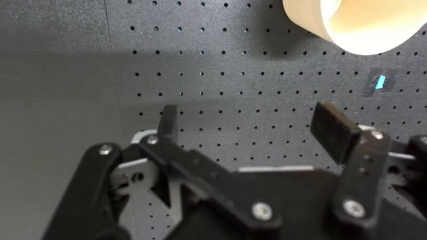} \\
\includegraphics[height=0.073\textheight]{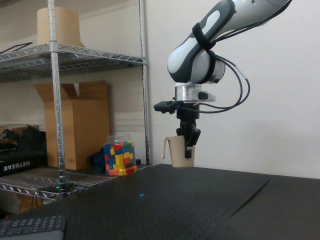} & \includegraphics[height=0.073\textheight]{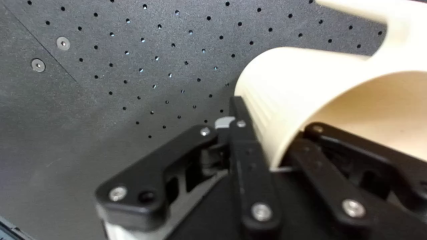} & \includegraphics[height=0.073\textheight]{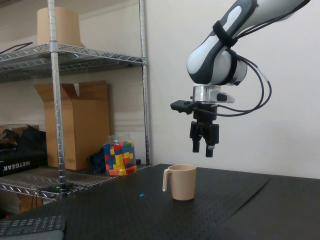} & \includegraphics[height=0.073\textheight]{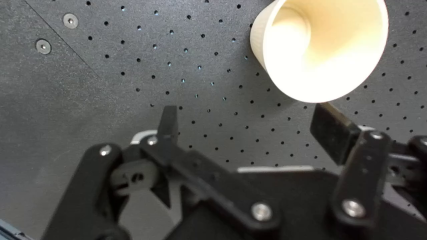} \\
\includegraphics[height=0.073\textheight]{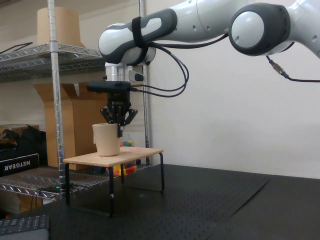} & \includegraphics[height=0.073\textheight]{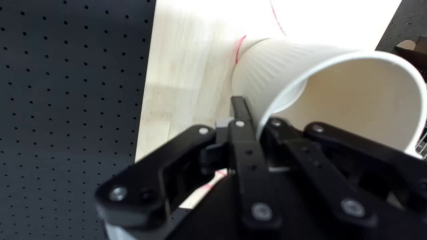} & \includegraphics[height=0.073\textheight]{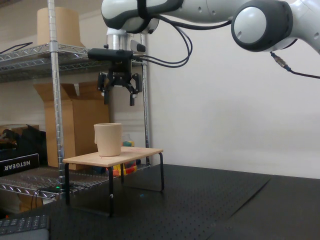} & \includegraphics[height=0.073\textheight]{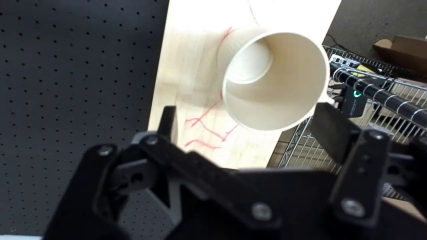} \\
\includegraphics[height=0.073\textheight]{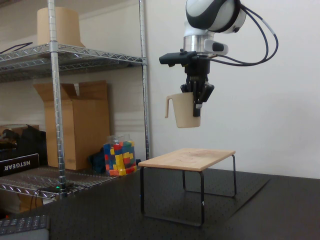} & \includegraphics[height=0.073\textheight]{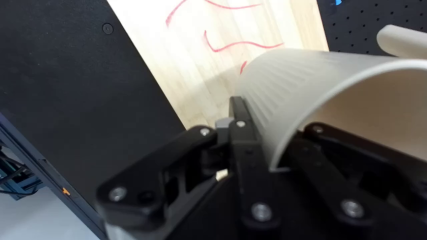} & \includegraphics[height=0.073\textheight]{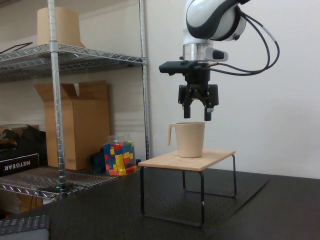} & \includegraphics[height=0.073\textheight]{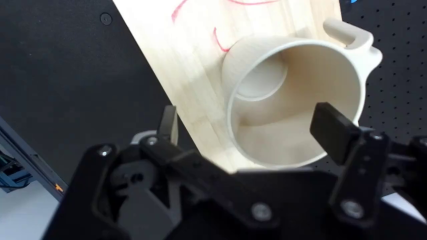} \\
\includegraphics[height=0.073\textheight]{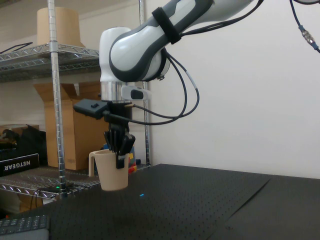} & \includegraphics[height=0.073\textheight]{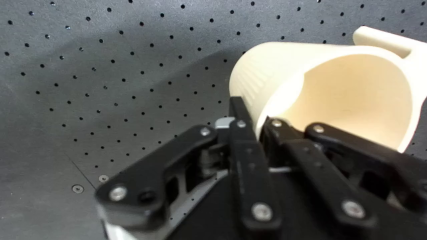} & \includegraphics[height=0.073\textheight]{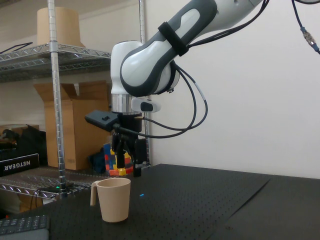} & \includegraphics[height=0.073\textheight]{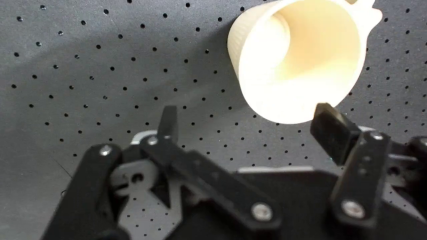} \\
\includegraphics[height=0.073\textheight]{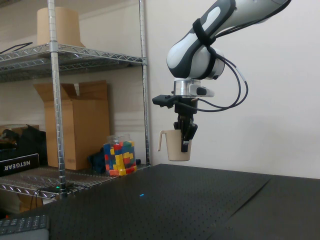} & \includegraphics[height=0.073\textheight]{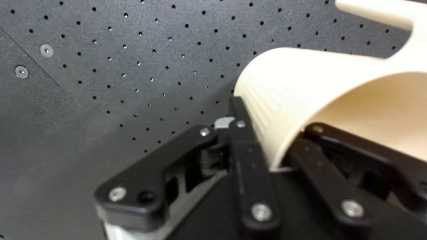} & \includegraphics[height=0.073\textheight]{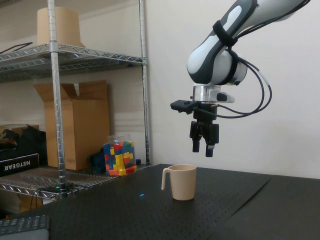} & \includegraphics[height=0.073\textheight]{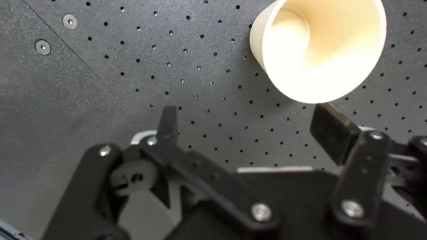} \\
\includegraphics[height=0.073\textheight]{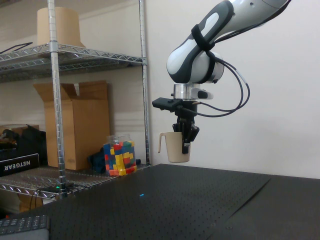} & \includegraphics[height=0.073\textheight]{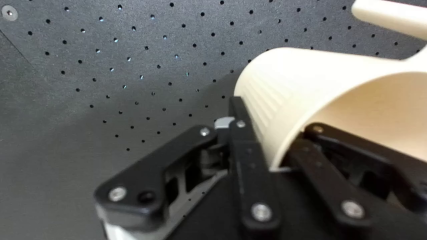} & \includegraphics[height=0.073\textheight]{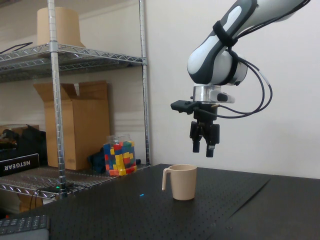} & \includegraphics[height=0.073\textheight]{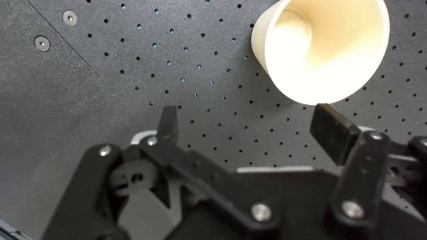} \\
\includegraphics[height=0.073\textheight]{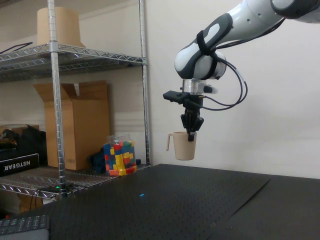} & \includegraphics[height=0.073\textheight]{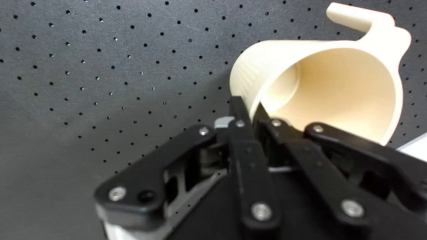} & \includegraphics[height=0.073\textheight]{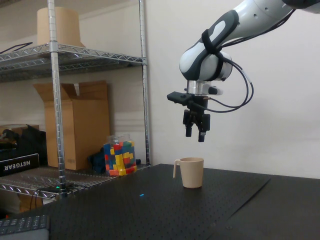} & \includegraphics[height=0.073\textheight]{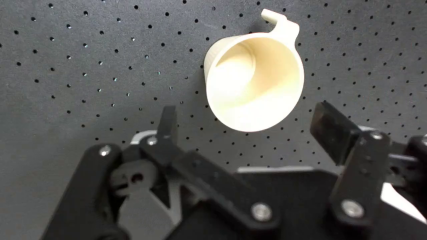} \\
\includegraphics[height=0.073\textheight]{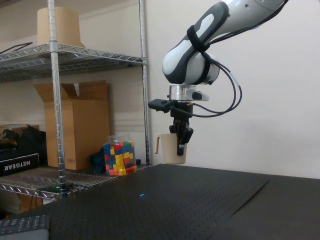} & \includegraphics[height=0.073\textheight]{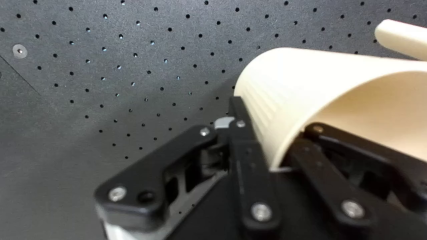} & \includegraphics[height=0.073\textheight]{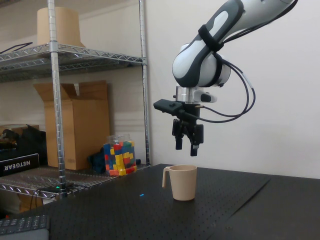} & \includegraphics[height=0.073\textheight]{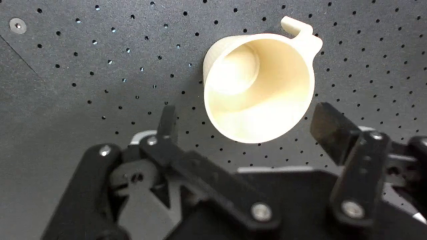} \\
\includegraphics[height=0.073\textheight]{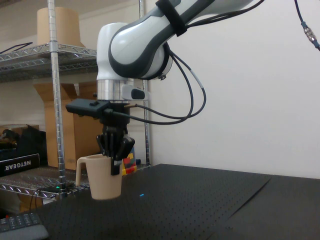} & \includegraphics[height=0.073\textheight]{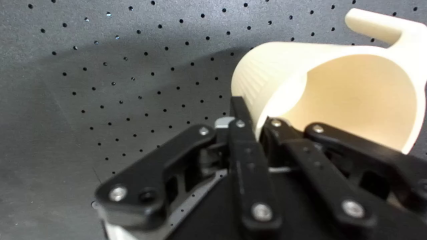} & \includegraphics[height=0.073\textheight]{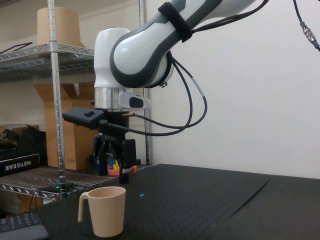} & \includegraphics[height=0.073\textheight]{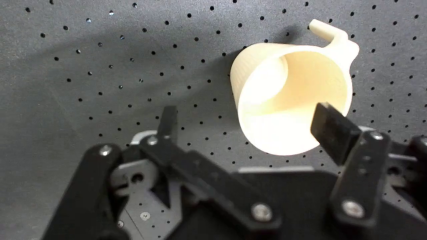} \\
\end{tabular}
\caption{\textbf{Real-robot Lift task: goal and initial observations.} Ten episodes (rows) of the
Lift task on the DROID/Franka platform. For each episode we show the goal image (left pair) and the
initial state (right pair), each from the external side camera and the wrist camera. The goal is
specified purely by these images; the planner drives the robot from the initial state toward the goal.}
\label{fig:deploy-lift}
\end{figure}

\begin{figure}[p]
\centering
\setlength{\tabcolsep}{2pt}
\renewcommand{\arraystretch}{0.5}
\begin{tabular}{cccc}
\textbf{Goal (side)} & \textbf{Goal (wrist)} & \textbf{Start (side)} & \textbf{Start (wrist)} \\[2pt]
\includegraphics[height=0.073\textheight]{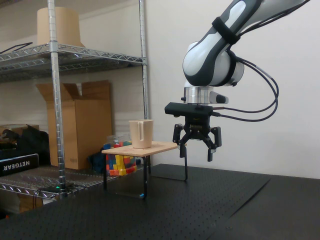} & \includegraphics[height=0.073\textheight]{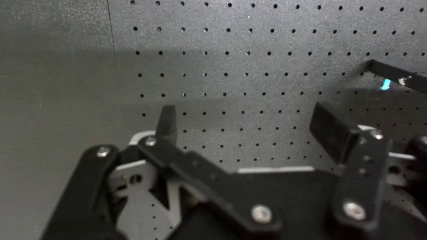} & \includegraphics[height=0.073\textheight]{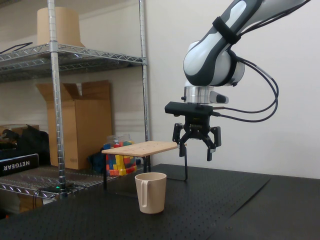} & \includegraphics[height=0.073\textheight]{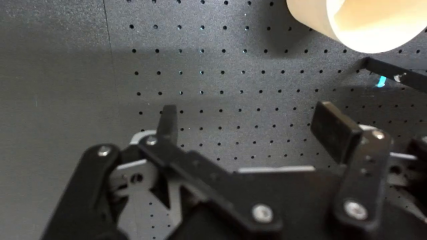} \\
\includegraphics[height=0.073\textheight]{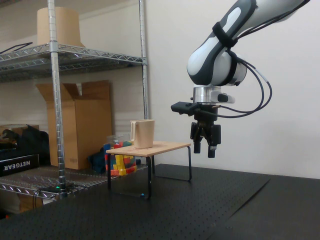} & \includegraphics[height=0.073\textheight]{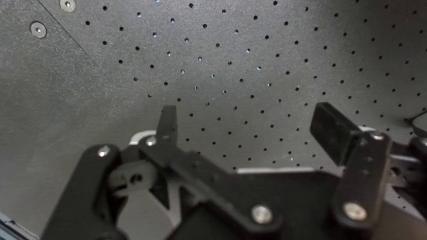} & \includegraphics[height=0.073\textheight]{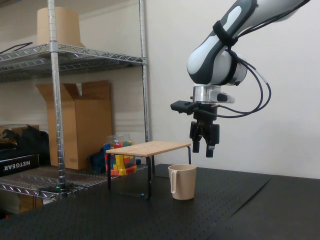} & \includegraphics[height=0.073\textheight]{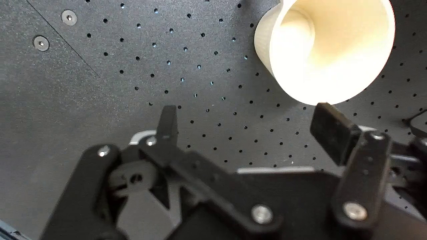} \\
\includegraphics[height=0.073\textheight]{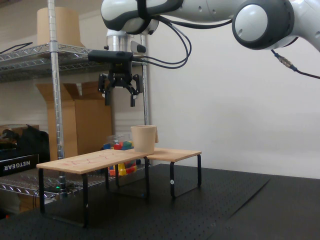} & \includegraphics[height=0.073\textheight]{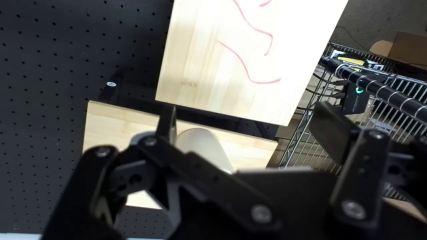} & \includegraphics[height=0.073\textheight]{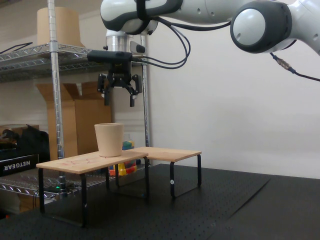} & \includegraphics[height=0.073\textheight]{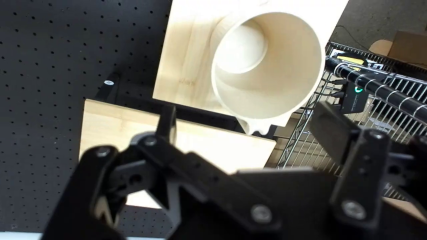} \\
\includegraphics[height=0.073\textheight]{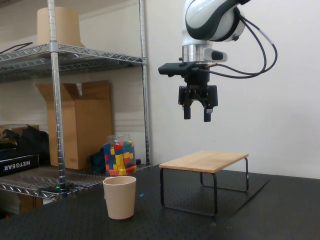} & \includegraphics[height=0.073\textheight]{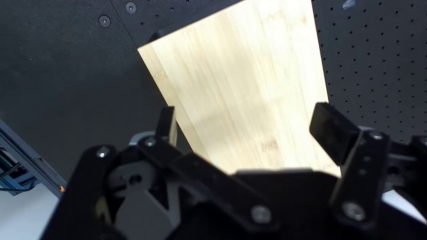} & \includegraphics[height=0.073\textheight]{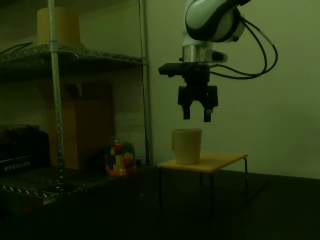} & \includegraphics[height=0.073\textheight]{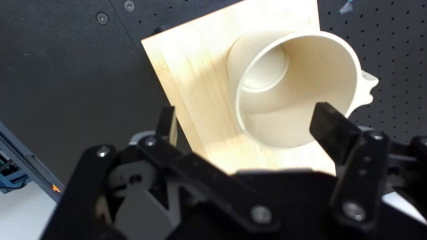} \\
\includegraphics[height=0.073\textheight]{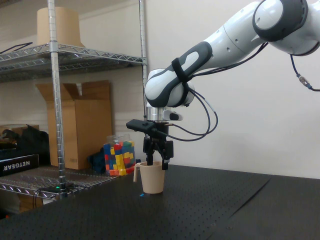} & \includegraphics[height=0.073\textheight]{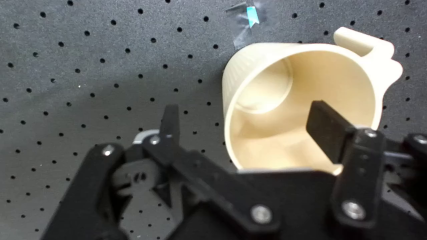} & \includegraphics[height=0.073\textheight]{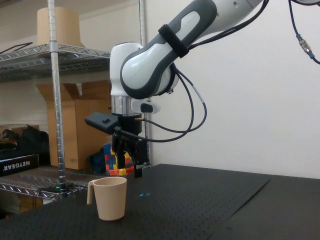} & \includegraphics[height=0.073\textheight]{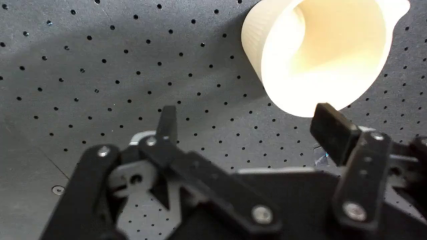} \\
\includegraphics[height=0.073\textheight]{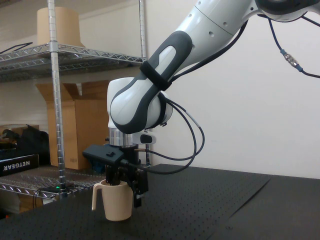} & \includegraphics[height=0.073\textheight]{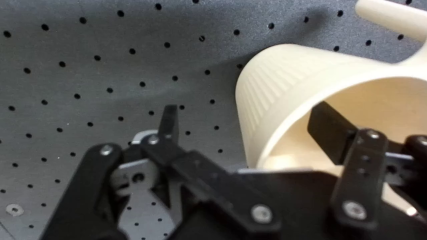} & \includegraphics[height=0.073\textheight]{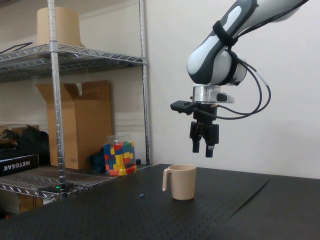} & \includegraphics[height=0.073\textheight]{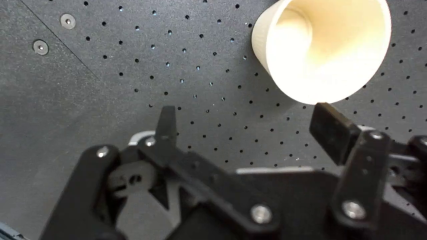} \\
\includegraphics[height=0.073\textheight]{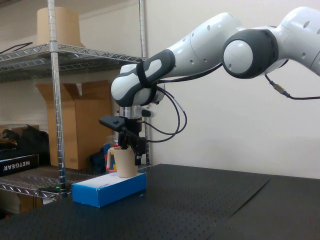} & \includegraphics[height=0.073\textheight]{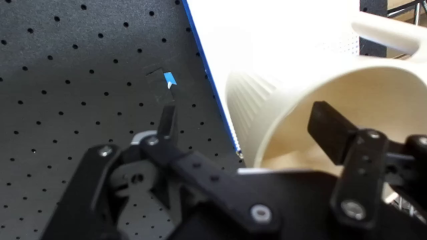} & \includegraphics[height=0.073\textheight]{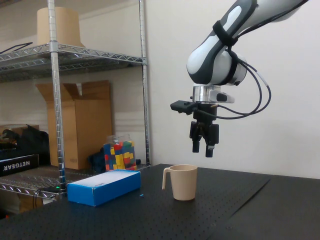} & \includegraphics[height=0.073\textheight]{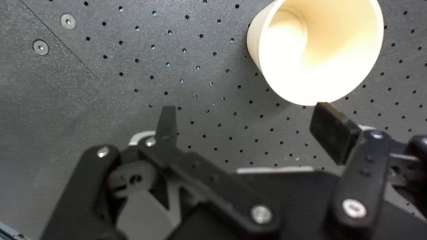} \\
\includegraphics[height=0.073\textheight]{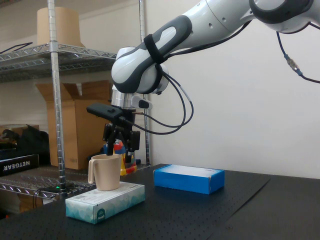} & \includegraphics[height=0.073\textheight]{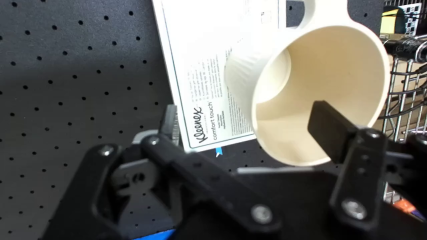} & \includegraphics[height=0.073\textheight]{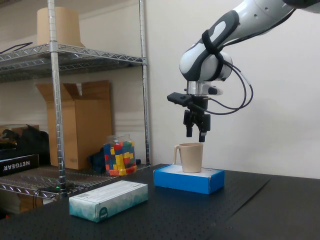} & \includegraphics[height=0.073\textheight]{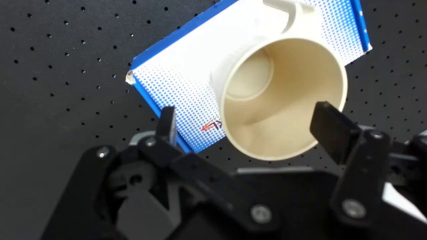} \\
\includegraphics[height=0.073\textheight]{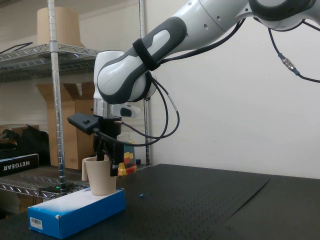} & \includegraphics[height=0.073\textheight]{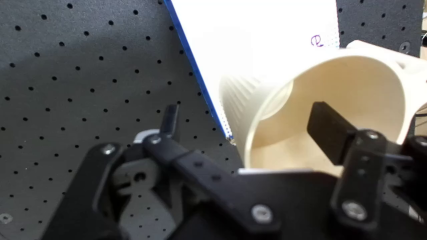} & \includegraphics[height=0.073\textheight]{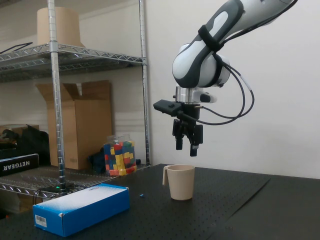} & \includegraphics[height=0.073\textheight]{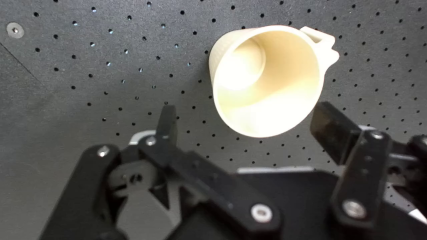} \\
\includegraphics[height=0.073\textheight]{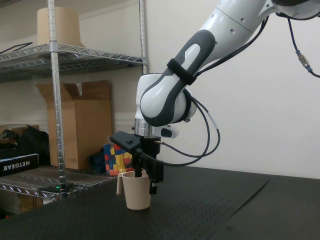} & \includegraphics[height=0.073\textheight]{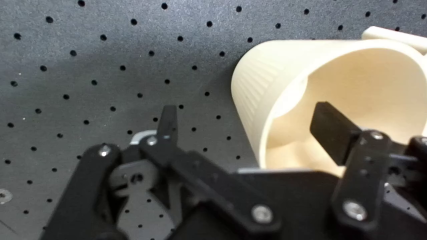} & \includegraphics[height=0.073\textheight]{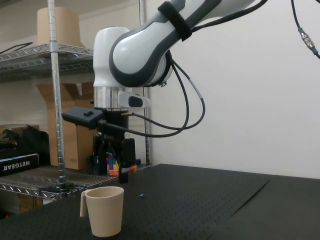} & \includegraphics[height=0.073\textheight]{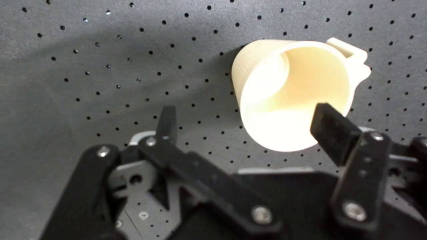} \\
\end{tabular}
\caption{\textbf{Real-robot Pick-and-Place task: goal and initial observations.} Ten episodes (rows) of the
Pick-and-Place task on the DROID/Franka platform. For each episode we show the goal image (left pair) and the
initial state (right pair), each from the external side camera and the wrist camera. The goal is
specified purely by these images; the planner drives the robot from the initial state toward the goal.}
\label{fig:deploy-pnp}
\end{figure}

\Cref{fig:deploy-grasp,fig:deploy-lift,fig:deploy-pnp} show goal and initial observations for Grasp, Object Lift, and Pick and Place, respectively.
We show goal and initial observations for the three real-robot deployment tasks (grasp, lift, and pick-and-place), ten episodes each, from both the external side camera and the wrist camera. The goal is provided to the planner as a single image; recall that the goal and the initial robot pose coincide, so the visible difference between the goal and initial columns is the configuration of the manipulated object.

\subsection{Real World Robotic Capabilities}
\label{appx:robotic-setup:real-world-discussion}

We observe qualitative differences in the behavior induced by planning with small and large world models. Small models' main weakness is poor gripper control, yet their successful episodes exhibit smooth end-effector trajectories and accurate approaches to objects. They struggle to recover after approaching an object from an unsuitable grasping position, whereas larger models can adapt and recover from such mistakes. As model size increases, behavior gradually shifts from smooth execution relying on a single grasping attempt toward more adaptive and robust execution. Intermediate models, such as the $300$M and $1$B variants, appear to lie between these behavioral regimes, combining some of the limitations of smaller models with emerging recovery capabilities.

When comparing RoboJEPA with the VLA reference models, we cannot fully disentangle the contribution of image-goal specification from that of world-model scaling. The differences in training data and controller interfaces described in \Cref{appx:robotic-setup:franka} also limit this comparison. Nevertheless, all RoboJEPA model sizes use the same goal specification and planning configuration, and the observed improvement across sizes suggests that model scaling contributes to the capabilities of the resulting planner. These qualitative observations complement the task progress and success rates reported in \Cref{tab:real-world-capabilities}.

\section{Key-Value Cache: Additional Details}\label{appx:kv}

For each predictor layer, we cache the keys and values of past timesteps so that autoregressive prediction processes only the new frame, action, and state tokens. Each timestep contains
\begin{equation}
N_{\mathrm{tok}} = V H_p W_p + n_a + n_s, \qquad n_a = n_s = 1,
\label{eq:tokens-per-step}
\end{equation}
where $V H_p W_p$ is the number of visual tokens across views. The per-layer cache tensors have shape
\begin{equation}
K^{(\ell)},\,V^{(\ell)} \in \mathbb{R}^{B\times h_{kv}\times L_{\mathrm{cache}}\times N_{\mathrm{tok}}\times d_h},
\label{eq:kv-shape}
\end{equation}
where $B$ is the batch size, $h_{kv}$ is the number of key/value heads, and $d_h$ is the head dimension. We store tokens in frame-sized slots and bound the cache by the predictor's local temporal attention window.

As prediction advances, the rolling cache replaces the oldest frame with the new one. RoPE uses absolute timestep indices, so cached positions do not need to be renumbered when the buffer wraps. Each new frame attends only to the retained window, keeping the amount of cached history bounded even for long rollouts.

At the start of each planning call, we process the observation context once and share its cached keys and values across the CEM candidates. Subsequent predictions maintain candidate-specific caches because their action sequences differ. This avoids repeatedly processing the same observation history, while each prediction step processes only the new tokens and attends to a fixed-size window of cached tokens. The compilation pipeline is described in \Cref{appx:infer}.

\section{Inference Engine: Additional Details}\label{appx:infer}

\emph{Expands \Cref{subsec:inference}.} The story of the inference engine is one long descent: start from the V-JEPA~2-AC~\citep{assran2025vjepa2} eager BF16 baseline, identify what binds it (host-side dispatch, not arithmetic; \Cref{appx:infer:dispatch}), unbind it with PyTorch Inductor lowering and ahead-of-time compilation that also makes cold start usable on a robot (\Cref{appx:infer:aot}), then squeeze the remaining matmul time with a static FP8 path on top (\Cref{appx:infer:fp8}). \Cref{appx:infer:failed} records the roads we abandoned because the workload simply was not where they help. All numbers are wall-clock measurements on a single NVIDIA H200 (140\,GB HBM3) running PyTorch~2.9, on the 1\,B predictor at 240p with grouped-query attention (12 query heads, 4 KV heads, 40 layers, 1536 hidden dim), and the rolling KV cache of $L_{\mathrm{cache}}{=}8$ frames.

\subsection{Where the time goes (why the model is not matmul-bound)}\label{appx:infer:dispatch}

\paragraph{Naive BF16 inference is bound by host-side dispatch, not by matrix multiplication.} At steady state ($B{=}16$, full KV cache, 2-view) GEMMs take only ${\sim}26\%$ of GPU time, attention ${\sim}41\%$, and fused pointwise ops (RoPE, GELU, normalizations, cache management) the remaining ${\sim}33\%$; on top of that, eager mode burns ${\sim}31\%$ of wall time in dispatch overhead. Doubling GEMM throughput would buy at most a $1.21\times$ end-to-end factor; removing dispatch and fusing the long pointwise tail are the first-order wins. At $B{=}1$ the same decomposition is even more dispatch-dominated (${\sim}47\%$ of wall time), which is why the headline speedups in \Cref{subsec:inference} are largest at small batch. Per-step kernel share within a single rollout is dominated by attention growing with cache occupancy: at step $0$ attention is ${\sim}14\%$ of step time; at step $4$ ${\sim}32\%$; at full cache (step $7^+$) ${\sim}41\%$. GEMM time stays roughly constant at ${\sim}25$\,ms per step regardless of cache depth. \emph{This is the computational-cost picture that the rest of the engine targets.}

\begin{table}[H]
\caption{Per-step decomposition into GPU kernel time and host-side dispatch overhead at $B{=}16$ on H200. AOT Inductor removes essentially all dispatch ($-65$\,ms) and additionally speeds up GPU compute by ${\sim}1.8\times$ through kernel fusion ($8{,}975 \to 1{,}291$ kernels for the 1-view model). $1$-view rows are retained here because dispatch behavior is qualitatively different at small sequence length, even though the deployed CEM planner runs the 2-view path.}
\label{tab:inference-dispatch}
\centering
\small
\begin{tabular}{l r r r r}
\toprule
Configuration & Wall (ms) & GPU kernel (ms) & Dispatch (ms) & Dispatch \% \\
\midrule
1-view eager       & 145.6 & 76.7  & 68.9 & 47.3\% \\
1-view AOT         & 45.0  & 41.9  &  3.1 &  6.8\% \\
2-view eager       & 217.1 & 148.7 & 68.4 & 31.5\% \\
2-view AOT         & 96.7  &  93.7 &  3.0 &  3.1\% \\
\bottomrule
\end{tabular}
\end{table}

\subsection{AOT Inductor compilation: from dispatch to deployment}\label{appx:infer:aot}

The natural answer to dispatch overhead is to lower the per-step graph into fused Triton kernels with PyTorch Inductor~\citep{ansel2024pytorch2}. A JIT \texttt{torch.compile} path does this, but it specializes into many shape variants on the first call; we measure $15$--$20$\,minutes of warm-up at robot start \emph{per autoregressive step pattern}, which is incompatible with closed-loop deployment where the operator turns the robot on and expects it to plan in seconds.

Autoregressive predictor inference, however, has a small predictable shape set. At step $k$ with batch size $B$, every tensor shape is fully determined; the rolling cache has only $9$ distinct configurations (steps $0$--$7$ grow the cache; step $8$ overwrites the oldest entry and repeats forever). With six compiled batch sizes $B\in\{1,2,4,8,16,32\}$ this produces $54$ static programs for the predictor; the diffusion decoder of \Cref{subsec:decoder} compiles to an analogous family of programs through the same pipeline. Each program is exported with \texttt{torch.export} under static shapes and packaged via \texttt{torch.\_inductor.aoti\_compile\_and\_package} into a $30$--$50$\,MB compiled C++ artefact.

The static-shape lowering has two prerequisites that drive engineering choices upstream. First, a \emph{mutation-free} cache: \texttt{torch.export}'s functionalisation rewrites in-place slices into hundreds of \texttt{aten.copy} ops, which inflated GPU buffer count by ${\sim}5\times$ in our first pass. We thread a functional cache through every block (each block reads the input cache tensors and returns fresh ones, the per-layer outputs are stacked, and a wrapper module bakes \texttt{cache\_size} and \texttt{input\_position} as compile-time constants). The functional path produces bitwise-identical outputs to the in-place version across all cache regimes (empty, mid, full). Second, \emph{shared weight tensors across programs}: because the $\sim$$2$\,GB BF16 weight set is held once rather than once per program, this drops resident memory from $\sim$108\,GB to ${\sim}14$\,GB --- the difference between fitting and not fitting on the deployment box. Non-compiled batch sizes are padded up to the nearest compiled value (e.g.\ $B{=}3 \to 4$) and sliced back on output, so the worst-case waste from bucketing is ${\sim}50\%$ when the requested $B$ is just above a compiled size.

End-to-end the AOT pipeline yields a $7\times$ reduction in launched kernels ($8{,}975\!\to\!1{,}291$ for the 1-view predictor), a $1.8\times$ GPU compute speedup from better fusion, and a $23\times$ reduction in dispatch overhead (\Cref{tab:inference-dispatch}). Crucially, cold start collapses from $15$--$20$\,minutes (JIT) to $1.2$--$1.5$\,seconds: the time to load the $54$ \texttt{.pt2} files and inject the shared weights. The computational cost of offline compilation ($40$--$60$\,minutes total, parallelised across multiple H200s by an offline compiler) is paid once per checkpoint; thereafter the runtime dispatcher loads the compiled archives and routes each \texttt{step()} call to the matching program. Inside the AOT pipeline we keep one positive low-level result that the failed-attempts ledger would otherwise hide: native GQA dispatched to FlashAttention~\citep{dao2022flashattention,dao2024flashattention2} avoids the implicit \texttt{repeat\_interleave} expansion that the cuDNN attention path uses, contributing the $\sim$$3\%$ on top that survives in the headline recipe.

\subsection{Static FP8 path: the last small lever}\label{appx:infer:fp8}

After AOT removes dispatch, what remains is steady-state GEMM and attention time. We can shave a final factor by running the predictor's linear layers in FP8. Because the workload is not matmul-bound, this is genuinely a \emph{small} lever --- which is itself the data point that justifies the engineering: order wins, not low-precision arithmetic. The FP8 path (\Cref{tab:inference-fp8}) runs through \texttt{torch.compile(fullgraph=True)} rather than the AOTI path because the AOTI loader currently mishandles many co-resident FP8 programs; the JIT path with the Inductor disk cache is instant after the first compile. Activation scales are calibrated once on a small held-out set and stored as fixed buffers; weights are kept as \texttt{float8\_e4m3fn} with fast-accumulate FP8 GEMMs. Outputs match the BF16 reference to within the spatial-sensitivity benchmark's noise floor (\Cref{subsec:correlations}). \Cref{tab:inference-fp8} reports per-cache-step latency at $B{=}16$ on the 2-view model: the FP8 advantage is largest at low cache occupancy where the GEMM share is relatively higher; once attention dominates at full cache it shrinks to ${\sim}5\%$.

\begin{table}[H]
\caption{Static FP8 vs.\ BF16 AOTI on the 2-view 1B predictor at $B{=}16$, averaged across the rolling-cache trajectory.}
\label{tab:inference-fp8}
\centering
\small
\begin{tabular}{l r r r}
\toprule
Cache step & BF16 AOTI (ms) & FP8 \texttt{torch.compile} (ms) & Speedup \\
\midrule
$0$       & 61.5  & 54.8  & $1.12\times$ \\
$4$       & 94.2  & 82.4  & $1.14\times$ \\
$7$--$9$  & ${\sim}108$ & ${\sim}103$ & $1.05\times$ \\
\midrule
average   & \textbf{93.1} & \textbf{84.1} & $\mathbf{1.11\times}$ \\
\bottomrule
\end{tabular}
\end{table}

\subsection{Optimisations that did not work}\label{appx:infer:failed}

We document the negative results because they sharpen the picture of what the dominant computational cost actually is.

\paragraph{Dynamic FP8.}
A first FP8 attempt with per-call activation amax computation was \emph{slower} than BF16 ($177$\,ms vs.\ ${\sim}100$\,ms): the three extra ops per linear (cast to FP32, amax reduction, quantise) have a higher computational cost than the FP8 GEMM saves on a model that is not matmul-bound. Static, pre-calibrated scales (\Cref{appx:infer:fp8}) eliminate this overhead.

\paragraph{Weight-only INT8 and W8A8 / F8A8 variants.}
The torchao implementations were either incompatible with the dispatcher's requirement that linear outputs match the BF16 KV-cache dtype required by FlashAttention, or produced ${\sim}2\times$ slower forwards because dequantisation overhead dominates on a compute-bound (not bandwidth-bound) model.

\paragraph{CUDA graphs.}
Reduce-overhead capture segfaults inside the captured graph because the KV cache slice assignment mutates a GPU tensor whose target offset changes between replays. A clone-based workaround runs but is slower ($96$\,ms vs.\ $91$\,ms) than the non-graph path: once AOT has reduced dispatch overhead to ${\sim}3$\,ms there is essentially nothing left for graph capture to amortise.

\paragraph{TensorRT.}
Both whole-model and per-block TensorRT wrappers were slower ($0.5$--$0.7\times$) and produced incorrect outputs, tripping a known multi-output-export bug. AOT Inductor already produces near-optimal fused Triton kernels for this architecture, so TensorRT did not add value.

\paragraph{RoPE pre-computation.}
Pre-computing cos/sin tables as registered buffers gave $0\%$ speedup under AOT, because Inductor already fuses RoPE inline into its surrounding Triton kernels.

\paragraph{cuDNN attention in AOT.}
\texttt{sdpa\_cudnn} is $2.4\times$ faster than \texttt{xformers\_fmha} in isolation but only ${\sim}3\%$ faster in the AOT pipeline: \texttt{torch.export} captures all SDPA variants as the same ATen op and AOT Inductor lowers to flash-attention regardless of context-manager hint. The native-GQA-without-\texttt{repeat\_interleave} factor that survives is reported as part of the AOT pipeline (\Cref{appx:infer:aot}).

\section{Scaling Laws: Additional Parametric Forms}\label{appx:scaling-laws}

\emph{Expands \Cref{subsec:scaling_laws} and \Cref{tab:scaling-extrapolation}.} In addition to the
saturating power law and the second-order power law of \Cref{subsec:scaling_laws}, we fit two
published functional forms as extrapolation baselines: the Broken Neural Scaling Law
(BNSL)~\citep{caballero2022bnsl} and the Unified Neural Scaling Law (UNSL)~\citep{caballero2026unsl}.
Here we give their full functional forms and the fitted parameter values for the single-break
($n{=}1$) instances whose extrapolation errors are reported in \Cref{tab:scaling-extrapolation}.

\paragraph{Fitting protocol.} For each episode we pick a random starting timestep, encode its initial
observation, and roll the predictor forward for nine steps under the recorded action sequence,
giving a ten-step sequence including the initial observation. We then
encode the corresponding ground truth future frames and average the $L_1$ distance between predicted
and encoder features at each token over the rollout. We average this over roughly 1{,}024 episodes
with randomized starting timesteps.

For both laws the prediction target is the loss $L$, defined as the mean
rollout-$\ell_1$ over horizons $1$--$9$. We fit each law on the $47$ checkpoints of the small models
($22$M--$2$B; a fixed set of cooldown steps per size) and report held-out extrapolation error as the
mean absolute error over the $33$ large-model checkpoints ($21$ at $4$B and $12$ at $8$B). We
deliberately avoid the $C\!\approx\!6ND$ approximation: compute is taken from the empirical per-size
accounting, $C = P_N\cdot s$, where $s$ is the number of optimization steps and $P_N$ the measured
FLOPs per step for each size ($22$M${=}1.56\!\times\!10^{15}$, \dots, $300$M${=}1.57\!\times\!10^{16}$,
\dots, $8$B${=}3.05\!\times\!10^{17}$). Training data is $D = s\cdot T_{\text{tok}}$ with
$T_{\text{tok}} = B(V_1{+}V_2)(2T{-}1)(HW{+}A{+}S) = 7{,}190{,}016$ tokens per optimization step, and
model size $N$ is the nominal parameter count. All curves are fit by minimizing the root-mean-squared
log error between predicted and observed $L$, with light $\ell_2$ regularization
($\lambda\!=\!3\!\times\!10^{-6}$) on the parameters, log-transformed and standardized inputs, and
multiple random restarts to mitigate local optima.

\subsection{Broken Neural Scaling Law (BNSL)}\label{appx:scaling-laws:bnsl}

The general BNSL functional form with $n$ breaks is~\citep{caballero2022bnsl}
\begin{equation}
L = a + \big(b\,x^{-c_0}\big)\prod_{i=1}^{n}\left(1 + \left(\frac{x}{d_i}\right)^{1/f_i}\right)^{-c_i f_i}.
\label{eq:bnsl-general}
\end{equation}
The instance in \Cref{tab:scaling-extrapolation} is univariate in compute ($x{=}C$) with a single
break ($n{=}1$),
\begin{equation}
L(C) = E + A\,C^{-\alpha}\left(1 + (C/d)^{1/f}\right)^{-\beta f},
\label{eq:bnsl-n1}
\end{equation}
with an irreducible floor $E$, a first-segment scale/exponent $(A,\alpha)$, and one smooth transition
of sharpness $f$ at compute scale $d$ that changes the log--log slope of $L-E$ from $-\alpha$ to $-(\alpha+\beta)$. It assumes the loss depends
on $N$ and $D$ only through total compute $C$.

\begin{table}[H]
\caption{Fitted BNSL ($n{=}1$, \Cref{eq:bnsl-n1}) parameters for the DROID and RoboCasa holdouts.}
\label{tab:bnsl-params}
\centering
\begin{tabular}{lcccccc}
\toprule
Dataset & $E$ & $A$ & $\alpha$ & $d$ & $f$ & $\beta f$ \\
\midrule
DROID    & $0.201$ & $0.115$ & $\approx 0$ & $3.15\times10^{19}$ & $0.651$ & $0.14$ \\
RoboCasa & $0.184$ & $3.17$  & $0.067$     & $2.60\times10^{36}$ & $3.11$  & $7.9\times10^{4}$ \\
\bottomrule
\end{tabular}
\end{table}

\subsection{Unified Neural Scaling Law (UNSL)}\label{appx:scaling-laws:unsl}

The UNSL functional form~\citep{caballero2026unsl} generalises BNSL to the multivariate setting in
which several quantities are scaled simultaneously. Its core building block is the Multivariate
Broken Neural Scaling Law (MBNSL) kernel
\begin{equation}
K = b\prod_{i} x_i^{-c_{i0}}\left(1 + \left(\frac{\prod_i x_i^{c_{i1}}}{d}\right)^{1/f}\right)^{-f}.
\label{eq:unsl-mbnsl}
\end{equation}
The instance in \Cref{tab:scaling-extrapolation} is the multivariate form in $(N,D)$,
\begin{equation}
L(N,D) = a_0 + \big(R^{-1} + A_3^{-1}\big)^{-1}, \qquad R = K_{ND} + K_{N} + K_{D},
\label{eq:unsl-n1}
\end{equation}
where each component $K$ is a broken power law with one break ($n{=}1$). Because our runs are
single-epoch, we specialize the general UNSL by (i) collapsing dataset size and training steps into a
single data axis $D$ and dropping the overfitting component, (ii) omitting the hyperparameter
(``oppositional-force'') terms since no hyperparameters are swept, and (iii) taking the evaluation
metric to be unbounded above (no $a_2$ term). Unlike the other three candidates, the UNSL is a
\emph{surface} in $(N,D)$ rather than a curve in $C$: it does not admit a closed-form $L(C)$ (the
compute-optimal envelope $L^{*}(C)=\min_N L(N,D(N,C))$ has no analytic solution for this form), so its
extrapolation error is evaluated per checkpoint at each large model's actual $(N,D)$.

\begin{table}[H]
\caption{Fitted UNSL ($n{=}1$, \Cref{eq:unsl-n1}) parameters for the DROID and RoboCasa holdouts. Each
$K$ component is an MBNSL kernel (\Cref{eq:unsl-mbnsl}); for $K_{ND}$ the exponents $c_0,c_1$ are
vectors over $(N,D)$.}
\label{tab:unsl-params}
\centering
\resizebox{\textwidth}{!}{%
\begin{tabular}{ll}
\toprule
Dataset & Fitted parameters \\
\midrule
DROID &
\shortstack[l]{$a_0{=}0.098,\ A_3{=}3.71$ \\
$K_{ND}\!:\ b{=}1.23\times10^{11},\ c_0{=}({-}0.38,1.31),\ f{=}1.81,\ d{=}2.18\times10^{9},\ c_1{=}(2.09,{-}0.68)$ \\
$K_{N}\!:\ b{=}3174,\ c_0{=}0.50,\ f{=}0.65,\ d{=}2.76,\ c_1{=}0.06$ \\
$K_{D}\!:\ b{=}1.36,\ c_0{=}0.064,\ f{=}{-}0.19,\ d{=}1114,\ c_1{=}0.27$} \\
\midrule
RoboCasa &
\shortstack[l]{$a_0{=}0.046,\ A_3{=}1.54$ \\
$K_{ND}\!:\ b{=}3505,\ c_0{=}({-}0.12,0.48),\ f{=}0.67,\ d{=}1.23\times10^{15},\ c_1{=}({-}0.75,1.81)$ \\
$K_{N}\!:\ b{=}5.03\times10^{5},\ c_0{=}0.76,\ f{=}0.55,\ d{=}1.85\times10^{-5},\ c_1{=}{-}0.55$ \\
$K_{D}\!:\ b{=}2.62\times10^{-12},\ c_0{=}{-}0.92,\ f{=}0.65,\ d{=}2.52\times10^{8},\ c_1{=}0.73$} \\
\bottomrule
\end{tabular}%
}
\end{table}

\subsection{Scaling Laws for Short- and Long-Horizon Prediction Training}\label{appx:scaling-laws:compute-cooldown}

\begin{figure}[t]
\centering
\includegraphics[width=0.72\linewidth]{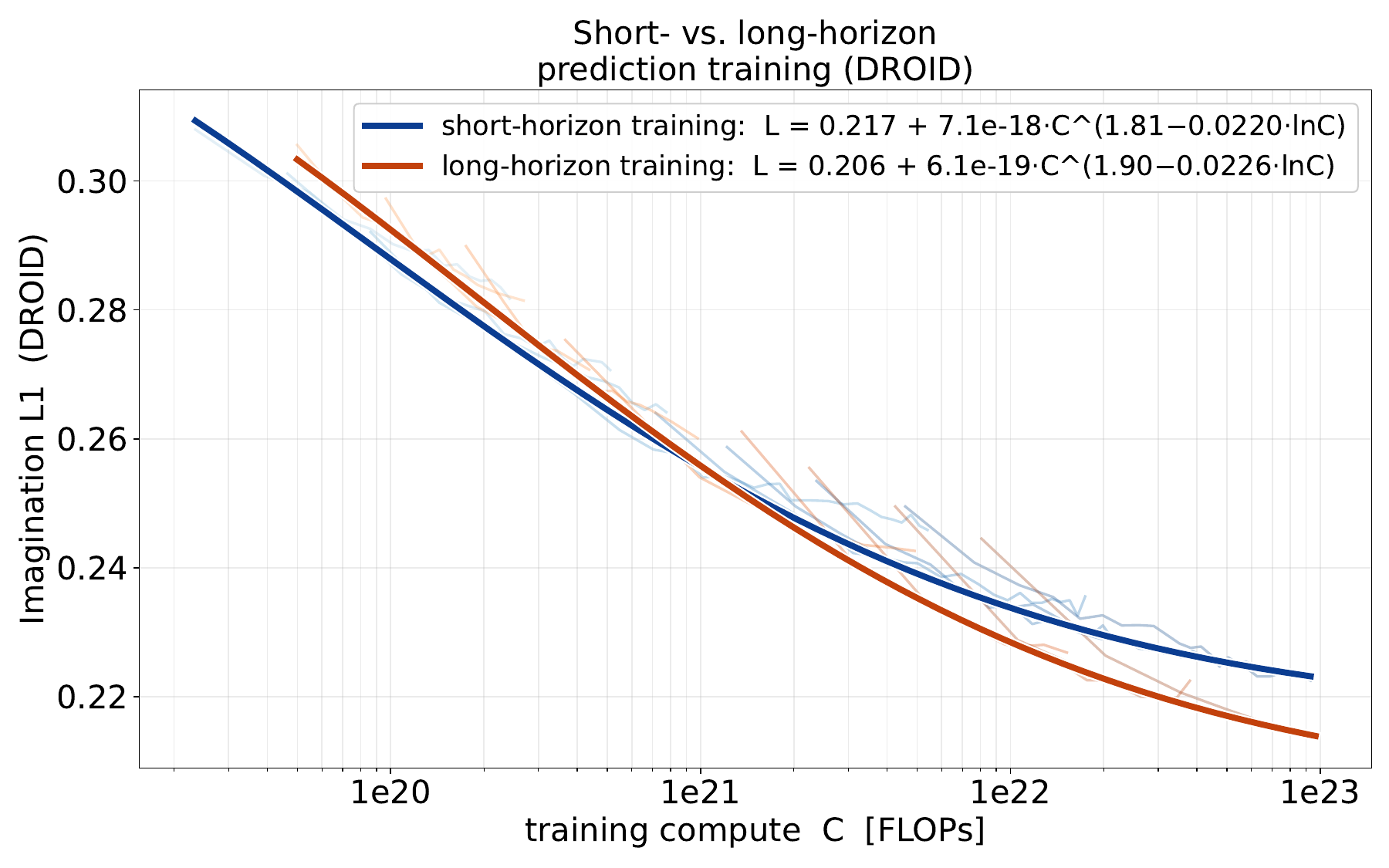}
\caption{\textbf{Short- vs.\ long-horizon prediction training on DROID.} Out-of-distribution imagination L1 on the
DROID holdout versus training compute. Faded curves are per-size checkpoints (blue: short-horizon
training, orange: long-horizon training; shade encodes model size); the bold curves are the two
second-order power laws fit on each recipe's frontier. The long-horizon training law overtakes the short-horizon training
one at high compute and reaches a lower irreducible floor.}
\label{fig:cooldown-hl-droid}
\end{figure}

\begin{figure}[t]
\centering
\includegraphics[width=0.72\linewidth]{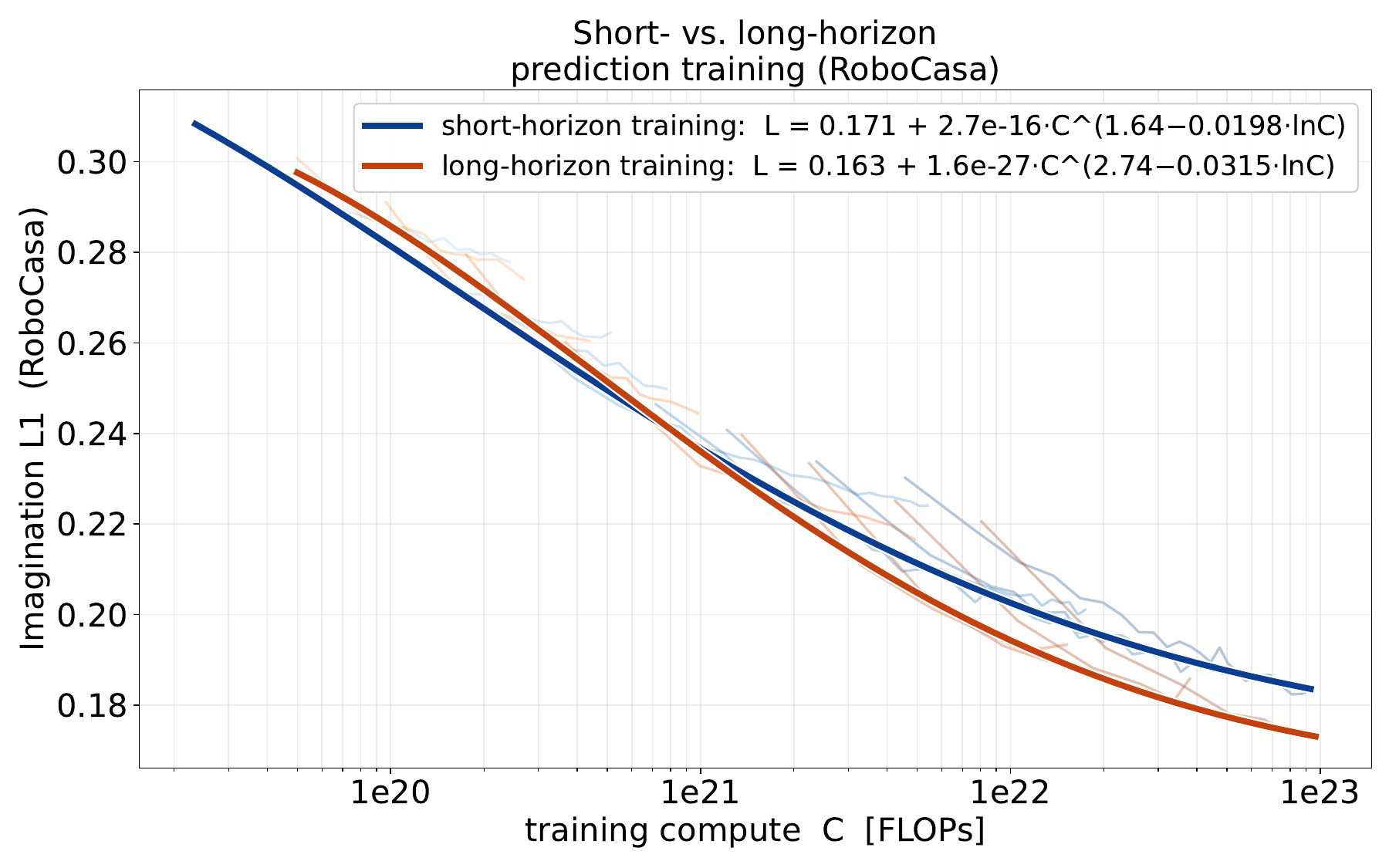}
\caption{\textbf{Short- vs.\ long-horizon prediction training on RoboCasa.} As \Cref{fig:cooldown-hl-droid}, for the
RoboCasa target evaluation. The same crossover and lower floor for long-horizon training hold
on the simulated suite.}
\label{fig:cooldown-hl-robocasa}
\end{figure}

\begin{figure}[t]
\centering
\includegraphics[width=0.72\linewidth]{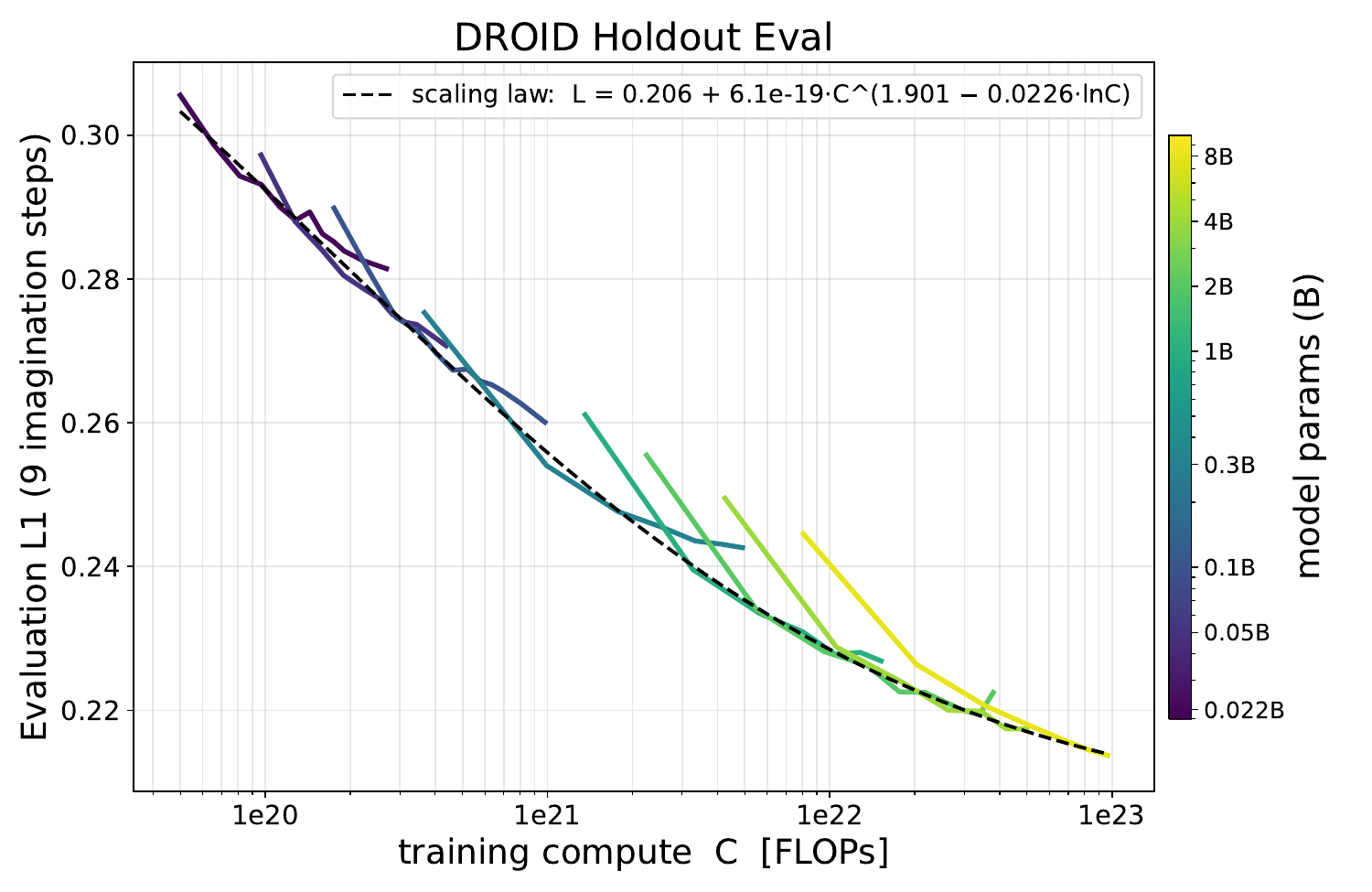}
\caption{\textbf{Long-horizon training scaling law on DROID.} Long-horizon prediction training alone: per-size
checkpoint curves colored by parameter count, with the fitted second-order power law (dashed) on the
compute-optimal frontier of the DROID holdout evaluation.}
\label{fig:cooldown-hi-droid}
\end{figure}

\begin{figure}[t]
\centering
\includegraphics[width=0.72\linewidth]{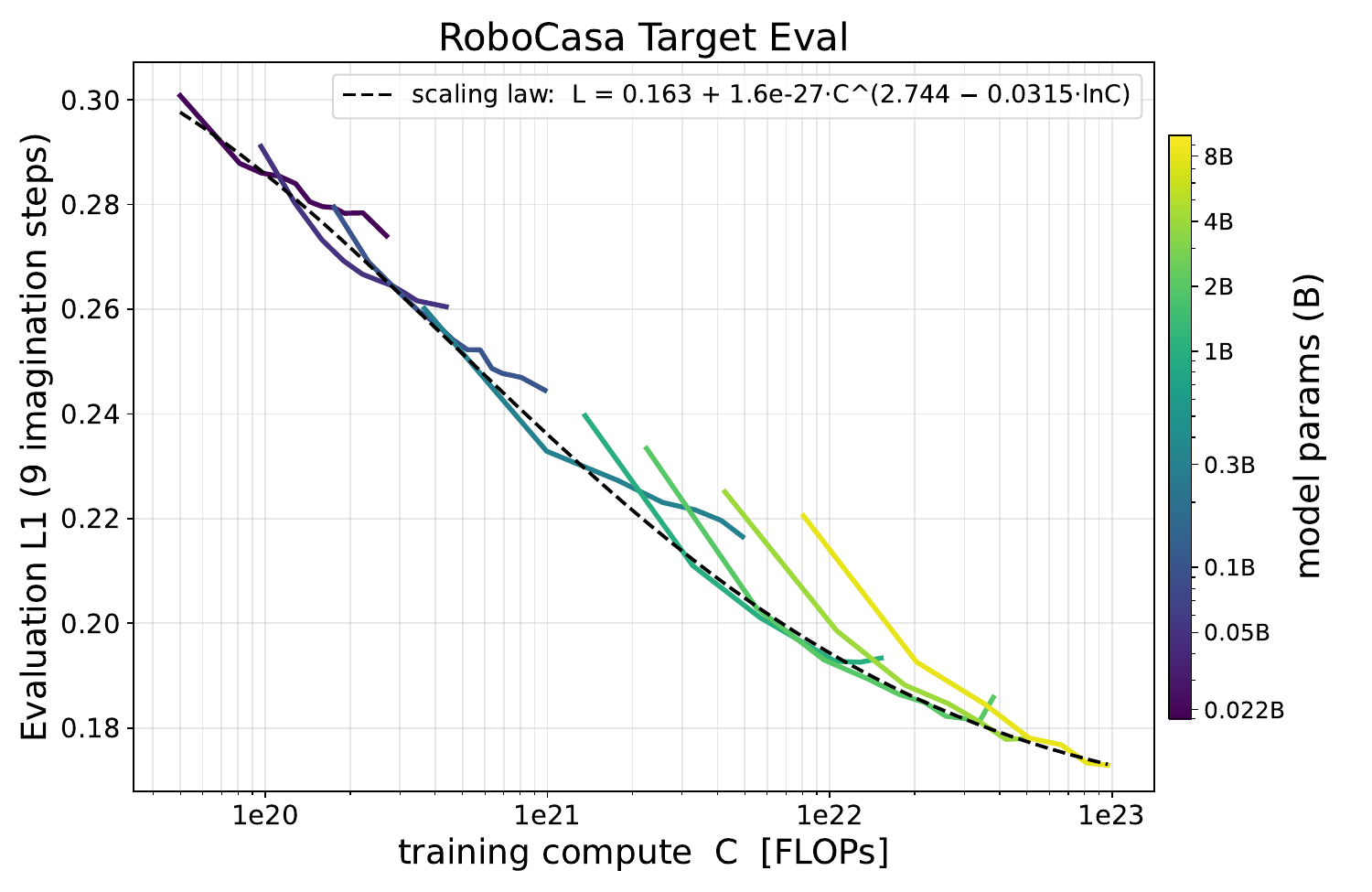}
\caption{\textbf{Long-horizon training scaling law on RoboCasa.} As \Cref{fig:cooldown-hi-droid}, for the
RoboCasa target evaluation.}
\label{fig:cooldown-hi-robocasa}
\end{figure}

\emph{Expands the ``Towards better scaling'' discussion of \Cref{subsec:towards-better-scaling}.} There we
show that spending extra compute \emph{only} in the final learning-rate annealing (cooldown) phase ---
resumed from the same flat-LR pretraining checkpoints --- improves the world model's scaling without any
new data. We compare two cooldown recipes, both fit with the second-order power law
$L(C){=}E{+}A\,C^{\alpha-\gamma\ln C}$ on the compute-optimal frontier:
\begin{itemize}
  \item \textbf{short-horizon prediction training} --- the standard cooldown: a short autoregressive rollout ($K{=}2$) at a
        single $240{\times}320$ resolution;
  \item \textbf{long-horizon prediction training} --- a cooldown with a $K{=}10$ autoregressive
        rollout computed with a KV cache, with mixed $240{\times}320$ and $480{\times}640$ resolutions, which has several times
        the computational cost per optimizer step.
\end{itemize}
The main-text \Cref{fig:high-low-compute-scaling} plots the combined (DROID${+}$RoboCasa) metric;
\Cref{fig:cooldown-hl-droid,fig:cooldown-hl-robocasa} break the same comparison down per evaluation
suite, and \Cref{fig:cooldown-hi-droid,fig:cooldown-hi-robocasa} show the long-horizon training law on its own.
On both suites the long-horizon training frontier crosses below the short-horizon training one at ${\sim}5{\times}10^{20}$
FLOPs and continues to a lower floor, i.e.\ long-horizon prediction training scales strictly better at
large compute budgets, confirming that the imagination scaling law can be improved with compute alone.

\subsection{Per-Step Training Compute across Stages}\label{appx:scaling-laws:per-step-compute}

\emph{Underpins the compute axis of \Cref{subsec:scaling_laws,subsec:towards-better-scaling}.} The
training-compute value $C$ on every scaling-law plot is accumulated as
$C = (\text{per-step FLOPs}) \times (\text{optimizer steps})$, so the per-step computational cost of each stage sets
the horizontal scale. We compute it for each run from its real configuration (view-group layout,
per-cell batch size, resolution grid, view count, rollout steps, and node/GPU count), counting both the
forward and backward pass. A single ``per step'' figure is the total the whole job performs for one
optimizer update, and therefore already includes all gradient-accumulation microbatches, all
data-parallel replicas, and --- for rollout recipes --- the autoregressive rollout unrolled over its
configured steps. We report compute in $\mathrm{PF}$ ($1\,\mathrm{PF}=10^{15}$ FLOPs).

RoboJEPA training has three stages with very different per-step computational costs. The long \emph{pretraining}
(flat learning-rate) phase and the \emph{short-horizon prediction training} cooldown share the same cheap recipe (a
short $K{=}2$ rollout at a single $240{\times}320$ resolution). The \emph{long-horizon prediction training} cooldown of
\Cref{subsec:towards-better-scaling} instead uses a $K{=}10$ rollout with a KV cache and mixed
$240{\times}320$ and $480{\times}640$ resolutions, and the \emph{DROID $720{\times}1280$ finetune}
(\Cref{appx:robotic-setup:franka}) adds three high-resolution views on top of the KV-cached $K{=}10$
rollout. \Cref{tab:per-step-compute-size} gives the per-step computational cost of the two cooldown/pretraining
recipes at every model size, and \Cref{tab:per-step-compute-8b} compares all three stages at $8$B.

\begin{table}[H]
\caption{Per-optimizer-step training compute (in $\mathrm{PF}$, i.e.\ $10^{15}$ FLOPs) for each model
size. All values report total FLOPs per optimizer step across the full training job and are used
for the compute axis of the scaling-law plots. The short-horizon training column also applies to
pretraining. The short-horizon recipe is exact; the long-horizon recipe runs its rollout with a KV cache
and is the estimate without the cache scaled by the measured KV-cache factor $0.36$.}
\label{tab:per-step-compute-size}
\centering
\small
\begin{tabular}{lcc}
\toprule
Model size & Short-horizon training & Long-horizon training \\
           & ($K{=}2$, $240{\times}320$) & ($K{=}10$, $240{\times}320 + 480{\times}640$) \\
\midrule
$22$M  & $1.56$   & $6.86$    \\
$50$M  & $3.10$   & $13.11$   \\
$100$M & $5.74$   & $23.53$   \\
$300$M & $15.66$  & $41.9$    \\
$1$B   & $47.77$  & $176.15$  \\
$2$B   & $81.17$  & $286.4$   \\
$4$B   & $157.84$ & $534.45$  \\
$8$B   & $305.36$ & $1002.75$ \\
\bottomrule
\end{tabular}
\end{table}

\begin{table}[H]
\caption{Per-step training compute at $8$B for the three stages. All values report total FLOPs
per optimizer step across the full training job. One finetune step is ${\sim}19\times$ a
long-horizon training step and ${\sim}62\times$ a short-horizon training step. The finetune uses a longer
sequence: a $720{\times}1280$ image is a $45{\times}80{=}3{,}600$-token grid per view ($\times 3$
views $=10{,}800$ tokens, versus ${\sim}300$ tokens/view at $240{\times}320$), and self-attention is
$\mathcal{O}(n^2)$ in tokens.}
\label{tab:per-step-compute-8b}
\centering
\small
\resizebox{\textwidth}{!}{%
\begin{tabular}{lcc}
\toprule
Stage & Recipe & Total per-step PFLOPs \\
\midrule
\shortstack[l]{Pretraining /\\short-horizon training} & $K{=}2$, $240{\times}320$ ($\mathrm{tp}{=}1$) & $305$\,PF \\
Long-horizon training             & $K{=}10$ with KV cache, multi-res ($\mathrm{tp}{=}2$)         & $1{,}003$\,PF \\
DROID $720{\times}1280$ finetune   & $3$-view $720{\times}1280$, $K{=}10$ with KV cache ($\mathrm{tp}{=}4$) & ${\sim}18{,}900$\,PF \\
\bottomrule
\end{tabular}%
}
\end{table}

\subsection{Stochastic upper-envelope selection}\label{appx:frontier-selection}

\emph{Extracts the compute-optimal frontier of the downstream planning success-rate (reward) curves
reported in the visual abstract (\Cref{fig:visual-abstract}).}
This procedure applies to the RoboCasa
downstream planning tasks, where higher success rate is better. Every model size contributes many
checkpoints, so at a fixed size the (compute, success-rate) pairs are stochastic rather than a single
clean curve: evaluation noise and training instabilities give a scatter of points at nearby compute
values. Fitting a scaling law requires the compute-optimal upper envelope --- the best achievable
success rate at each compute --- rather than the raw scatter, but the obvious envelope estimators are
both unsatisfying. A running maximum over compute is not robust: a single lucky, high-variance
checkpoint pulls the envelope up prematurely and the estimate never recovers. Fitting the law to all
points instead is dragged toward the mean of the scatter, since many dominated, lower checkpoints
outnumber the few near the ceiling.

We instead select frontier points with a stochastic upper-envelope procedure that tracks a running
estimate of the achievable ceiling and admits only points consistent with it, rejecting downward
outliers before they can corrupt the running estimate. Concretely, we pool all (compute, success-rate)
pairs $(C_i, y_i)$ across model sizes (higher $y$ is better) and sort them by ascending compute. The running
mean $m$ and variance $v$ are initialized from the first three points, which are always kept. Each
subsequent point is compared against the current upper trend $m - k s$, where $s=\sqrt{v}$ and
$k \geq 0$ is a rejection threshold in standard-deviation units: a point is accepted onto the frontier
only if it does not fall more than $k$ standard deviations below the running mean. Accepted points
update $m$ and $v$ with an exponentially-weighted (West) moving-average and moving-variance update
governed by a retention factor $\alpha \in (0,1)$, so recent frontier points dominate the running
estimate while stale ones decay away; rejected points leave $m$ and $v$ untouched, so a single
downward outlier can neither join the frontier nor bias the trend used to judge later points. The
scaling law is then fit only on the resulting frontier set. We use $\alpha = 0.9$ throughout, with a
per-task rejection threshold matched to the four RoboCasa downstream planning tasks of the visual
abstract: $k \approx 0.5$ for the two shorter-horizon, greedy tasks, a larger $k \approx 2$ for the
noisier long-horizon obstacle task, and no rejection ($k \to \infty$, all points kept) for the
object-push task, reflecting the differing success-rate evaluation noise of each task (short-horizon vs.
long-horizon, and whether object interaction is involved).

\begin{algorithm}[H]
\caption{Stochastic upper-envelope (frontier) selection}
\label{alg:frontier-selection}
\begin{algorithmic}[1]
\Require points $\{(C_i, y_i)\}$ pooled across model sizes (higher $y$ is better); retention factor
$\alpha \in (0,1)$; rejection threshold $k \geq 0$
\Ensure frontier index set $S$
\State Sort points by ascending compute $C$
\State Initialize running mean $m$ and variance $v$ from the first three points
\State $S \gets \{1, 2, 3\}$
\For{each subsequent point $(C_i, y_i)$}
    \State $s \gets \sqrt{v}$
    \If{$y_i > m - k s$} \Comment{accept: not an anomalous downward dip}
        \State $S \gets S \cup \{i\}$
        \State $d \gets y_i - m$
        \State $m \gets m + (1 - \alpha) d$
        \State $v \gets \alpha \big(v + (1 - \alpha) d^2\big)$
    \Else
        \State \textbf{reject} $i$ \Comment{$m, v$ left unchanged}
    \EndIf
\EndFor
\State \Return $\{(C_i, y_i) : i \in S\}$
\end{algorithmic}
\end{algorithm}

The new-point weight $(1-\alpha)$ controls how quickly the running estimate adapts: $\alpha$ close to
$1$ retains a long memory of past frontier points and adapts slowly, while smaller $\alpha$ tracks
recent points more aggressively. Because $m$ and $v$ are updated only on accepted points, the procedure
is robust to an arbitrary number of downward outliers at any compute scale, unlike a plain running
maximum. A reference implementation is provided in \texttt{make\_plot.py} as the function
\texttt{stochastic\_envelope}.

\clearpage
\section{Encoder Choice and World-Model Scaling}\label{appx:encoder-scaling}

Initially, we started the project by scaling the predictor on top of the V-JEPA~2 encoder. However, we faced numerous issues and instabilities during training. The predictor often collapsed beyond a certain model size, despite the frozen encoder, and increasing the dataset size often led to training instabilities or sometimes divergence. For both encoders, we worked with image features only. We hypothesize that this behavior can be attributed to the type of features the encoder computes.

Table~\ref{tab:encoder-published-results} restates published evaluations of V-JEPA~2~\citep{assran2025vjepa2} and V-JEPA~2.1~\citep{mur2026vjepa21} on the classification, semantic segmentation, and depth estimation benchmarks shown in Figure~2 of the latter paper, together with IntPhys2~\citep{bordes2025intphys2} results reported by \citet{punzo2026physics}. These are results from the cited papers, not evaluations conducted in this work.

Table~\ref{tab:encoder-published-results} shows that the V-JEPA~2 and V-JEPA~2.1 encoders perform on par on image and video understanding tasks, measured through image and video classification, and V-JEPA~2.1 even obtains a lower reported score on the IntPhys2 intuitive physics task than V-JEPA~2. On the other hand, V-JEPA~2.1 substantially outperforms V-JEPA~2 on dense tasks such as depth estimation and semantic segmentation. Given this large gap, we hypothesize that high-quality dense features are the most important features for scaling robotic world models, even more important than features associated with physical understanding.

\begin{table}[htbp]
\caption{Restated results from \citet{mur2026vjepa21} (Tables~8--9) and \citet{punzo2026physics} (Table~1, IntPhys2); V-JEPA~2 classification results are also reported by \citet{assran2025vjepa2} (Table~4). We distinguish global tasks (action recognition, image classification, and intuitive physics) from dense tasks (semantic segmentation and depth estimation), following the grouping in Figure~2 of \citet{mur2026vjepa21}. Classification uses attentive probes; segmentation and depth use linear probes, with frozen encoders. IntPhys2 reports mean violation-of-expectation (VOE) accuracy over three seeds with a temporal attentive probe at each model's best reported layer. Higher accuracy and mIoU, and lower RMSE, are better. Model sizes and pretraining recipes differ, so this is not a controlled ablation.}
\label{tab:encoder-published-results}
\centering
\begin{tabularx}{\linewidth}{@{}lXlrr@{}}
\toprule
Type & Benchmark & Metric & V-JEPA~2 & V-JEPA~2.1 \\
 & & & ViT-g (1B) & ViT-G (2B) \\
\midrule
Global & SSv2 action recognition & Top-1 (\%) $\uparrow$ & 77.3 & 77.7 \\
Global & K400 action recognition & Top-1 (\%) $\uparrow$ & 87.3 & 87.7 \\
Global & IN1K image classification & Top-1 (\%) $\uparrow$ & 85.1 & 85.5 \\
Global & IntPhys2~\citep{bordes2025intphys2} & VOE (\%) $\uparrow$ & 66.0 & 58.8 \\
\midrule
Dense & ADE20K semantic segmentation & mIoU (\%) $\uparrow$ & 24.4 & 47.9 \\
Dense & NYUv2 depth estimation & RMSE $\downarrow$ & 0.642 & 0.307 \\
\bottomrule
\end{tabularx}
\end{table}

\stopcontents[appendix]

\end{document}